\documentclass[a4paper,fleqn]{cas-dc}

\usepackage[numbers]{natbib}
\usepackage{hyperref}
\usepackage{textcomp} 
\usepackage{rotating}

\usepackage{fancyhdr}
\usepackage{lastpage}
\usepackage{amsmath,amssymb,amsfonts}
\usepackage{graphicx}
\usepackage{textcomp}
\usepackage{xcolor}
\usepackage{algorithm}
\usepackage{algpseudocode}
\usepackage{subcaption}
\usepackage{amsmath}

\usepackage[utf8]{inputenc} 
\usepackage[T1]{fontenc}
\usepackage{CJKutf8} 

\def\tsc#1{\csdef{#1}{\textsc{\lowercase{#1}}\xspace}}
\tsc{WGM}
\tsc{QE}
\tsc{EP}
\tsc{PMS}
\tsc{BEC}
\tsc{DE}

\begin{document}
\let\WriteBookmarks\relax
\def\floatpagepagefraction{1}
\def\textpagefraction{.001}

\shorttitle{Leveraging LLM-based Room-Object Relationships
Knowledge for Enhancing Multimodal-Input Object Goal
Navigation}

\shortauthors{Leyuan Sun et~al.}

\title [mode = title]{Leveraging Large Language Model-based Room-Object Relationships Knowledge for Enhancing Multimodal-Input Object Goal Navigation}                      
\tnotemark[1,2]

\tnotetext[1]{This document is the results of the research project, JPNP20006, commissioned by the New Energy and Industrial Technology Development Organization (NEDO).}

\tnotetext[2]{This document is partially supported by JSPS KAKENHI Grant Number 23H03426 in Japan.}

%
\author[1]{Leyuan Sun}

\cormark[1]


\ead{son.leyuansun@aist.go.jp}



\affiliation[1]{organization={Computer Vision Research Team, Artificial Intelligence Research Center (AIRC), National Institute of Advanced Industrial Science and Technology (AIST)},
    city={Tsukuba, Ibaraki},
    postcode={305-8560}, 
    country={Japan}}

\author[2]{Asako Kanezaki}
\affiliation[2]{organization={Tokyo Institute of Technology},
    city={Meguro, Tokyo},
    postcode={152-8550}, 
    country={Japan}}

\ead{kanezaki@c.titech.ac.jp}

\author[3,4]{Guillaume Caron}
\affiliation[3]{organization={University of Picardie Jules Verne, MIS laboratory},
    city={Amiens},
    postcode={80025}, 
    country={France}}

\ead{guillaume.caron@u-picardie.fr}

\author[1,4]{Yusuke Yoshiyasu}
\affiliation[4]{organization={CNRS-AIST Joint Robotics Laboratory (JRL), IRL, National Institute of Advanced Industrial Science and Technology (AIST)},
    city={Tsukuba, Ibaraki},
    postcode={305-8560}, 
    country={Japan}}

\ead{yusuke-yoshiyasu@aist.go.jp}





\cortext[cor1]{Corresponding author}



\begin{abstract}
Object-goal navigation is a crucial engineering task for the community of embodied navigation; it involves navigating to an instance of a specified object category within unseen environments. Although extensive investigations have been conducted on both end-to-end and modular-based, data-driven approaches, fully enabling an agent to comprehend the environment through perceptual knowledge and perform object-goal navigation as efficiently as humans remains a significant challenge. Recently, large language models have shown potential in this task, thanks to their powerful capabilities for knowledge extraction and integration. In this study, we propose a data-driven, modular-based approach, trained on a dataset that incorporates common-sense knowledge of object-to-room relationships extracted from a large language model. We utilize the multi-channel Swin-Unet architecture to conduct multi-task learning incorporating with multimodal inputs. The results in the Habitat simulator demonstrate that our framework outperforms the baseline by an average of 10.6\% in the efficiency metric, Success weighted by Path Length (SPL). The real-world demonstration shows that the proposed approach can efficiently conduct this task by traversing several rooms. For more details and real-world demonstrations, please check our project webpage (\url{https://sunleyuan.github.io/ObjectNav}).
\end{abstract}



\makeatletter\def\Hy@Warning#1{}\makeatother
\maketitle 
\section{Introduction}
\begin{figure}[!t]
	\centering
		\includegraphics[scale=.45]{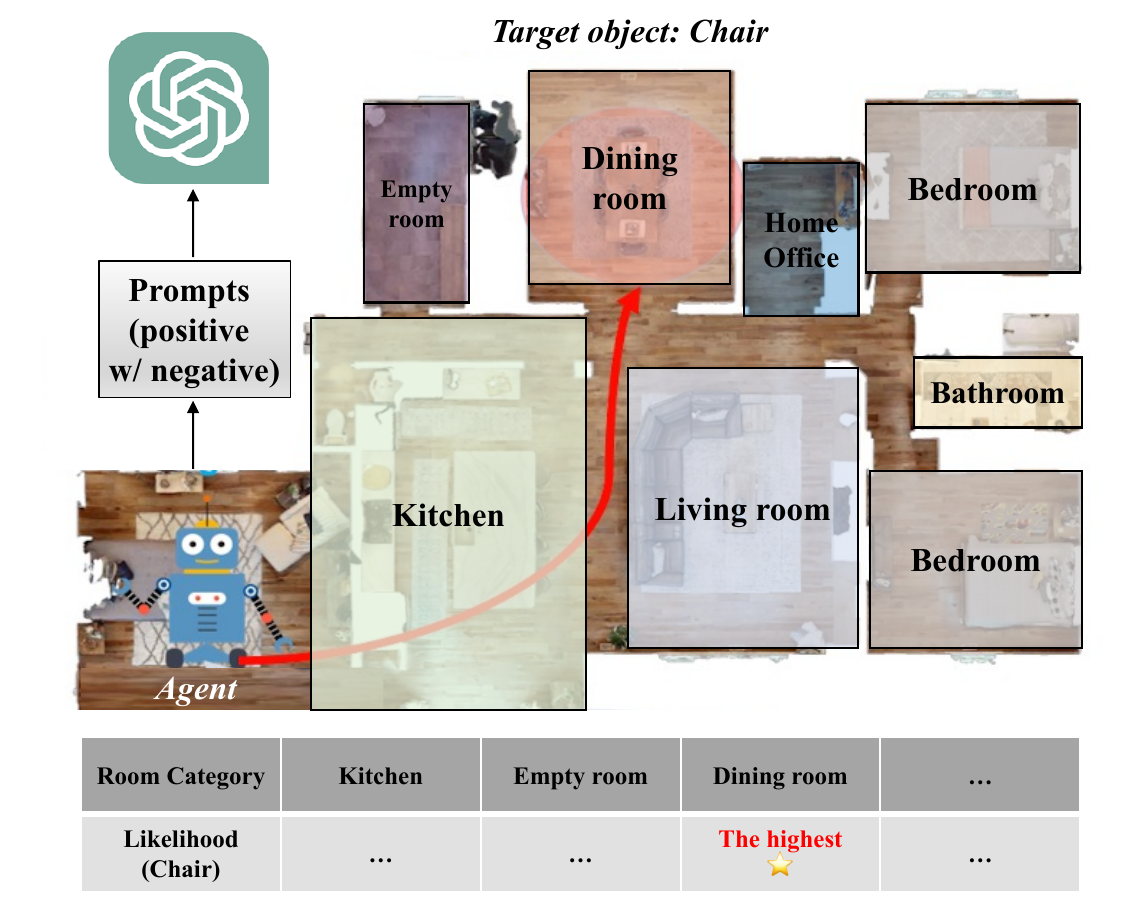}
	\caption{This study proposes utilizing LLM-based knowledge of object-to-room relationships to improve the efficiency of the object-goal navigation task. Positive and negative prompts are combined to determine the likelihood of the target object's presence in various room categories. }
	\label{FIG:proposal}
\end{figure}

The demand for indoor robots that can autonomously navigate in unseen real-world scenarios is growing for various applications, including rescue missions, assisting the disabled, and vacuum cleaning \cite{rayguru2023introducing}. A crucial capability for these robots is the ability to locate a specific object and move towards it, a task known as Object Goal Navigation (ObjectNav) \cite{anderson2018evaluation}. For example, the agent is given a target object category, e.g., a chair, as shown in Fig. \ref{FIG:proposal}, and it must explore this unseen environment to find it successfully without wasting time on unrelated spaces, in order to achieve better performance in the evaluation of the efficiency metric.

Since the release of the Habitat Challenge in 2019 and the introduction of standardized definitions and evaluation metrics, there has been growing interest in research on ObjectNav. Initial studies in ObjectNav primarily explored end-to-end approaches \cite{mousavian2019visual}\cite{liu2020indoor}\cite{shen2019situational}\cite{de2020deep}, mapping observations directly to actions. These studies largely focused on utilizing visual representations to enhance decision-making in policies. However, such approaches demand extensive computational power and time. In response, modular techniques \cite{semexp}\cite{ramakrishnan2022poni}\cite{yu2023l3mvn}\cite{zhai2023peanut}\cite{FSE} have been developed. These strategies combine the benefits of traditional pipelines with learning-based approaches, aiming to mitigate the demands on computational resources and time consumption. Modular techniques are structured around a mapping component that depicts the surroundings, a network that sets a long-term goal objective, and a local deterministic route planner for goal-directed movement.

Compared with end-to-end approaches in real-world applications, modular-based methods have achieved a 90\% Success Rate (SR) in extensive real-world house environment testing, thanks to robust deterministic sub-modules and domain-agnostic grid maps as inputs \cite{li2022object}\cite{wang2023survey}\cite{semexp}\cite{gervet2023navigating}. In contrast, end-to-end approaches have achieved the 77\% SR. However, there is still a gap that needs to be explored and improved in the evaluation of the efficiency metric, Success weighted by Path Length (SPL). The challenge lies in enabling the embodied agent to understand common-sense knowledge from perceptions as humans do (e.g., a bed is typically found in the bedroom, a TV has a higher possibility of being in the living room than in the bathroom).

Some data-driven, modular-based reinforcement learning (RL) approaches utilize only the distance to the target as a reward \cite{semexp}\cite{FSE}, while other supervised learning approaches directly predict the position of the target object \cite{ramakrishnan2022poni}\cite{zhai2023peanut}. However, even though they implicitly learn the relationship between the target and observed objects using semantic maps, both approaches do not fully exploit the geometric and semantic cues present in the map. On the other hand, some studies have explicitly explored the relationships between objects, represented as nodes \cite{zhou2023optimal}\cite{chen2023think}\cite{MON} (e.g., chairs near tables, couches near TVs). These studies suggest that understanding such relationships can enhance the efficiency of ObjectNav tasks using the graph transformer architecture \cite{yun2019graph}. However, these approaches can lead to complexity in the state space as scenes change, making navigation through expansive state spaces inefficient due to redundancy and inaccuracies in prior knowledge \cite{chen2023think}.

Recently, the ability of knowledge extraction and integration within Large Language Models (LLMs) for robotics has been extensively investigated \cite{zeng2023large}. However, when relying solely on LLMs \cite{yu2023l3mvn}\cite{shah2023navigation}, performance is limited when insufficient semantic observations available to enable reasoning for efficient decision-making. To address these issues, we propose an approach known as LROGNav (\textbf{L}LM-based \textbf{R}oom-\textbf{O}bject \textbf{G}oal \textbf{Nav}igation). This method merges the benefits of data-driven, modular-based methodologies with the incorporation of common-sense knowledge from LLMs regarding object-to-room relationships, aiming to improve the efficiency of ObjectNav.

As illustrated in Fig. \ref{FIG:proposal}, we utilize positive and negative promptings to query the LLM about the likelihood between the target object and different room categories as a form of common sense knowledge. This information is then injected into augmented room segmentation maps for supervised training. Similar to our baseline approach, PONI \cite{ramakrishnan2022poni}, LROGNav employs a frontier-based weights prediction approach. Its primary task is to predict the score on the frontiers, where a higher score indicates closer proximity to the target object. One of the auxiliary tasks predicts the score on the frontier representing unexplored space, while the other concerns the Object-to-Room (O2R) relationship score, where a higher score indicates a greater likelihood of the target object appearing in that room. Then, these weights are combined to decide the long-term goal, while using a deterministic local path planner, the fast marching method \cite{sethian1999fast}, to gradually approach the goal until the target is found. 

The model is trained using a multi-channel Swin-Unet \cite{cao2022swinunet}\cite{faulkenberry2024visual} with multimodal inputs, leveraging the Transformer architecture's capability for multimodal fusion \cite{xu2023multimodal}. We use the CLIP \cite{clip} model to estimate room categories during inference as room-related knowledge for the O2R auxiliary task learning. The proposed LROGNav is evaluated both on the Habitat simulator \cite{ramrakhya2022habitat} and in a real-world, house-like environment with the Kobuki mobile robot. The demonstrations show that LROGNav outperforms the baseline approach in simulation, particularly in terms of efficiency, while also navigating through multiple rooms to locate the target effectively in real-world scenarios.

In summary, our study has the following main contributions:
\begin{enumerate}
  \item We proposed a supervised, modular-based ObjectNav approach named LROGNav, which utilizes common-sense knowledge of object-to-room category relationships extracted from the large language model via chain-of-thought positive with negative promptings.
  \item To inject the common-sense knowledge extracted from LLM into the training dataset, we generated sufficent room segmentation floor maps based on both the Gibson \cite{xia2018gibson} and Matterport3D \cite{chang2017matterport3d} photorealistic datasets. LROGNav has been trained using a multi-channel Swin-Unet encoder-decoder architecture that incorporates multimodal inputs, to fully leverage the benefits of multimodal fusion.
  \item Simulation experiments have been evaluated on both the Gibson and Matterport3D. The proposed LROGNav achieves competitive SR compared with prior techniques. Regarding the SPL metric representing the efficiency of ObjectNav, LROGNav shows an average improvement of 10.6\% compared to the second-best related works across both datasets.
  \item We deployed our framework on a real robot for real-world experiments. The demonstrations verified the concept of the proposed approach, which guides the agent in traversing several rooms to find the target efficiently. The encountered issues have been discussed in the context of transferring to future real-world applications.
\end{enumerate}

\begin{table*}[!t]
\caption{The comparison of existing visual navigation approaches and our proposed LROGNav.
}
\label{relatedworks}
\renewcommand\arraystretch{1.5}
\begin{center}
\scalebox{0.47}{
\begin{tabular}{c|cccccc}
\hline
Method                    & Type & Task & Common sense reasoning & Network & Dataset & \begin{tabular}[c]{@{}c@{}}Real-world\\ deployment\end{tabular} \\ \hline
SemExp (NeurIPS 2020) \cite{semexp}     &  Modular-based RL    & ObjectNav     &  Distance to target as reward for long-term goal position      & Fully connected layers      & Gibson \cite{xia2018gibson} \& MP3D \cite{chang2017matterport3d}       &     \begin{tabular}[c]{@{}c@{}}Yes \\  (across several rooms)  \end{tabular}                      \\ \hline
PONI (CVPR 2022) \cite{ramakrishnan2022poni}  &  Modular-based supervised learning      &  ObjectNav     & Distance to target \& Area occupancy    &        U-Net            &  Gibson \& MP3D     &      No                          \\ \hline
L3MVN (IROS 2023) \cite{yu2023l3mvn}        & Modular-based (zero-shot)     &  ObjectNav    &  LLM-based object-to-object relationship                  &  Training not required      &    Gibson \& HM3D \cite{ramakrishnan2021habitathm3d}     &   \begin{tabular}[c]{@{}c@{}}Yes \\  (across several rooms)  \end{tabular}                        \\ \hline
FSE-VN (ICRA 2023) \cite{FSE}       &  Modular-based RL    &  ObjectNav     &  Distance to target as reward for frontier selection                  &   Fully connected layers    &       Gibson \& HM3D   &      \begin{tabular}[c]{@{}c@{}}Yes \\  (in single scene/room)  \end{tabular}                     \\ \hline
RIM (IROS 2023) \cite{chen2023objectRIM}          & End-to-end RL     & ObjectNav     &      \begin{tabular}[c]{@{}c@{}}Behavior cloning via expert trajectory\\ \& Auxiliary tasks (visual feature, explicit map, etc.)\end{tabular}                          &  Multi-layer Transformer     &   MP3D      &          \begin{tabular}[c]{@{}c@{}}Yes \\  (in single scene/room)  \end{tabular}     \\ \hline
PEANUT (ICCV 2023) \cite{zhai2023peanut}       &  Modular-based supervised learning     &  ObjectNav    &  Predict the unseen target object position via semantic map                  &  PSPNet \cite{zhao2017pyramid}      &    HM3D \& MP3D     &    No                   \\ \hline
LFG (CoRL 2023) \cite{shah2023navigation}           & Modular-based (LLM as proposal)     &    ObjectNav  &  LLM-based Object-to-object relationship               &   Training not required    &    HM3D     &          \begin{tabular}[c]{@{}c@{}}Yes \\  (in single scene/room)  \end{tabular}                \\ \hline
MON (IEEE TIP 2023) \cite{MON}      &  Modular-based RL    &   \begin{tabular}[c]{@{}c@{}}Multi-object navigation\\ (include single object)\end{tabular}    &   \begin{tabular}[c]{@{}c@{}}Object-to-object relation graph \\ \& Distance to target as reward\end{tabular}                &    Fully connected layers   &  Gibson \& MP3D        &        No               \\ \hline
CER (IEEE TCSVT 2023) \cite {chen2023think}        & End-to-end RL     &  ObjectNav    & Predict the object-to-object relationship via contrastive learning                   &  Graph Transformer \cite{yun2019graph}      &    MP3D     &  No                     \\ \hline
CKR (IEEE TPAMI 2023) \cite{CKR}     &   End-to-end RL   &   Embodied referring expression     &  Room-room and object-object relation graphs from CLIP                  &    Transformer encoder-decoder   &   R2R \cite{anderson2018vision}      &      No                 \\ \hline
ESC (ICML 2023) \cite{zhou2023esc}   &  Modular-based (zero-shot)    & Zero-shot ObjectNav \cite{majumdar2022zson}      &   LLM-based object and room reasoning                 &      Training not required  &    MP3D \& HM3D \& RoboTHOR    &      No                \\ \hline
GTV (ADV ENG INFORM 2023) \cite{zhou2023optimal}   & End-to-end RL     & ObjectNav     &   Object-to-object relation graph                 &  Graph Transformer    &   AI2THOR \cite{kolve2017ai2}      &  No                     \\ \hline
LROGNav (Ours) & Modular-based supervised learning & ObjectNav     & \begin{tabular}[c]{@{}c@{}}Distance to target\\ \& Area occupancy\\ \& LLM-based O2R relationship\end{tabular}     &  \begin{tabular}[c]{@{}c@{}}Multi-channel Swin U-Net\\ w/ multi-decoders\end{tabular}                & Gibson \& MP3D      & \begin{tabular}[c]{@{}c@{}}Yes \\  (across several rooms)  \end{tabular}                                \\ \hline
\end{tabular}
}
\end{center}
\label{tab:related}
\end{table*}
\section{Related work}


\subsection{Visual navigation}
Tasks in Embodied Navigation are categorized by their goals as follows: Point Goal Navigation \cite{ye2021auxiliary} (aiming for a precise coordinate, for instance, moving 7m south and 8m east from the starting position), Object Goal Navigation \cite{semexp} (searching for a specific object, such as a couch), Zero-Shot Object-goal Navigation (ZSON) \cite{al2022zero} (open-vocabulary ObjectNav), Area Goal Navigation \cite{savva2017minos} (targeting a defined area, such as the bedroom), and Vision-and-Language Navigation (VLN) \cite{anderson2018vision} (navigating according to a descriptive route text, for example, proceed past the dining table and through the corridor straight ahead). In this study, the proposed LROGNav specifically addresses the challenge of the ObjectNav task. 
\subsubsection{End-to-end approach}
End-to-end approaches directly translate observations into actions utilizing reinforcement learning (RL) or imitation learning (IL). Within this structure, the agent initially encodes observations to visual features suitable for the input of the policy network. Subsequently, the policy network is trained on action selection, guided by the rewards received through environmental interaction.

Chen \emph{et al.} proposed the RIM \cite{chen2023objectRIM}, which is an implicit spatial map designed to enhance end-to-end learning techniques for navigating towards specific objects. This map is dynamically updated with fresh observations through the use of a transformer, facilitating continual refinement. Chen \emph{et al.} introduced Continuous Environmental Representations (CER) \cite{chen2023think}, which uniquely leverages the agent's capability to envision spatial and semantic information beyond its Field of View (FoV) and delves CER through self-supervised learning techniques. Zhou \emph{et al.} developed a knowledge graph and Graph Transformer Viterbi Network (GTV) \cite{zhou2023optimal}, which utilizes a commonsense knowledge graph for encoding and facilitates action inference in the end-to-end manner, leading to the derivation of an optimal policy that enhances the agent’s capability for exploration.

End-to-end strategies seamlessly integrate the entire navigation process without the necessity for explicit map construction, offering a straightforward approach. Current studies delve into various visual representations \cite{chen2023objectRIM} and policy development. However, there remains a considerable gap for these strategies to become applicable in real-world scenarios since these methods must master localization, mapping, and route planning simultaneously. Moreover, the transition from simulated to real environments presents challenges due to discrepancies between simulated and real-world perceptions \cite{gervet2023navigating}. Furthermore, the computational demands and time requirements for end-to-end learning, utilizing reinforcement or imitation learning \cite{ye2021auxiliaryred}\cite{ramakrishnan2022poni}, pose significant hurdles for deployment in real-world navigation applications. Consequently, there is growing interest in investigating modular-based approaches to mitigate these issues.

\subsubsection{Module-based approach}

Instead of adopting an end-to-end approach, modular techniques typically divide the system into separate components, including mapping, long-term goal prediction, and path-planning modules. Initially, the agent processes observations to identify key features. It then employs the mapping module to construct a representation of the environment, such as a grid map \cite{semexp}\cite{ramakrishnan2022poni} or graph \cite{MON}. These extracted features, along with the environmental representation, are utilized by the long-term goal network module to predict a long-term objective. Subsequently, the path-planning module uses this long-term goal to determine the subsequent actions.

Chaplot \emph{et al.} developed SemExp \cite{semexp}, which is the initial modular-based approach utilizing a semantic top-down 2D map for semantic reasoning. It adopts an interactive RL approach that uses the distance to the target as the reward. Subsequently, FSE-VN \cite{FSE}, PEANUT \cite{zhai2023peanut}, and PONI \cite{ramakrishnan2022poni} utilized the same semantic mapping module from SemExp \cite{semexp}, differing primarily in the long-term goal prediction module. FSE-VN combines word embeddings with the frontier map to train a policy for deciding which frontier to explore, similar to the traditional frontier-based approach \cite{yamauchi1997frontier}. Meanwhile, PEANUT \cite{zhai2023peanut} directly predicts the location of the target object in a supervised manner. However, PONI \cite{ramakrishnan2022poni} demonstrated that implicitly predicting the area and object potential functions yields better performance than directly predicting the location of targets.

Contrary to end-to-end approaches, modular techniques segment ObjectNav tasks into distinct modules: mapping, long-term goal formulation, and path planning. The mapping module offers a domain-independent representation, distinguishing perception from policy formulation and path planning, thereby enhancing applicability for real-world transfer. Additionally, modular approaches generally require fewer computational resources and are less time-intensive to some extent \cite{ramakrishnan2022poni}. However, in this structured approach, the importance of the mapping module cannot be overstated, as it significantly influences the decision-making process.
\subsubsection{LLM-based approach}

Because the information content of text instructions in ObjectNav is relatively low compared to VLN, many LLM-based VLN studies have attempted to integrate forecasts from language models with planning or probabilistic analysis \cite{shah2023lm}. The goal is to reduce reliance on solely using the language model's progressive predictions for initiating actions. Instead, these approaches aim to eliminate impractical strategies, such as preventing a robot from attempting unsuitable actions, and focus primarily on understanding and executing instructions rather than relying on language models for semantic guidance. Conversely, our approach not only explicitly extracts common-sense knowledge from LLMs for semantic reasoning but also maintains robustness even when the extracted knowledge does not align correctly with the observations. This is because the proposed LROGNav is data-driven. It doesn't solely rely on LLM-based common-sense guidance but also predicts the distance to the target and the direction of unexplored spaces as part of a multi-task learning process.

L3MVN \cite{yu2023l3mvn}, utilizing a combination of LLM and search strategies, slightly outperforms FBE \cite{yamauchi1997frontier} but does not fully leverage the semantic potential of the LLM. While it shares similarities with LROGNav and LFG \cite{shah2023navigation}, it encounters some limitations: firstly, it adopts a zero-shot approach that requires no training, relying solely on the LLM. This approach offers limited reliability when scant semantic information is detected, as the capability of the LLM to reason using its inherent common-sense knowledge is reduced; secondly, it employs a basic logarithmic probability-based scoring method, which has been shown to be less efficient \cite{shah2023navigation}. LFG \cite{shah2023navigation} combines LLM-based semantic reasoning, using object-to-object relationships, with traditional FBE \cite{yamauchi1997frontier} strategy, which takes over the navigation process in the event of LLM failures. To some extent, this takeover mechanism improves robustness by compensating for incorrect decisions made by the LLMs. 

Differing from traditional ObjectNav approaches, ZSON focuses on improving the ability to identify objects from open-vocabulary categories and achieving greater efficiency during the exploration process. Zhou \emph{et al.} developed a ZSON method called Exploration with Soft Commonsense constraints (ESC) \cite{zhou2023esc}. This technique employs commonsense knowledge from pre-trained models for object navigation in unfamiliar settings, eliminating the need for prior navigation experience or additional training. However, these semantic planning methods consistently execute decision-making at predetermined intervals. This can lead to agents setting intermediate goals from suboptimal locations due to limited data, impeding the full realization of the LLM’s inferential capabilities \cite{wu2024voronav}.

\subsection{Common sense knowledge reasoning}
We categorize all the approaches mentioned above in Table \ref{tab:related}. To enhance the efficiency of ObjectNav tasks, a crucial component is providing the robot with the capability for semantic reasoning. Typically, this skill is mainly developed through training that incorporates common-sense knowledge from object-to-object \cite{yu2023l3mvn}\cite{chen2023think}\cite{shah2023navigation}\cite{zhou2023optimal}\cite{MON}, object-to-room \cite{zhou2023esc}, and room-to-room \cite{CKR} relationships, as well as the layout and geometry of the environment, whether implicitly or explicitly.

Many map-based approaches use the distance to the target as a supervised signal to train the long-term goal prediction module \cite{semexp}\cite{FSE}\cite{ramakrishnan2022poni}. This allows for implicit learning of semantic reasoning based on the room's layout and the relationships between the target object and observed objects. Such learning is enhanced by the creation of a semantic map that includes historical memory, progressively enriching the agent's understanding over time. The Graph Transformer, as introduced in \cite{yun2019graph}, provides a methodical framework for assembling graphs that illustrate the relationships among objects, where each relationship is represented by weighted edges connecting various nodes. This setup is instrumental for facilitating semantic reasoning, as it encapsulates the intricacies of object-to-object interactions within a structured format \cite{chen2023think} \cite{zhou2023optimal} \cite{MON}. Imitation learning represents another viable method for acquiring semantic reasoning skills for ObjectNav, wherein the model is trained via behavior cloning \cite{chen2023objectRIM}. This training involves utilizing expert trajectories from datasets, allowing the model to learn optimal navigation strategies by replicating the actions of proficient agents. 

Another approach to imbue semantic reasoning capabilities involves utilizing LLMs, as mentioned earlier \cite{yu2023l3mvn}\cite{zhou2023esc}. Furthermore, language models can be enhanced by incorporating direct image analysis capabilities, akin to those found in foundational vision-language models \cite{chen2024mapgpt} like GPT-4V. This development marks a significant advancement in augmenting the utility of language models for decision-making scenarios. This integration not only enriches the understanding derived from textual data but also bridges the gap between visual and linguistic information, promoting a more comprehensive approach to semantic reasoning.

\subsection{Multi-modal learning with Transformer}

Multimodal Machine Learning (MML) has emerged as a significant field of study over the past several decades, playing a crucial role in human societal interactions. This is because we exist in a world characterized by multimodal inputs and outputs. This is particularly relevant for an AI-powered navigation robot that utilizes various sensor technologies to effectively navigate the physical world. The agent processes multimodal information, including, but not limited to, RGB-D observations, 3D pose data from GPS and compass, textual instructions, and even audio signals (as in audio-visual navigation \cite{chen2021semantic}). In the era of deep learning, advancements in deep neural networks have significantly accelerated progress in MML. The Transformer architecture, in particular, has introduced new challenges and opportunities in the MML domain. Specifically, the remarkable achievements of LLMs underscore the transformative impact and versatility of Transformer-based approaches, laying the groundwork for multimodal applications.

Exploring the advantages of Transformers in the realm of MML remains a significant and unresolved area. Key insights derived from existing studies include \cite{xu2023multimodal}: (1) Transformers possess the capability to capture underlying knowledge without explicit direction. (2) Their architecture, characterized by multiple heads, permits diversification in modeling approaches, thereby broadening the model's interpretative capabilities. (3) By their design, Transformers are adept at consolidating information from a wide array of sources, identifying patterns that extend beyond immediate local interactions. (4) The capability to conceptualize input data in graph form allows for seamless integration across different data types. (5) Their substantial architecture allows them to better navigate and adapt to complex domain variations, such as those between linguistic and visual information, particularly with comprehensive pretraining across extensive datasets. (6) In contrast to RNNs, Transformers exhibit enhanced efficiency in both training and inference phases for sequential data analysis, like time series, benefiting from their ability to perform parallel computations. (7) Finally, their tokenization process provides a flexible framework for handling diverse multimodal inputs. This ability to reconfigure data inputs makes Transformers a versatile tool in the MML field.

\section{Methodology}
\subsection{ObjectNav task definition}
Following the definition of the Object-Goal Navigation task as outlined in \cite{semexp}\cite{anderson2018evaluation}, the ObjectNav task requires the agent to find a specific target category object (\textit{T}) in an unknown environment. The inputs to the embodied agents are RGB-D images and the agent's pose (\textit{x, y, $\theta$}) at each timestamp. The agent is expected to find the target objects via an efficient path, using four output actions: \texttt{move\underline{ }forward}, \texttt{turn\underline{ }left}, \texttt{turn\underline{ }right}, and \texttt{stop}. A successful case is identified when the \texttt{stop} action is activated while the agent is less than 1 meter away from the target object. The entire search episode should last no more than 500 timestamps.
\subsection{Method overview}
\begin{figure*}[!t]
	\centering
		\includegraphics[scale=.5]{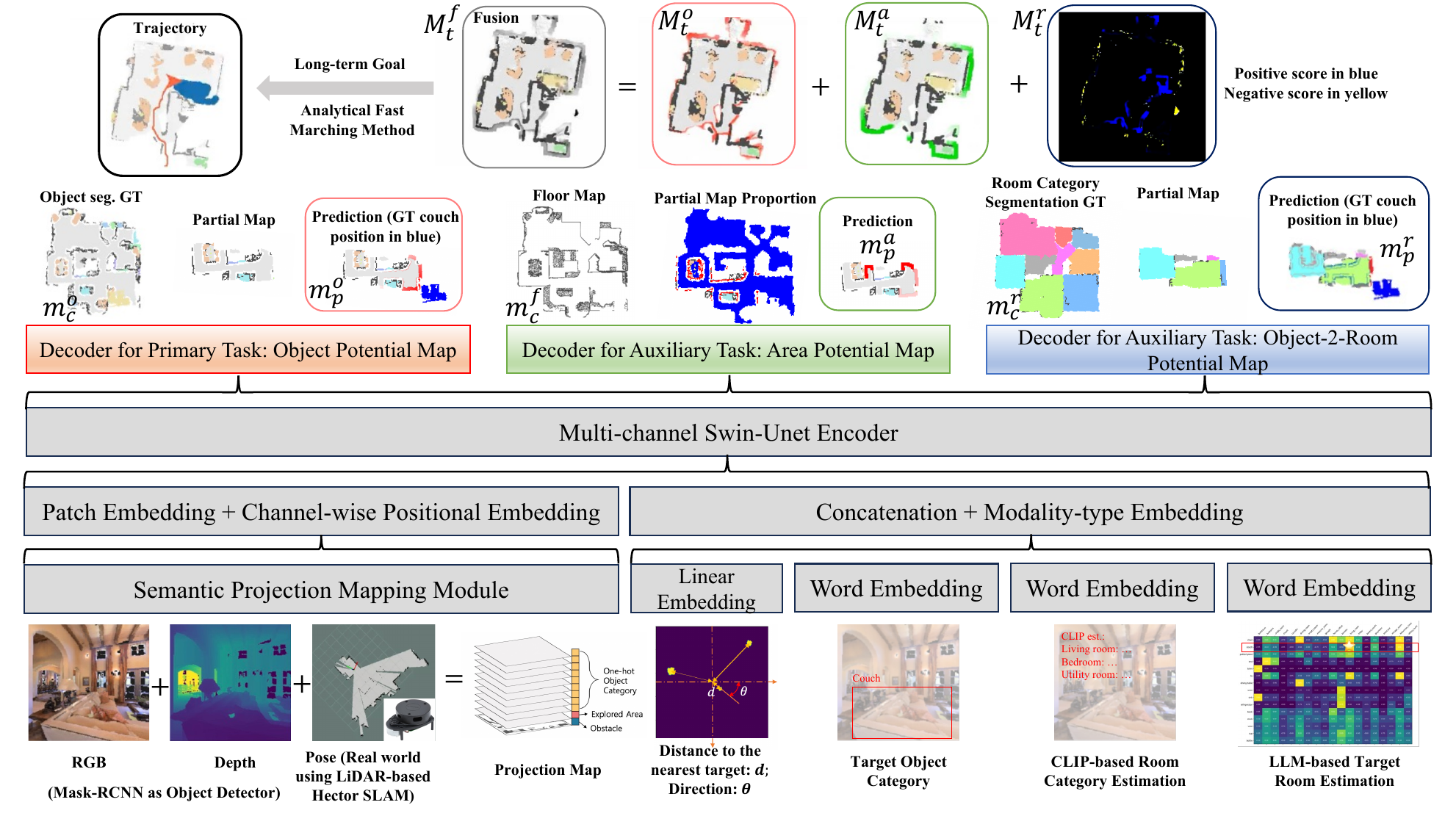}
	\caption{This overview illustrates the proposed approach LROGNav. It involves encoding RGB-D with pose data for a semantic projection mapping module. The direction and distance to the nearest target object are captured and projected using linear embedding, along with three other word embeddings: the target object, CLIP-based room category estimations, and LLM-based potential rooms. A multi-channel Swin-Unet is employed to integrate these modalities. The primary task by one of decoders is to predict frontiers close to the target object. One auxiliary task focuses on predicting frontiers that require further exploration, while another auxiliary task assigns high scores to those frontiers located in rooms with a high likelihood of containing target objects, as pre-determined by LLM-based knowledge. These three tasks are combined to determine the long-term goal, followed by an analytical method to gradually approach the goal until the target is detected.}
	\label{FIG:overview}
\end{figure*}
Figure \ref{FIG:overview} illustrates the overview of the proposed approach, namely LROGNav. LROGNav is a hierarchical, modular-based approach that integrates multimodal inputs and multi-task learning. Observations, including RGB-D images and pose at each timestamp, are projected onto a semantic 2D floor map. This map uses multiple channels to represent different object categories, as identified by Mask R-CNN \cite{he2017mask}. Another modality, projected by a linear embedding layer, captures the direction and distance to each nearest object category. Three word embeddings are used to encode the target object category, the CLIP-based \cite{clip} room category estimations, and LLM-based potential target room estimations. 

A multi-channel Swin-Unet encoder with three decoders integrates these modalities for the primary and auxiliary tasks. In the object-goal navigation task, the primary task is to approach the target. Therefore, $M^{t}$ is a map that assigns high values to frontiers close to the target object. Collecting semantic reasoning in an unknown environment is also vital as one of the auxiliary tasks. The high-value frontier in $M^{a}$ indicates the direction with more unexplored space that requires further exploration. To enhance search efficiency utilizing LLM-based common sense knowledge, another auxiliary task assigns high scores to frontiers in rooms that are highly likely to contain target objects, as represented by $M^{r}$. 

By fusing the three tasks, the long-term goal, which is the frontier with the highest value, can be determined. An analytical path planner is used to guide the agent towards this long-term goal until the target object is detected, within the maximum number of episodes. Next, we will introduce the individual components of LROGNav, which include LLM promptings, the Object-2-Room dataset, network architecture, loss functions, and more.

\subsection{LLM-based Object-2-Room relationship knowledge}
\label{sec:o2r relationship}

\begin{figure}[ht]
	\centering
		\includegraphics[scale=.32]{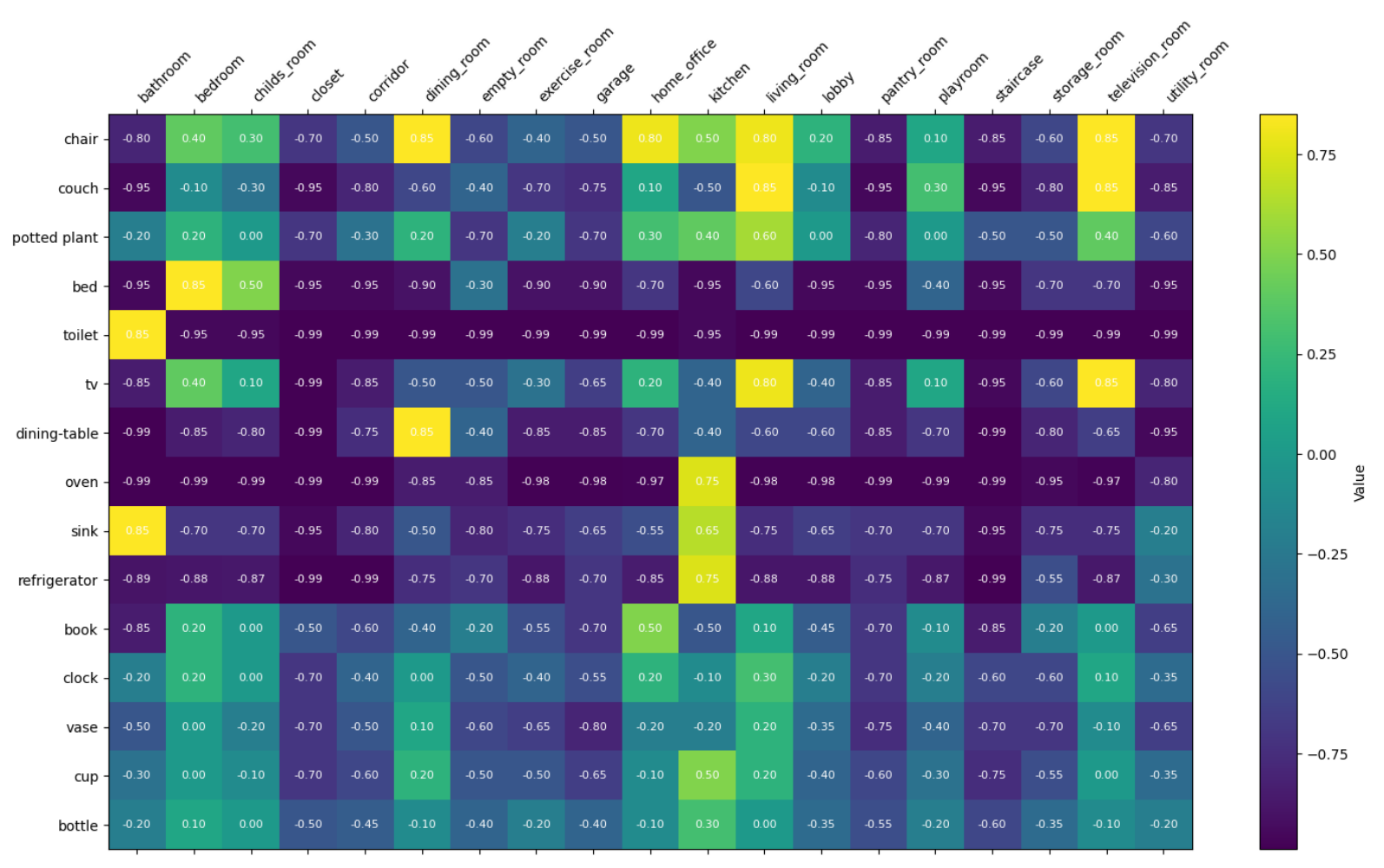}
	\caption{The Object-2-Room relationship matrix utilizing LLM-based knowledge. In the Gibson dataset, the room categories represented on the y-axis are: \texttt{"bathroom", "bedroom", "child's room", "closet", "corridor", "dining room", "empty room", "exercise room", "garage", "home office", "kitchen", "living room", "lobby", "pantry room", "playroom", "staircase", "storage room", "television room", "utility room"}. The object categories on the x-axis include: \texttt{"chair", "couch", "potted plant", "bed", "toilet", "tv", "dining table", "oven", "sink", "refrigerator", "book", "clock", "vase", "cup", "bottle"}.}
 \label{FIG:matrix}
\end{figure}

\begin{figure*}[ht]
	\centering
		\includegraphics[scale=.53]{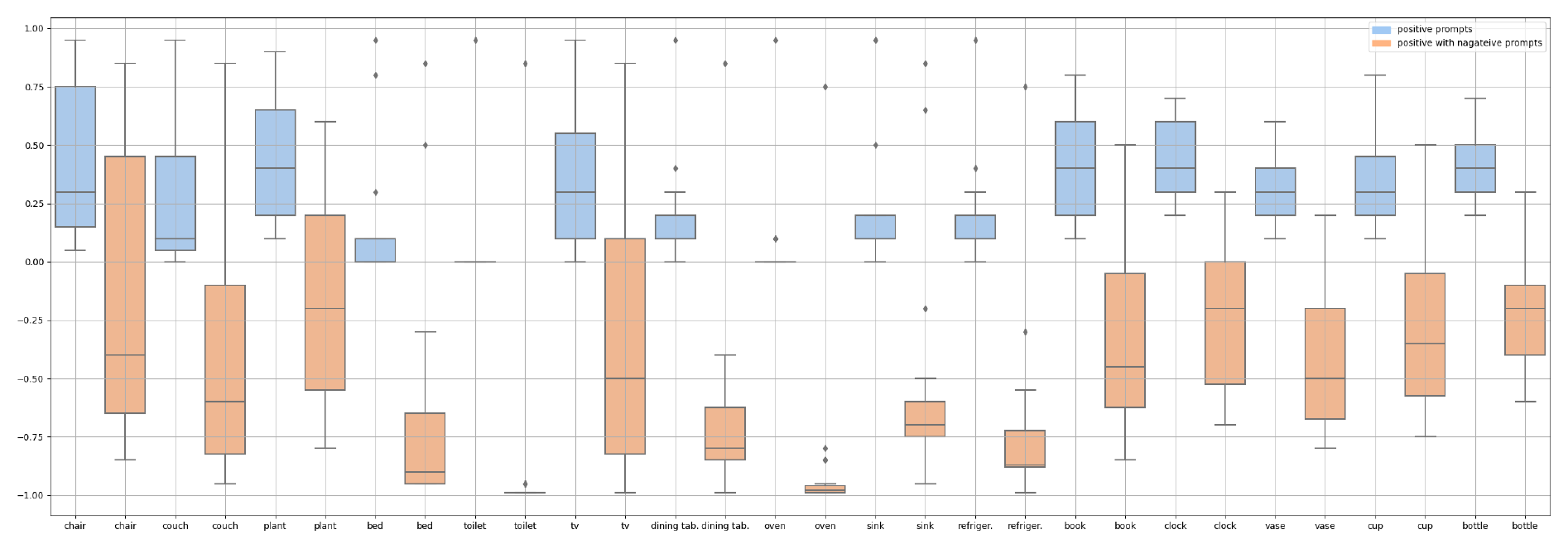}
	\caption{The box plot compares the scores of "Positive only" prompts ($LLM_{pos}(r,o)$) with those of "Positive w/ negative" prompts ($LLM_{pos}(r,o) - LLM_{neg}(r,o)$).}
 \label{FIG:prompts score}
\end{figure*}

Humans have the ability to identify the most likely room when searching for a target object, which is a significant piece of common sense knowledge for the ObjectNav task. To collect this object-to-room relationship knowledge, we query GPT-4 to rank different room categories for each target object based on likelihood estimation.

Inspired by Dhruv \emph{et al.} \cite{shah2023navigation}, who proposed querying to extract likelihood estimations between target objects and observed object clusters in the ObjectNav task, we designed prompts to infer the object-to-room relationships. In our designed prompts, we rely not only on positive prompts (e.g., in CLIPGraphs \cite{agrawal2023clipgraphs}, \texttt{"Which of the following rooms would you expect to find a "target\_object" in?"}), but also on negative prompts (e.g., \texttt{"Which of the following rooms is least likely to be relevant for the "target\_object"?"}). Relying solely on the scores of positive prompts may lead to uncertainty about any subgoal in the LLM, as verified in \cite{shah2023navigation}. Additionally, Chain-of-Thought (CoT) prompting \cite{wei2022chain} is a common strategy to elicit reliable likelihood estimations from LLM \cite{shah2023navigation}, enhancing the LLM's interpretability and reasoning capabilities.

In summary, we collected all room ($r$) and object categories ($o$) from the Gibson \cite{xia2018gibson} and Matterport3D \cite{chang2017matterport3d} datasets (see Fig. \ref{FIG:matterpord3d_matrix} in Appendix). The scores of CoT positive prompts $LLM_{pos}(r,o)\in [0,1]$ are combined with negative scores $LLM_{neg}(r,o)\in [0,1]$ for each $r$ and $o$. We plotted the relationship matrix as shown in Fig. \ref{FIG:matrix}, representing the score $h_{r,o}$ in Eq. (\ref{eq:score}): 

\begin{equation}
LLM_{pos}(r,o) - LLM_{neg}(r,o) = h_{r,o} \in [-1,1]
\label{eq:score}
\end{equation}

Observing Fig. \ref{FIG:matrix}, we can identify distinct room-specific objects. For instance, the oven, sink, and refrigerator predominantly appear in the kitchen, while the toilet is primarily found in the bathroom. We believe these specific characteristics could be leveraged to enhance the efficiency of the ObjectNav task. Additionally, objects like plants, TVs, books, clocks, vases, cups, and bottles have a relatively high likelihood of appearing in multiple rooms, making them ubiquitous objects.

As illustrated by the box plot in Fig. \ref{FIG:prompts score}, we compare the score distributions between the "Positive only" and "Positive w/ negative" prompts. By presenting the differences in probabilities of occurrence and non-occurrence, we can more clearly identify which objects are strongly related to specific rooms. For example, if an object has a high difference value in a particular room, this suggests that its likelihood of appearing in that room is significantly greater than not appearing, indicating a strong correlation. We aim for positive scores to guide the agent towards rooms more likely to contain the target object, while negative scores dissuade the agent from less likely rooms. Detailed CoT prompts are in Appendix materials, and the ablation study comparing "Positive only" with "Positive w/ negative" for ObjectNav is elaborated in Section \ref{sec:ablation}.

\subsection{Object-2-Room relationship dataset}
To incorporate the previously obtained LLM-based O2R knowledge into the dataset for training an ObjectNav task, the first step is to acquire the room segmentation dataset. Both the Gibson \cite{xia2018gibson} and Matterport3D \cite{chang2017matterport3d} datasets provide ground truth annotations of object and room/area categories for each house (e.g., the 3d scene graph \cite{armeni20193dscene} with annotations provided for Gibson). In this subsection, we outline the process of generating a room-segmented 2D floor map and how this map is combined with the LLM-based O2R knowledge to train our proposed approach, LROGNav.

\begin{figure*}[ht]
	\centering
		\includegraphics[scale=.56]{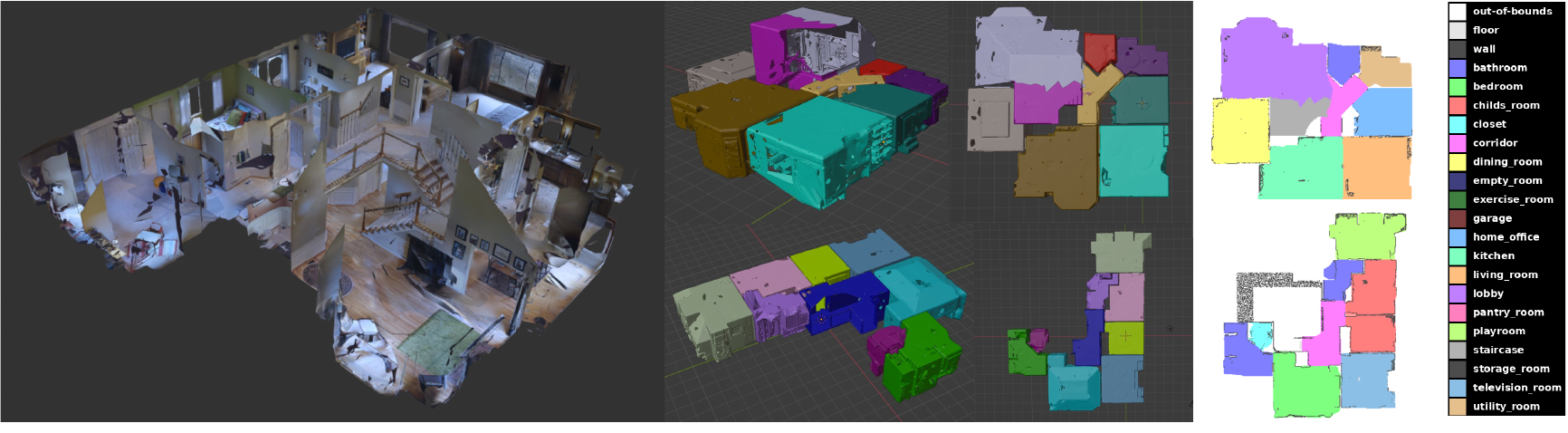}
	\caption{One example from the Gibson \cite{xia2018gibson} dataset is the Beechwood house. From left to right, it shows the mesh with texture, the mesh with room annotations, and the room segmentation for each floor.}
	\label{FIG:room_seg}
\end{figure*}

\begin{figure*}[bh]
	\centering
		\includegraphics[scale=.5]{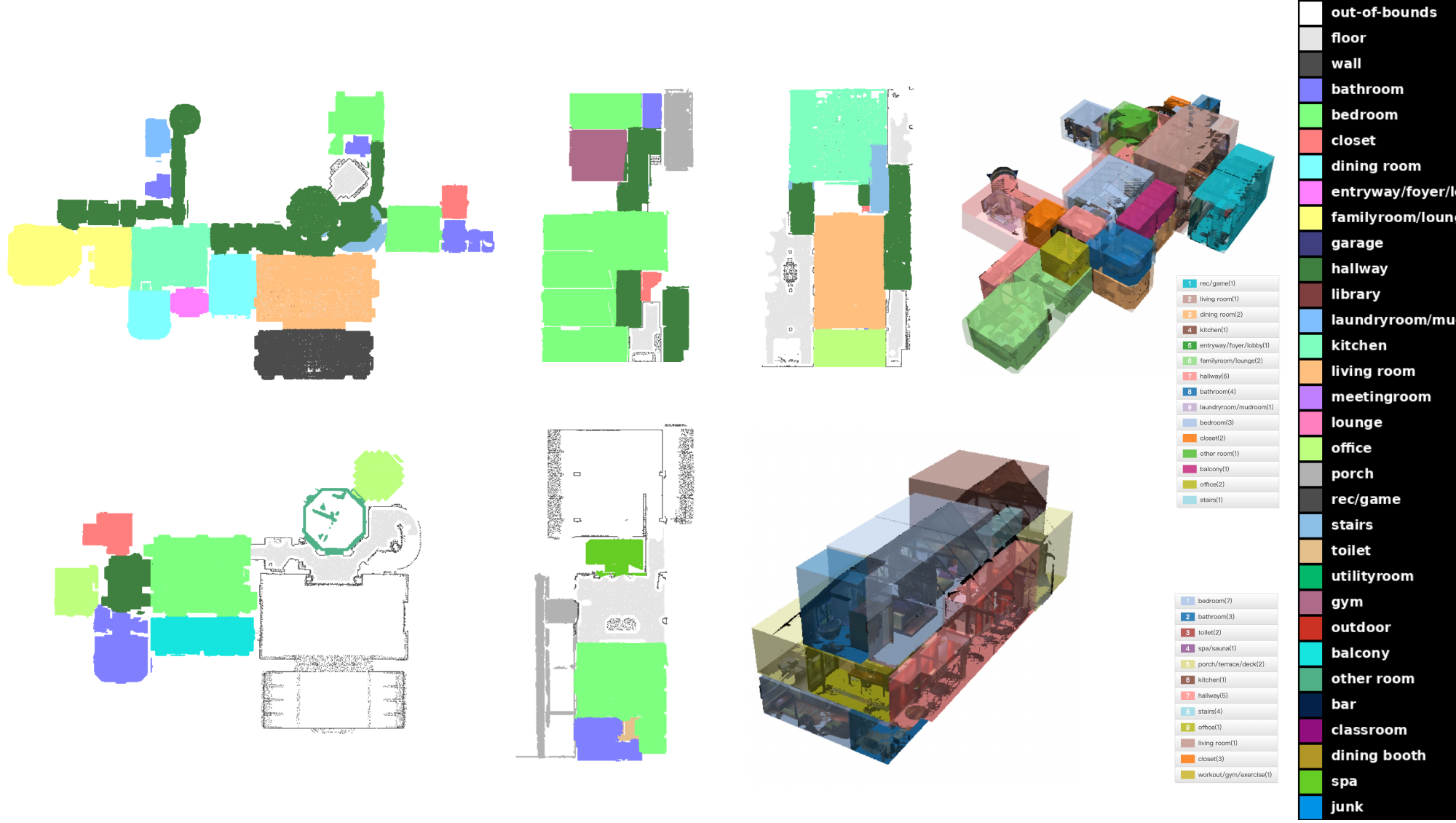}
	\caption{Room semantic segmentations derived from the Matterport3D dataset. The rightmost figures correspond to the left 3D house depiction, showing the ground truth bounding boxes for each room. The room categories in the Matterport3D dataset include: \texttt{"bathroom", "bedroom", "closet", "dining room", "entryway", "family room", "garage", "hallway", "library", "laundry room", "kitchen", "living room", "meeting room", "lounge", "office", "porch", "recroom", "stairs", "toilet", "utility room", "gym", "outdoor", "other-room", "bar", "classroom", "dining booth", "spa", "junk"}.}
	\label{FIG:matterpord3d_room_semantic}
\end{figure*}

The room annotations provided in these two datasets differ, but the general pre-processing step involves extracting face IDs based on the three vertex numbers in the mesh file, to which room annotations are assigned as different materials. Using the extracted face numbers and key house parameters such as the number of floors and floor height, we can calculate the 3D bounding box of each room. With this bounding box information, we can project the original house point cloud into a 2D floor map for each floor, incorporating the room segmentation as shown in Fig. \ref{FIG:room_seg}. More examples of room semantic segmentation based on the Gibson and Matterport3D datasets can be found in the Appendix, as shown in Fig. \ref{FIG:gibson_room_semantic}, \ref{FIG:matterpord3d_room_semantic}.

\begin{figure*}[ht]
    \centering
    \begin{subfigure}{0.8\textwidth}
        \includegraphics[width=\textwidth]{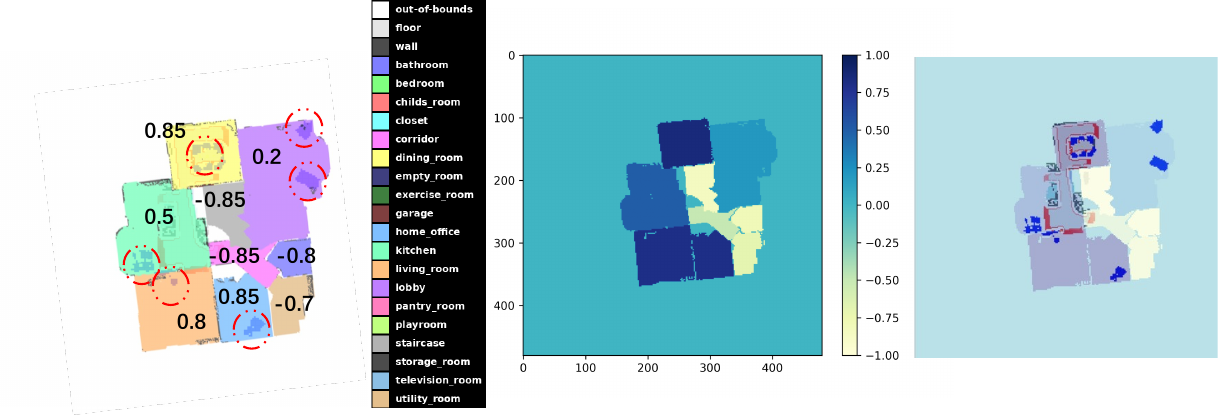}
        \caption{LLM-based chair-with-room maps}
    \end{subfigure}
    \hspace{0.1\textwidth}
    \begin{subfigure}{0.8\textwidth}
        \includegraphics[width=\textwidth]{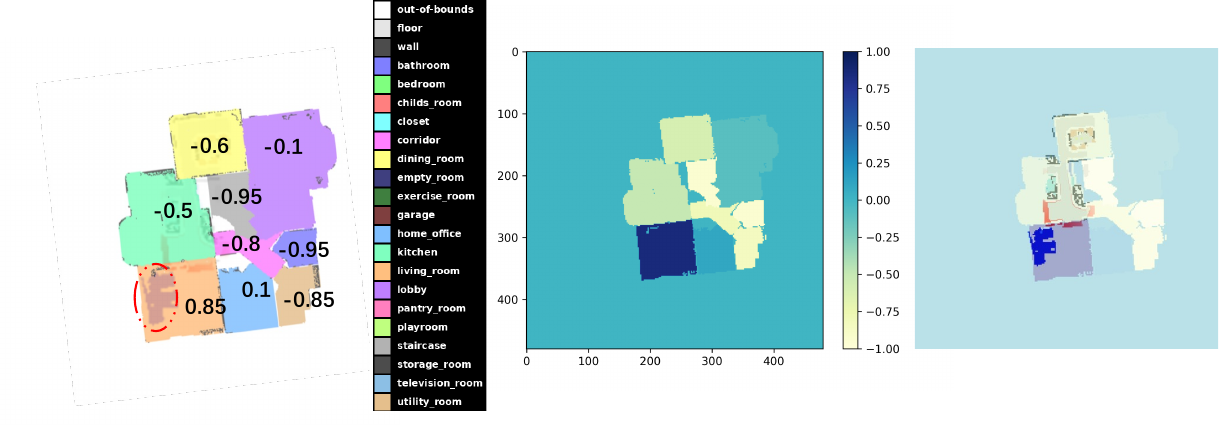}
        \caption{LLM-based couch-with-room maps}
    \end{subfigure}
    \caption{LLM-based O2R relationship map for chairs and couches. From left to right: (a.1) room segmentation overlapping with the ground truth chair/couch position marked by a red circle and the LLM-based score; (a.2) LLM-based O2R score map; (a.3) LLM-based O2R score map overlapped with the ground truth chair/couch position (same positions as shown in the red circle of (1)), highlighted in blue.}
    \label{FIG:room_object_score_map}
\end{figure*}

For each object category $o_{i}$, the LLM-based O2R knowledge score $h_{r_{i},o_{i}}$ is assigned to each room $r_{i}$. The O2R relationship map $M_{r,o}\in\mathbb{R}^{H \times W \times 15}$, where the channel number 15 represents the object categories detectable by Mask R-CNN \cite{he2017mask}. Fig. \ref{FIG:room_object_score_map} illustrates the injection of LLM-based knowledge into the O2R dataset for two object categories: \texttt{"chair", "couch"}. As this knowledge is derived from the common sense understanding of LLM, we lack ground truth labels for it. Therefore, we visualize the LLM-based O2R knowledge score $h_{r_{i},o_{i}}$, overlapping it with room semantic labels and object ground truth positions in Fig. \ref{FIG:room_object_score_map}.(a/b.1). Through this visualization, we can observe that high/positive scores are assigned to rooms where the object is likely to occur, such as \texttt{chairs} in the \texttt{dining room} and \texttt{home office} with a score of 0.85. Conversely, low/negative scores are indicated in rooms/areas where the object's appearance is rare. For example, a \texttt{couch} in the \texttt{kitchen} and on the \texttt{staircase} are assigned scores of -0.5 and -0.95, respectively. This concept is similarly demonstrated in Fig. \ref{FIG:room_object_score_map}.(a/b.3), where darker colors (high/positive scores) appear at the object's ground truth positions (shown in blue). In contrast, most rooms that do not contain the object are represented with lighter colors (low/negative scores). More examples are shown in the Appendix, Fig. \ref{FIG:obj_room_wholemap} and \ref{FIG:obj_room_gt_vis}.

\subsection{Dataset augmentation and ground truth calculation}
\label{sec: grondtruth}

Similar to the proposed LLM-based O2R relationship dataset that predicts frontiers with LLM-based likelihood scores of object-to-room relationships, our primary task involves predicting the frontier with scores representing the distance to the target object. An additional auxiliary task is to predict the frontier with scores indicating unexplored areas. To train these two tasks, we utilize the dataset generation process from PONI \cite{ramakrishnan2022poni}, an efficient non-interactive, frontier-based approach. In this subsection, we demonstrate how to generate ground truth frontier-based $m_{p}^{o}, m_{p}^{a}$ and $m_{p}^{r}$ for training the three tasks in the proposed LROGNav.

\begin{figure}[ht]
	\centering
		\includegraphics[scale=.5]{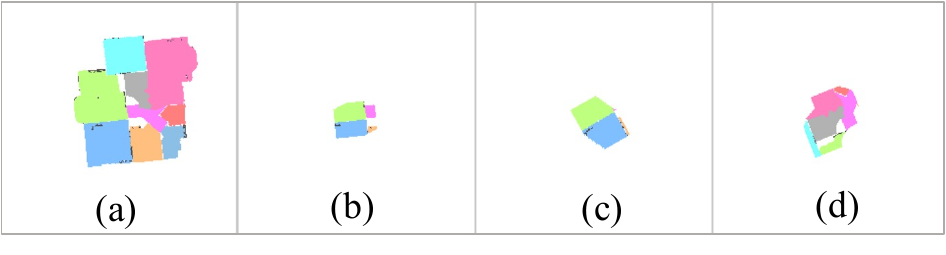}
	\caption{(a) complete whole map $m_{c}^{r}$ with room annotations; (b/c/d) partial map $m_{p}^{r}$ extracted from (a) by deploying random translation and rotation with room annotations.}
	\label{FIG:partial_map}
\end{figure}

Assuming we have a complete semantic object map $m_{c}^{o}$, we apply random translation and rotation as data augmentation to generate sufficient training data for a data-hungry Transformer-based approach. We then randomly select two locations in the navigable space of the complete map and compute the shortest path using the deterministic Fast Marching Method (FMM) path planner \cite{sethian1999fast}. For each location along this path, we sample a $S (30cm) \times S$ square patch around it to denote the area that the agent has explored. These patches are collected and combined to form the partial map $m_{p}$ for three tasks, such as the partial room semantic segmentation map $m_{c}^{r}$ shown in Fig. \ref{FIG:partial_map}.

\begin{figure}[ht]
	\centering
		\includegraphics[scale=.46]{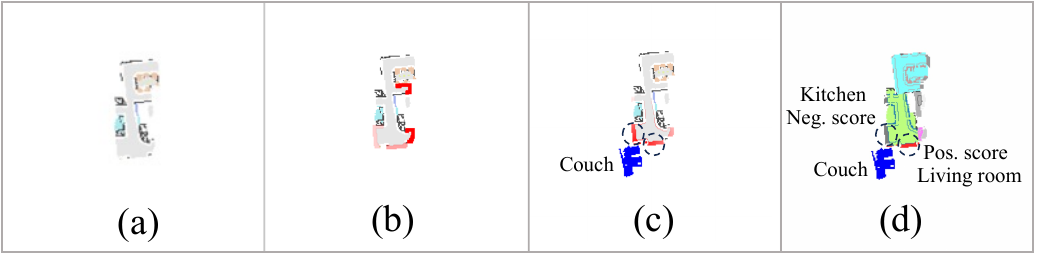}
	\caption{(a) partial map $m_{p}$ as one of input modalities; (b) ground truth area potential map $\hat{m}_{p}^{a}$; (c) ground truth object potential map $\hat{m}_{p}^{o}$ with couch ground truth position, highlighted in blue; (d) ground truth O2R potential map $\hat{m}_{p}^{r}$, positive score in red, negative score in gray.}
	\label{FIG:all_PF_map}
\end{figure}

The frontiers are the edges located between explored and unexplored areas in non-obstacle spaces. We calculate the connected components and associate them with a frontier only when the components are 4-connected neighbors of some pixels within the frontier. We then sum the components of each frontier in $m_{p}^{a}$ and normalize this by the total navigable area in $m{c}$. In summary, the score of frontiers in $m_{p}^{a}$ represents the unexplored space area in that direction, as shown in Fig. \ref{FIG:all_PF_map} (b). The partial map $m_{p}$ is extracted from the relative left space in the complete map $m_{c}$, resulting in more unexplored space on the right than the left. This is represented in $m_{p}^{a}$, where the frontier on the right side has a higher value than the frontiers on the left (in red color).

For calculating the scores in the object potential map $m_{p}^{o}$, as shown in Fig. \ref{FIG:all_PF_map} (c), the distance from a frontier position $x$ to the target object $o_{t}^{g}$ is calculated as follows:
\begin{equation}
d_{(o_{g},x)} = \max(1-\frac{1-d_{g}(o_{g},x)}{d_{max}},0)
\label{eq:object_distance}
\end{equation}
where $d_{g}$ represents the geodesic distance between the frontier location and the success circle (diameter = 1m) around the nearest target object $o_{g}$, and $d_{max}$ is the distance at which the score $d_{(o_{g},x)}$ decays to 0. This parameter should be tuned experimentally. In the Gibson dataset, it is set to 5m, aligning with the configuration of PONI \cite{ramakrishnan2022poni}. As illustrated in Fig. \ref{FIG:all_PF_map} (c), the left and bottom, two frontiers are similarly close to the target object, the couch (highlighted in blue), resulting in these two frontiers having higher scores compared to the rightmost one, which is farther away from the couch. 

\begin{figure*}
	\centering
		\includegraphics[scale=.85]{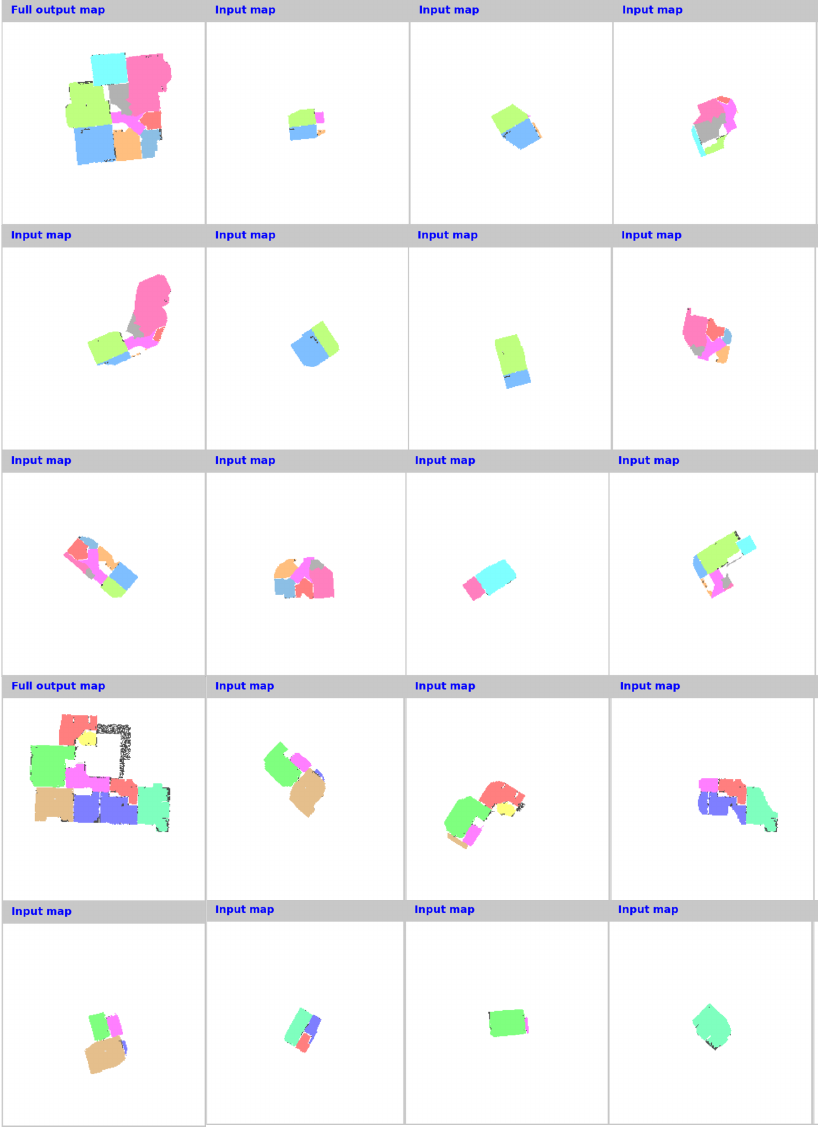}
	\caption{Data augmentation of room semantic segmentation map, two complete room semantic map $m_{c}^{r}$ are the first figure of the first and third rows, the rest figures are partial maps extracted from them.}
	\label{FIG:data_aug}
\end{figure*}

\begin{figure*}
	\centering
		\includegraphics[scale=0.85]{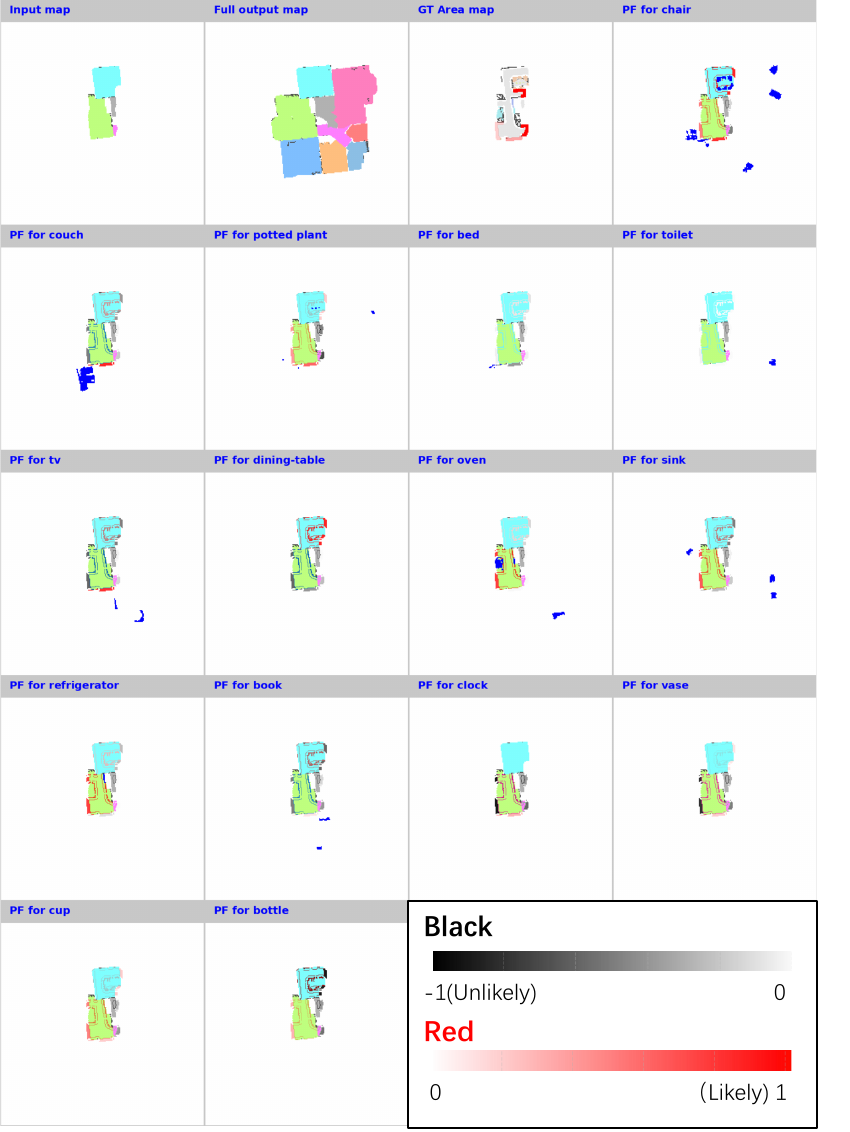}
	\caption{Ground truth O2R potential maps $\hat{m}_{p}^{r}$, the first map in the first row is a partial map extracted from a complete map, as shown in the second map. The third map is the ground truth area potential map $\hat{m}_{p}^{a}$. The subsequent 15 figures represent the O2R potential maps $\hat{m}_{p}^{r}$ for 15 object categories, the object position is highlighted in blue. The scores on the frontiers, based on the LLM, are color-coded: black indicates a low likelihood of finding the object in that room, while red suggests a high possibility of the object appearing in the room where this frontier is located. }
	\label{FIG:all_obj_room_PF}
\end{figure*}

To generate the proposed LLM-based O2R relationship map $m_{p}^{r}$ , we assign the scores $h_{r,o}$ extracted from the relationship matrix based on the object category and room type. As illustrated in Fig. \ref{FIG:all_PF_map} (d), unlike $m_{p}^{o}$, the left frontier has a lower score than the bottom one because it is located in a kitchen, whereas the bottom frontier is in a living room. It is evident that the likelihood of finding a couch in the living room is higher than in the kitchen. However, this aspect is overlooked in the $m_{p}^{o}$ from PONI \cite{ramakrishnan2022poni}, where the supervisory signal in $m_{p}^{o}$ is based solely on distance. In summary, the proposed LROGNav regresses three frontier prediction tasks, as demonstrated in Fig. \ref{FIG:all_PF_map} (b/c/d). The dataset $D$ we collected for training the proposed LROGNav is described below:
\begin{equation}
D = \left \{ [\hat{m}_{c}^{o}, \hat{m}_{c}^{a}, \hat{m}_{c}^{r}] \overset{\text{aug.}}{\Rightarrow} (\hat{m}_{p_{1}}^{o_{i}}, \hat{m}_{p_{1}}^{a}, \hat{m}_{p_{1}}^{r_{i}}), (\hat{m}_{p_{2}}^{o_{i}}, \hat{m}_{p_{2}}^{a}, \hat{m}_{p_{2}}^{r_{i}}), \dots \right \} 
\label{eq:dataset}
\end{equation}

where $m_{c}$ and $m_{p}$ denote the complete and partial maps, respectively, while $o$/$a$/$r$ stand for object, area, and room. The terms with and without the hat symbol ( $\hat{}$ ) represent the ground truth and prediction in partial maps, respectively. It is noteworthy that all the ground truth labels for training are only for the pixels at frontiers, the other semantic information is just for visualization. Additionally, $i$ denotes the different object categories. More dataset samples are illustrated in Appendix, Fig. \ref{FIG:obj_room_partialmap}, \ref{FIG:obj_room_gt_vis}, \ref{FIG:data_aug}, \ref{FIG:all_obj_room_PF}.

\subsection{Network architecture}
\label{sec:network}
\subsubsection{Multi-modal input}
We utilize a multimodal-input strategy to enhance the performance of the proposed LROGNav, which employs a Swin Transformer-based \cite{liu2021swin} encoder-decoder network architecture. To regress the primary task $U^{o}$ and two auxiliary tasks $U^{a}$ and $U^{r}$, we encode object semantic information from RGB $I_{t}^{v}$ and depth $I_{t}^{d}$ observations at each timestamp $t$, integrating it with the pose $(x_{t},y_{t},\theta_{t})$ to construct a semantic projection map $M_{t}^{s} \in\mathbb{R}^{\hat{h}\times\hat{w}\times17}$. The first two layers of initial map represent the explored area and obstacle map, while the remaining layers depict the positions of different object categories collected during navigation, following the approach used in SemExp \cite{semexp}, PONI \cite{ramakrishnan2022poni}, PEANUT \cite{zhai2023peanut}, and FSE \cite{FSE}. 

\begin{figure}[!t]
	\centering
		\includegraphics[scale=.6]{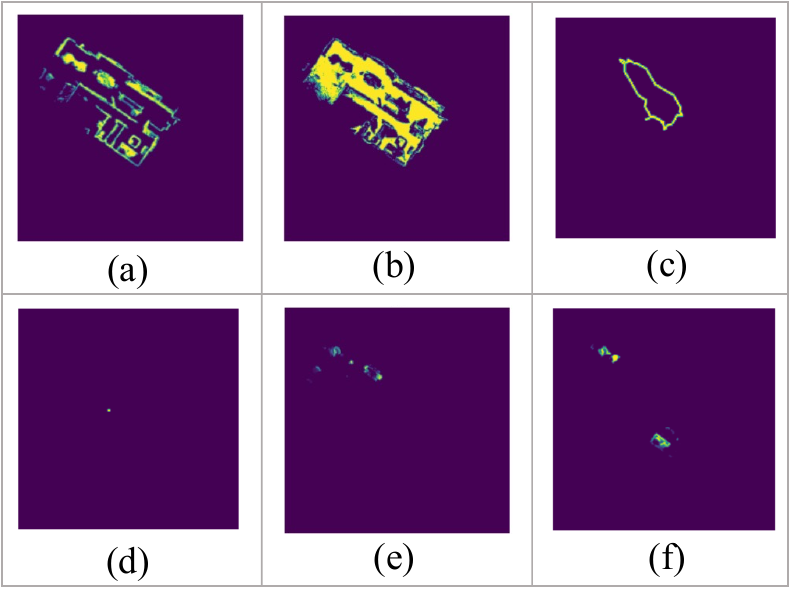}
	\caption{Some layers visualization of input to multi-channel Swin-Unet and position-related information. (a) obstacle map; (b) explored area; (c) history trajectory; (d) agent current position; (e/f) '\texttt{chair, couch}' positions}
	\label{FIG:input_map}
\end{figure}

\begin{figure}[ht]
	\centering
		\includegraphics[scale=.39]{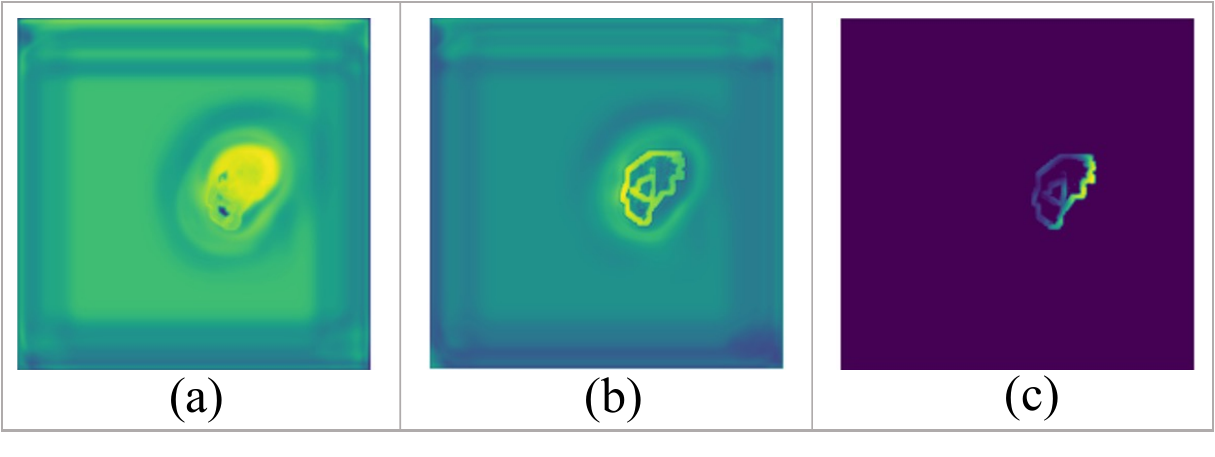}
	\caption{Frontier predictions at (a) initial, (b) intermediate and (c) final periods in LROGNav.}
	\label{FIG:frontier}
\end{figure}

Each channel has a different semantic and independent meaning in $M_{t}^{s}$, as illustrated in Fig. \ref{FIG:input_map}. Furthermore, the predicted frontier of decoders in LROGNav, as shown in Fig. \ref{FIG:frontier}, requires reasoning across the different input channels, especially the obstacle and explored area maps. To enhance this multi-channel reasoning, we employ a learnable channel-wise positional embedding (PE), a concept similar to those proposed in ChannelViT \cite{bao2023channelvit} and SAT3D \cite{ibrahim2023sat3d}. This channel-wise PE in Eq. \ref{eq:channel-wise} is applied after the patch embedding $\tilde{X}_{s}=[x_{t}^{s,ij}]_{\hat{h} \times \hat{w}}$ with $\hat{h} \times \hat{w}$ cells. Each cell $(i, j)$ contains a latent feature $x_{t}^{s,ij} \in \mathbb{R}^{c}$, where $c$ represents the embedded feature dimensionality before feeding into the Swin Transformer. More details of channel-wise PE are provided as pseudocode in Algorithm \ref{al:channel-wise code}.

\begin{equation}
\mathbf{X}_{s} = [x^{1,1}_{t}, x^{1,2}_{t}, ..., x^{h,w}_{t}] + \mathbf{x}_{channel\_PE}
\label{eq:channel-wise}
\end{equation}

\begin{algorithm}[!t]
\caption{Channel-wise Position Embedding}
\label{al:channel-wise code}
\begin{algorithmic}[1]
\Require Patch embeddings $X$ from semantic projection \\ $X$ $\in \mathbb{R}^{batch\_size \times num\_patches \times embed\_dim}$
\Ensure Patch embeddings with channel position encoding
\Function{Add\_Channel\_Position\_Embedding}{$X$}
    \State $in\_channels \gets 17$ \Comment{Original number of channels}
    \State $embedding\_size $ \Comment{Size after linear layer, to align with other CLIP embeddings}
    \State Initialize $ChannelPosOriginal \in \mathbb{R}^{1 \times in\_channels}$ with values $0$ to $16$
    \State $LinearExpand \gets \text{LinearLayer}(in\_channels, embedding\_size)$
    \State $ChannelPosExpanded \gets LinearExpand(ChannelPosOriginal)$
    \State $X \gets X + ChannelPosExpanded[:, \text{None}, :]$
    \State \Return $X$
\EndFunction

\end{algorithmic}
\end{algorithm}

Until now, these input modalities have not included room-based knowledge, which is essential for the proposed auxiliary task $U^{r}$ and crucial for effective multi-task network convergence. To address this, we integrate room information using a CLIP-based contrastive model \cite{clip} during inference, employing a frozen "ViT-B/32" pre-trained model to estimate the current potential room type $r_{t}^{v_{i}}$ from RGB observations, denoted as $r_{t}^{v_{i}}=CLIP(I_{t}^{v})$, where $i =3 $ represents the top 3 candidates, more details are formulated as pseudocode in Algorithm \ref{al:CLIP}. However, to preserve the advantage of the efficient non-interactive training pipeline proposed in PONI, the room category is determined by the center location represented in $\hat{m}_{p_{1}}^{r_{i}}$ (robot-centric) during the training process. Similarly, the potential target rooms $r_{t}^{LLM_{i}}$, extracted from the LLM-based O2R relation matrix (as introduced in Section \ref{sec:o2r relationship}), along with the target object category $o_{t}^{g}$, are encoded using the same frozen CLIP model. To summarize, all these textual modalities are embedded by the CLIP model, denoted as $[ {\frac{1}{i} \textstyle \sum_{1}^{i}}C(r_{t}^{v_{i}})p_{r}^{i}, {\frac{1}{i} \textstyle \sum_{1}^{i}} C(r_{t}^{LLM_{i}})p_{LLM}^{i}, C(o_{t}^{g})]$, where the $p_{r}^{i}$ and $p_{LLM}^{i}$ represent the confidence score of CLIP/LLM-based estimation and then subsequently concatenated for integration into the multi-channel Swin-Unet.

\begin{algorithm} [!t]
\caption{Estimating Room Types and Confidence Score Using a Pre-trained CLIP Model}
\label{al:CLIP}
\begin{algorithmic}[1]
\Require $image$ \Comment{The input RGB observation $I_{t}^{v}$}
\Require $model$ \Comment{Pre-trained CLIP model}
\Ensure $predictions$ \Comment{Top 3 predicted room types $r_{t}^{v_{i}}$ and their confidence scores $p_{r}^{i}$}

\State $room\_types \gets$ List of defined room types
\State $predictions \gets []$ \Comment{Initialize list for predictions}

\Function{PredictRoomTypes}{$image, model, room\_types, device$}
    \State $image\_features \gets model.encode\_image(image)$ \Comment{Encode the image with CLIP}
    \State $text\_tokens \gets clip.tokenize(room\_types)$
    \State $text\_features \gets model.encode\_text(text\_tokens)$
    \State $similarities \gets image\_features @ text\_features.T$ \Comment{Compute similarities}
    \State $top\_probs, top\_labels \gets similarities.softmax(dim=-1).topk(3)$ \Comment{Extract top 3 predictions}
    \For{$index$ in $top\_labels[0]$}
        \State $predicted\_room \gets room\_types[index]$ \Comment{Map indices to room types $r_{t}^{v_{i}}$}
        \State $score \gets top\_probs[0][index].item()$ \Comment{Extract confidence scores $p_{r}^{i}$}
        \State $predictions.append((predicted\_room, score))$ 
    \EndFor
    \State \Return $predictions$
\EndFunction
\end{algorithmic}
\end{algorithm}

The distance and direction to target object are significant reward signals for training an end-to-end reinforcement learning ObjectNav policy \cite{al2022zero}. Since our proposed LROGNav is a frontier-based navigation approach \cite{ramakrishnan2022poni} \cite{FSE} \cite{yamauchi1997frontier}, the distance to the nearest target object $d_{t}^{o_{i}}$ and direction $\theta_{t}^{o_{i}}$ are also crucial for predicting the frontier shape and values ($m_{p}^{o}$), serving as geometrical constraint information. During dataset generation, we assume all partial maps $m_{p}$ to be robot-centric, meaning the robot is at the center of the map. We then select the nearest target object of each category by identifying the connected components in the map. The details of this computation are outlined in the pseudocode Algorithm \ref{al:direction and location}. We define east as 0 and consider clockwise as positive for direction. The output distance $d_{t}^{o_{i}} \in \mathbb{R}^{1 \times 15}$ and direction $\theta_{t}^{o_{i}} \in \mathbb{R}^{1 \times 15}$ are projected through a multilayer perceptron (MLP)-based linear embedding (denotes as $L(\cdot)$ in Eq. \ref{eq:modal-type}) to align the dimensions and then concatenated with three other word embeddings.

\begin{algorithm}[!t]
\caption{Compute Directions and Locations of each object}
\label{al:direction and location}
 \hspace*{\algorithmicindent} \textbf{Input}: $sem\_map$ -- a 3D multi-layer semantic maps \\
 \hspace*{\algorithmicindent} \textbf{Output}: $out\_dirs$ -- directions to closest object, $out\_locs$ -- normalized coordinates of closest object 
\begin{algorithmic}[1]
\For{each object layer in $sem\_map$}
\State $H, W \gets$ dimensions of $sem\_map$
\State $H_{1/2}, W_{1/2} \gets H/2, W/2$
\State $centroids \gets$ connectedComponents($sem\_map$)
\If{$centroids = \emptyset$}
\State append default values to $out\_dirs$, $out\_locs$
\State \textbf{continue}
\EndIf
\State $min\_idx \gets$ argmin$_i$ distance($centroids[i]$, $(H_{1/2}, W_{1/2})$)
\State $obj_y, obj_x \gets centroids[min\_idx]$
\State $dists \gets \sqrt{(obj_y - H_{1/2})^2 + (obj_x - W_{1/2})^2}$
\State $obj\_dir \gets$ arctan($obj_y - H_{1/2}$, $obj_x - W_{1/2}$)
\State append $obj\_dir$ to $out\_dirs$
\State append $dists$ to $out\_dist$
\EndFor
\State \Return $out\_dirs, out\_locs$
\end{algorithmic}
\end{algorithm}

Similar to other transformer-based multi-modal fusion frameworks, such as ViLT \cite{kim2021vilt}, AFT-VO \cite{kaygusuz2022aft}, and TransFusionOdom \cite{sun2023transfusionodom}, a learnable modal-type embedding is added to the tokens to differentiate each modality source before performing the attention mechanism calculation: 

\begin{equation}
\begin{split}
\mathbf{X}_{t} = [& {\frac{1}{i} \textstyle \sum_{1}^{i}}(C(r_{t}^{v_{i}})\times p_{r}^{i}), {\frac{1}{i} \textstyle \sum_{1}^{i}}(C(r_{t}^{LLM_{i}}) \times p_{LLM}^{i}), \\ & C(o_{t}^{g}), 
L(d_{t}^{o_{i}}), L(\theta_{t}^{o_{i}}) ] + \mathbf{x}_{\text{modal\_type}}
\end{split}\label{eq:modal-type}
\end{equation}
where ${x}_{modal\_type} \in \mathbb{R}^{c}$ is the learnable modal-type embedding, having the same dimensions as the patch embedding of the semantic projection map $\mathbf{X}_{s}$ in Eq. \ref{eq:channel-wise}. These multi-modal embeddings, denoted as $[\mathbf{X}_{s}, \mathbf{X}_{t}]$ are fed into a fully-connected linear layer ($C_{emb} = 128$ to align with the pre-trained Swin-B Transformer model) and then undergo a partitioning process into the multi-channel Swin-Unet.

\subsubsection{Multi-channel Swin-Unet}

\begin{figure*}[t]
	\centering
		\includegraphics[scale=.5]{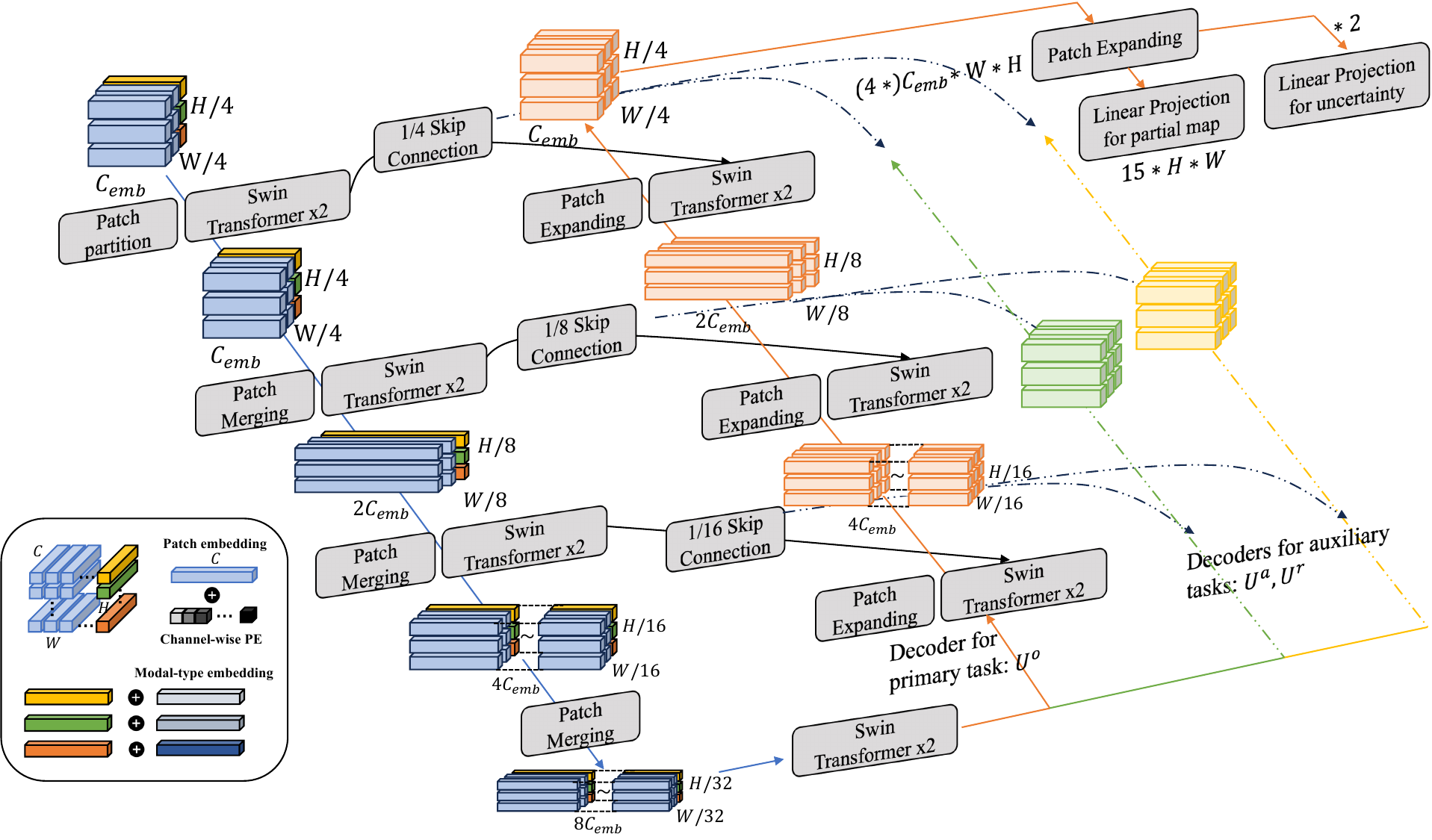}
	\caption{The network architecture of proposed LROGNav. Input embeddings are processed through channel-wise positional encoding (PE) and summed with the modal-type embeddings, then feed into the encoder and three decoders. These components work to predict three different maps and compute the loss by comparing them with their respective ground truth maps generated in Section \ref{sec: grondtruth}.}
	\label{FIG:network}
\end{figure*}

The network architecture of the proposed LROGNav is illustrated in Fig. \ref{FIG:network}. We utilize a Swin Unet-based \cite{cao2022swinunet} multi-channel encoder to encode the multimodal input, which has been concatenated in the previous processes. These patch tokens pass through several Swin Transformer \cite{liu2021swin} blocks and patch merging layers, producing hierarchical feature representations. The patch merging layers are used for downsampling and increasing dimensionality, while the Swin Transformer blocks facilitate feature representation learning. Inspired by the UNet architecture \cite{ronneberger2015unet}, three symmetrical Transformer-based decoders are designed to predict three maps ($m_{p}^{o}$, $m_{p}^{a}$, and $m_{p}^{r}$), each consisting of Swin Transformer blocks and patch expanding layers. The contextual features extracted are fused with the encoder's multi-scale features via skip connections, compensating for spatial information loss due to downsampling. The last patch expanding layer, along with two MLP-based linear projections, regress the outputs as $m_{p}^{o_{i}} \in \mathbb{R}^{15 \times H \times W}$, along with its corresponding adaptive weight for fusion \cite{sun2023transfusionodom} \cite{certianodom} \cite{whatuncertainty} \cite{balancedvio}, denoted as $\sigma_{p/s}^{o_{i}}$. The same decoder structure are deployed for two auxiliary tasks $U^{a}$ and $U^{r}$.

To efficiently handle multi-channel input data, the patch partition is embedded using a separable convolution layer proposed in MobileNet \cite{sinha2019thin}, which includes depthwise and pointwise convolutions. Conversely, the original Swin-Unet is implemented with non-separable convolution, as it is primarily used for grayscale and RGB images \cite{faulkenberry2024visual}. In each pair of successive Swin Transformer block units, there are primarily two modules in sequence: first, a window-based multi-head self-attention (W-MSA) module, followed by a shifted window-based multi-head self-attention (SW-MSA) module. This sequence can be formulated as follows:
\begin{equation}
\hat{\mathbf{x}}^l = W\text{-}MSA(LN(\mathbf{x}^{l-1})) + \mathbf{x}^{l-1},
\label{eq:swinblock1}
\end{equation}

\begin{equation}
\mathbf{x}^l = MLP(LN(\hat{\mathbf{x}}^l)) + \hat{\mathbf{x}}^l,
\label{eq:swinblock2}
\end{equation}

\begin{equation}
\hat{\mathbf{x}}^{l+1} = SW\text{-}MSA(LN(\mathbf{x}^l)) + \mathbf{x}^l,
\label{eq:swinblock3}
\end{equation}

\begin{equation}
\mathbf{x}^{l+1} = MLP(LN(\hat{\mathbf{x}}^{l+1})) + \hat{\mathbf{x}}^{l+1},
\label{eq:swinblock4}
\end{equation}
where $LN(\cdot)$ represents the LayerNorm layer, $\hat{\mathbf{x}}^l$ and $\mathbf{x}^{l}$ denote the outputs of the (S)W-MSA and the MLP layer, respectively, in the $l^{th}$ block, where $l \in [1,14]$. Additionally, the self-attention module is calculated as outlined in previous works \cite{cao2022swinunet} \cite{hu2018relation} \cite{hu2019local}, as follows:

\begin{equation}
\text{Attention}(\mathbf{Q}, \mathbf{K}, \mathbf{V}) = \text{SoftMax}\left(\frac{\mathbf{QK}^T}{\sqrt{d}} + \mathbf{B}\right)\mathbf{V}
\label{eq:swinblock6}
\end{equation}
consider the set of matrices $ \mathbf{Q}, \mathbf{K}, \mathbf{V} \in \mathbb{R}^{N \times d_k} $, where they symbolize the query, key, and value constructs, respectively. Here, \( N \) represents the count of discrete regions within a specified window, while \( d_k \) denotes the dimensionality of the query or key vectors. Furthermore, entries within matrix \( B \) are sourced from a predefined bias matrix \cite{liu2021swin} \( \mathbf{B} \in \mathbb{R}^{(2N-1) \times (2N+1)} \). In the ablation study Section \ref{sec:ablation}, the absolute, relative \cite{liu2021swin}, and channel-wise position embeddings are experimentally evaluated.

\subsubsection{Multi-task regression and loss functions}
In the original PONI framework \cite{ramakrishnan2022poni}, the image-to-image model is trained using a UNet with Mean Square Error (MSE) for the primary task $U^{o}$ and the auxiliary task $U^{a}$, which are then directly combined to form the final loss. Inspired by various depth estimation works \cite{bian2019unsupervised} \cite{song2023spatial} that also tackle image reconstruction tasks, we adopt photometric loss (MSE-based) $L_{p}$ and Structural Similarity Index Measure (SSIM) loss $L_{s}$ to train our supervised frontier learning network for each task. Different from PONI, which uses fixed weights or manually-tuned hyperparameters \cite{wu2023self} \cite{liu2023unsupervised} to combine the losses from two tasks, we employ uncertainty-based adaptive weights \cite{whatuncertainty} to fuse the losses between different loss terms among different tasks \cite{certianodom} \cite{iwaszczuk2021deeplio}.

In exploring aleatoric uncertainty through self-supervised learning, the authors introduce a model that utilizes a specific loss function for optimizing learnable uncertainty, as described below \cite{whatuncertainty}:
\begin{equation}
\mathcal{L}_{\text{uncer.}} = \frac{1}{N} \sum_{i=1}^{N} \left( \frac{1}{2\sigma(x_{i})^2} \left | y_{i} - f(x_{i}) \right |^2 + \frac{1}{2}\log\sigma(x_{i})^2 \right)
\label{eq:uncertainty}
\end{equation}
where $\sigma$ represents the estimated uncertainty for the input $x$. The terms $y_{i}$ and $f(x_{i})$ correspond to the ground truth and the model's prediction for the input $x_{i}$, respectively.

Moreover, the depth image reconstruction task \cite{bian2019unsupervised} \cite{song2023spatial} often utilizes photometric loss combined with the SSIM function to compare the predicted depth with ground truth, as shown below:
\begin{equation}
\begin{split}
\mathcal{L}_{\text{pho\_ssim}} = \frac{1}{|V|} \sum_{p \in V} \Bigl( \lambda_i &\left\lVert D(p) - D'(p) \right\rVert_1 \\
&+ \lambda_s \frac{1 - \text{SSIM}_{dd'}(p)}{2} \Bigr)
\end{split}
\label{eq:pho_ssim}
\end{equation}
where $V$ stands for the valid pixels, $SSIM_{dd'}$ denotes the element-wise similarity calculation between the predicted depth $D(p)$ and ground truth $D'(p)$, and $\lambda_{i/s}$ are manually-tuned hyperparameters which require extensive experimentation and are sensitive to different datasets.

By integrating the uncertainty regression loss (see Eq. \ref{eq:uncertainty}) with the image reconstruction loss (see Eq. \ref{eq:pho_ssim}), which has been utilized to enhance the robustness and accuracy of regression models as detailed in \cite{poggi2020uncertainty}, \cite{nie2021uncertainty}, and \cite{song2023spatial}, we propose the use of learnable uncertainty as an adaptive weight. This approach aims to effectively combine the photometric loss and SSIM loss across three tasks. A similar concept has been previously verified in our previous work \cite{certianodom}. For example, the formulation of the two losses for the primary task is as follows:

\begin{equation}
\begin{split}
\mathcal{L}_{\text{pho}} (\theta_{p}^{o}, \sigma_{p}^{o}) = \frac{1}{N} \sum_{i=1}^{N} \bigg( \frac{1}{\sigma_{p}^{o}(x_{i})} \left\lVert m_{p}^{o}(x_{i}) - \hat{m}^{o}_{p}(x_{i}) \right\rVert_1 \\
+ \log \sigma_{p}^{o}(x_{i}) \bigg)
\end{split}
\label{eq:pho}
\end{equation}
where N stands for the number of pixels, $m_{p}^{o}(x_{i})$ represents the prediction of pixel $x_{i}$ by the object potential decoder $U^{o}$, which is supervised by the ground truth $\hat{m}^{o}_{p}(x_{i})$. Moreover, $\sigma_{p}^{o}$ represents the uncertainty in photometric regression for the primary object potential task, and $\theta_{p}^{o}$ denotes the weights of the network for the photometric loss in this primary task.

\begin{equation}
\begin{split}
\mathcal{L}_{\text{ssim}} (\theta_{s}^{o}, \sigma_{s}^{o}) &= \frac{1}{N} \sum_{i=1}^{N} \Bigg( \frac{1}{\sigma_{s}^{o}(x_{i})} \left\lVert \frac{1 - \text{SSIM}\big(m_{p}^{o}(x_{i}), \hat{m}^{o}_{p}(x_{i}) \big)}{2} \right\rVert_1 \\
&\quad + \log \sigma_{s}^{o}(x_{i}) \Bigg)
\end{split}
\label{eq:ssim}
\end{equation}

where most variables have the same meaning as those in the previously defined photometric loss. $SSIM(\cdot)$ denotes the SSIM loss function, referencing \cite{wang2004image}, which compares the prediction with the ground truth. $\sigma_{s}^{o}$ represents the uncertainty in SSIM regression for the primary object potential task, and $\theta_{s}^{o}$ denotes the weights of the network for the SSIM loss in this primary task.


The final joint loss is defined as follows, incorporating both photometric and SSIM loss terms for the primary task $U^{o}$, as well as for the auxiliary tasks $U^{a}$ and $U^{r}$.

\begin{equation}
\begin{aligned}
\mathcal{L}(\theta,\sigma) = & \mathcal{L}_{\text{pho}} (\theta_{p}^{o}, \sigma_{p}^{o}) + \mathcal{L}_{\text{ssim}} (\theta_{s}^{o}, \sigma_{s}^{o}) \\
 + & \mathcal{L}_{\text{pho}} (\theta_{p}^{a}, \sigma_{p}^{a}) + \mathcal{L}_{\text{ssim}} (\theta_{s}^{a}, \sigma_{s}^{a}) \\
 + & \mathcal{L}_{\text{pho}} (\theta_{p}^{r}, \sigma_{p}^{r}) + \mathcal{L}_{\text{ssim}} (\theta_{s}^{r}, \sigma_{s}^{r})
\end{aligned}
\label{eq:total loss}
\end{equation}






\section{Experiment evaluation}
\subsection{Experiment setup}
We deploy the proposed LROGNav to the Gibson \cite{xia2018gibson} and Matterport3D (MP3D) \cite{chang2017matterport3d} datasets, employing the Habitat simulator \cite{ramrakhya2022habitat} for our experiments. Both datasets feature photorealistic 3D reconstructions of actual environments. For the experimental configurations, we adopt the settings from our baseline approach, PONI \cite{ramakrishnan2022poni}. Specifically, for the Gibson dataset, we utilize a subset known as Gibson Tiny, which includes 25 training scenes and 5 validation scenes, all equipped with relevant semantic annotations \cite{armeni20193dscene}. In the case of the MP3D dataset, we adhere to the standard division of data, which comprises 61 training scenes, 11 validation scenes, and 18 test scenes. 

In terms of specific datasets, for our Gibson-based experiments, we employ the ObjectNav dataset from SemExp \cite{semexp} and PONI \cite{ramakrishnan2022poni}, which focuses on six types of goals: "chair", "couch", "potted plant", "bed", "toilet", and "tv". For our MP3D-based studies, we utilize the Habitat ObjectNav dataset \cite{ramrakhya2022habitat}, which encompasses a broader set of 21 goal categories. Details of these categories are provided in the Appendix materials (see the category in Fig. \ref{FIG:matterpord3d_matrix}). These categories are also employed for training the multi-channel Swin-Unet within the LROGNav's network.

We utilize two main metrics in the evaluation of the ObjectNav task: Success Rate (SR) and Success weighted by Path Length (SPL) \cite{anderson2018evaluation}. SR is defined as $\frac{1}{N} {\textstyle \sum_{i=1}^{N}} S_{i}$, where $N$ represents the number of episodes (with 200 episodes used for testing in each scene), and $S_{i}$ is a binary indicator of success in episode $i$. The definition of SPL can be articulated as follows: $\frac{1}{N} \sum_{i=1}^{N} S_i \frac{L_i}{\max(P_i, L_i)}$, where $P_{i}$ and $L_{i}$ represent the actual and optimal path lengths, respectively, for each testing episode $i$. Essentially, SPL assesses the efficiency of the agent's trajectory in successfully reaching the goal compared to the best possible path for any given instance of the targeted object class in the scenario.

\subsection{Implementation details}

In our study of the detection and segmentation framework for the Gibson dataset, we deploy a refined Mask-RCNN model \cite{he2017mask}, originally trained on the COCO dataset \cite{lin2014microsoft}, as adapted from PONI \cite{ramakrishnan2022poni}, using images from the Gibson Tiny training split. This fine-tuning process incorporates 15 distinct object categories as detailed in SemExp \cite{semexp}. Moreover, for the MP3D dataset, our approach utilizes a RedNet segmentation model \cite{jiang2018rednet}, as trained and described in \cite{maksymets2021thda}, to identify and classify 21 different object categories, which are listed in the Appendix material.

\begin{figure}[t]
	\centering
		\includegraphics[scale=.4]{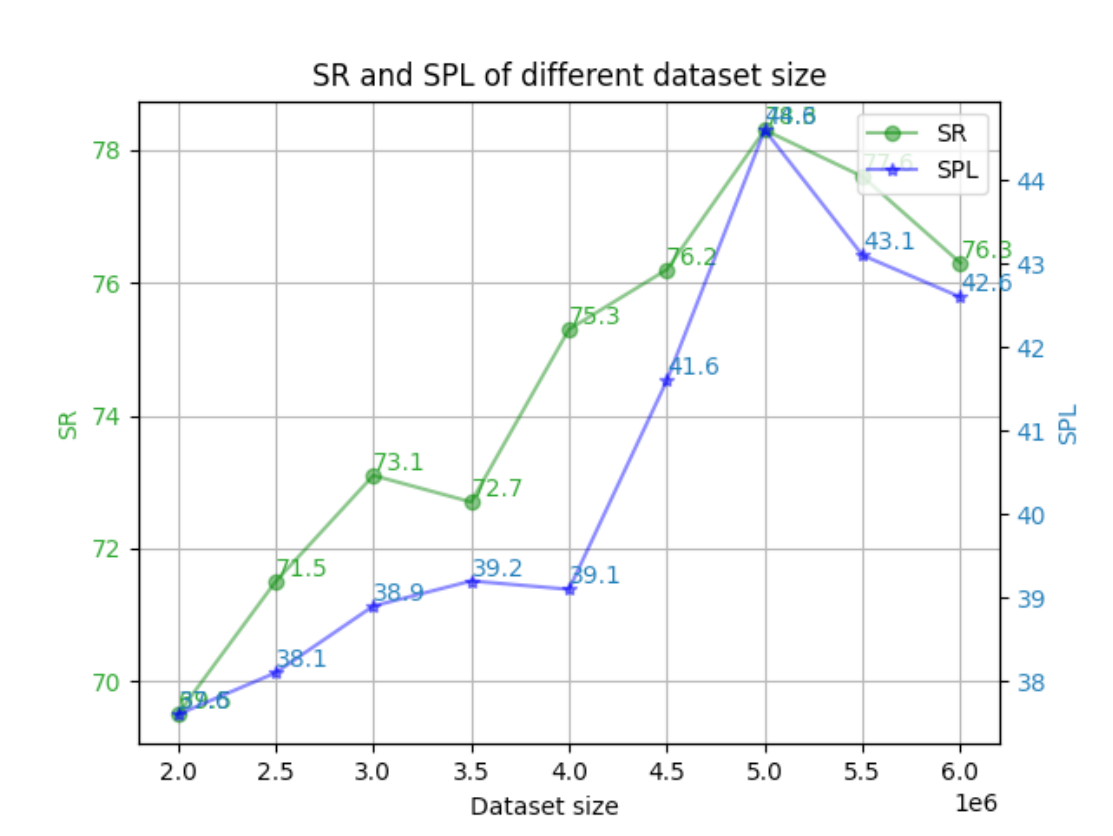}
	\caption{The SR and SPL for different size of generated Gibson-based training dataset.}
	\label{FIG:dataset size}
\end{figure}

The network for computing potential values on the frontiers is structured around a Swin-Unet \cite{cao2022swinunet} encoder-decoder design. This architecture is trained utilizing multi-modal inputs, as elaborated in Section \ref{sec:network}. For the Gibson dataset, we derive 63 training and 13 validation maps from every floor within the Gibson Tiny division. In the case of the MP3D dataset, we derive 108 training and 21 validation maps from each floor. Additionally, we generate and prepare 800,000 training and 2,000 validation combinations ($\hat{m}_{p}^{o}$, $\hat{m}_{p}^{a}$, $\hat{m}_{p}^{r}$) as outlined in Section \ref{sec: grondtruth}. We conduct exhaustive experiments with different dataset sizes as shown in Fig. \ref{FIG:dataset size} to determine the best performance in terms of SR and SPL. Compared with the approximately 25M parameters in the UNet-based \cite{ronneberger2015unet} PONI framework, the Swin-UNet-based LROGNav achieves the best SR/SPL with approximately 50M training data in the Gibson evaluation.

The training process takes within 2.5 days is conducted on 4 Tesla V100 with 16GB VRAM per GPU, over 30 epochs using the PyTorch framework \cite{paszke2019pytorch}, employing the Adam optimization algorithm \cite{kingma2014adam} with an initial learning rate of 0.001. This rate undergoes a decimation by a factor of 10 following the completion of every 10 epochs. 

\begin{table}[!t]
\caption{The performance of LROGNav across various pre-trained backbone models and input sizes.}
\renewcommand\arraystretch{1.9}
\begin{center}
\scalebox{0.78}{
\begin{tabular}{ccccc}
\hline
Backbone        & Resolution       & FPS         & SR            & SPL           \\ \hline
Swin-T          & 224*224          & 87          & 73.8          & 40.9          \\
Swin-S          & 224*224          & 52          & 74.6          & 41.2          \\
Swin-B          & 224*224          & 36          & 76.2          & 42.6          \\
\textbf{Swin-B} & \textbf{384*384} & \textbf{14} & \textbf{78.3} & \textbf{44.6} \\
Swin-L          & 224*224          & 17          & 77.6          & 44.2          \\
Swin-L          & 384*384          & 5           & 79.1          & 45.9          \\ \hline
\end{tabular}
}
\end{center}
\label{tab:fps}
\end{table}

In the process of transferring ObjectNav, the advantages of selecting a long-term goal every T=1 steps have been verified in PONI \cite{ramakrishnan2022poni}. However, when deploying the framework to real-world experiments, and considering the efficiency of inference balanced with the translation/rotation velocity for real robots (0.5m/s and 0.2rad/s), long-term goals are selected at intervals of $T=25$ steps, aligning with earlier modular-based methodologies \cite{semexp} \cite{chaplot2020learning} \cite{ramakrishnan2020occupancy}.

We evaluate the performance of LROGNav using different pre-trained models provided by Swin-Unet \cite{cao2022swinunet} and various input resolutions, as shown in Table \ref{tab:fps}. Since LROGNav is fundamentally an image-to-image model, its performance improves with higher input resolution. Considering the need for efficiency in real-world application transfer, we have selected the Swin-B model with a resolution of $384 \times 384$ for LROGNav.

\begin{table*}[!t]
\caption{ObjectNav validation results on Gibson and MP3D datasets. All results used for comparison are extracted from cited papers.}
\renewcommand\arraystretch{1.9}
\begin{center}
\scalebox{0.54}{
\begin{tabular}{c|c|ccccc|ccccccccccc}
\hline
\multirow{2}{*}{\textbf{Dataset}} & \multirow{2}{*}{\textbf{Metrics/ Method}} & \multicolumn{5}{c|}{\textbf{End-to-end}}                                                    & \multicolumn{11}{c}{\textbf{Modular-based}}                                                                                                                                                                    \\ \cline{3-18} 
                                  &                                           & \textbf{\begin{tabular}[c]{@{}c@{}}DD-PPO \\  \cite{wijmans2019dd} \end{tabular} } & \textbf{ \begin{tabular}[c]{@{}c@{}}Red-Rabbit \\  \cite{ye2021auxiliaryred} \end{tabular}} & \textbf{\begin{tabular}[c]{@{}c@{}}THDA \\  \cite{maksymets2021thda} \end{tabular}} & \textbf{ \begin{tabular}[c]{@{}c@{}}Habitat-web \\  \cite{ramrakhya2022habitat} \end{tabular}} & \textbf{\begin{tabular}[c]{@{}c@{}}RIM \\  \cite{chen2023objectRIM} \end{tabular}} & \textbf{ \begin{tabular}[c]{@{}c@{}}FBE \\  \cite{yamauchi1997frontier} \end{tabular}} & \textbf{\begin{tabular}[c]{@{}c@{}}ANS \\  \cite{chaplot2020learningans} \end{tabular}} & \textbf{\begin{tabular}[c]{@{}c@{}}SemExp \\  \cite{semexp} \end{tabular}} & \textbf{\begin{tabular}[c]{@{}c@{}}PONI \\  \cite{ramakrishnan2022poni} \end{tabular}} & \textbf{ \begin{tabular}[c]{@{}c@{}}L3MVN \\  (Zero-shot) \cite{yu2023l3mvn} \end{tabular} } & \textbf{\begin{tabular}[c]{@{}c@{}}L3MVN \\  (Fine-tune) \cite{yu2023l3mvn} \end{tabular}} & \textbf{\begin{tabular}[c]{@{}c@{}}FSE \\  \cite{FSE} \end{tabular}} & \textbf{\begin{tabular}[c]{@{}c@{}}STUBBORN \\  \cite{luo2022stubborn} \end{tabular}} & \textbf{\begin{tabular}[c]{@{}c@{}}PEANUT \\  \cite{zhai2023peanut} \end{tabular}} & \textbf{\begin{tabular}[c]{@{}c@{}}MON \\  \cite{MON} \end{tabular}} & \textbf{\begin{tabular}[c]{@{}c@{}}LROGNav \\  Ours \end{tabular} } \\ \hline
\multirow{2}{*}{\textbf{Gibson}}  & SR $ \uparrow $                                       & 15.0            & -                   & -             & -                    & -            & 64.3         & 67.1         & 71.7            & 73.6          & 76.1                       & 76.9                       & 71.5         & -                 & -               & 62.5         &        \underline{\textbf{78.3}}          \\
                                  & SPL $ \uparrow $                                       & 10.7            & -                   & -             & -                    & -            & 28.3         & 34.9         & 39.6            & 41.0          & 37.7                       & 38.8                       & 36.0         & -                 & -               & 31.5         &    \underline{\textbf{44.6}} ($ \uparrow $ 8.8\%)                \\ \hline
\multirow{2}{*}{\textbf{MP3D}}    & SR $ \uparrow $                                        & 8.0             & 34.6                & 28.4          & 35.4                 & \underline{\textbf{50.3}}           & 22.7         & 27.3         & -               & 31.8          & -                          & -                          & -            & 37.0              & 40.5                & 36.9         &          50.1           \\
                                  & SPL $ \uparrow $                                       & 1.8             & 7.9                 & 11.0          & 10.2                 & 17.0         & 7.2          & 9.2          & -               & 12.1          & -                          & -                          & -            & -                 & 15.8            & 16.3         &     \underline{\textbf{19.1}} ($ \uparrow $ 12.4\%)                 \\ \hline
\end{tabular}
}
\end{center}
\label{tab:eva_all}
\end{table*}

\begin{figure*}[!t]
	\centering
		\includegraphics[scale=.6]{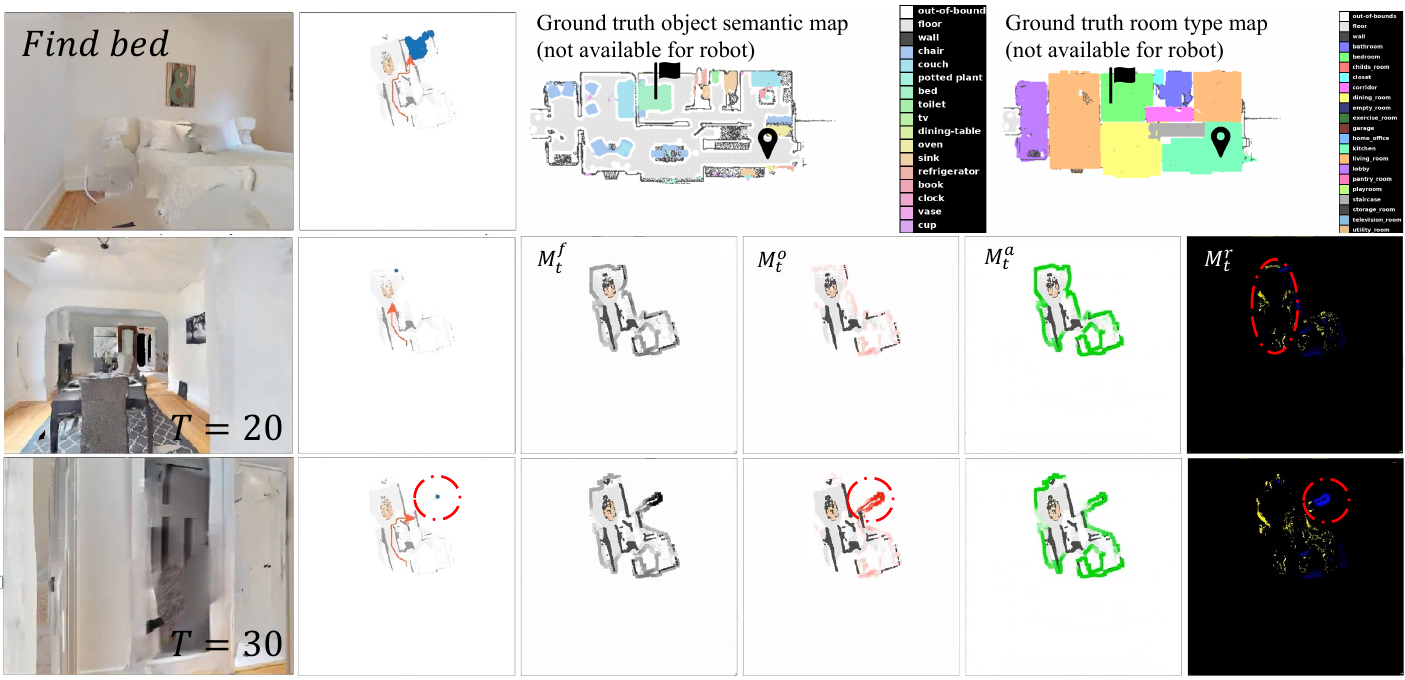}
	\caption{The qualitative example of the proposed LROGNav. The top-row four figures: The final observation(target) with entire trajectory, the ground truth floor map with semantic annotations for object and room types, with the start and target positions marked. The rows below display sequences from the LROGNav tasked to "Find bed" at different timestamps. From left to right, the sequence includes the RGB observation, real-time mapping (the long-term goal is denoted by the blue dot), frontier-based fusion map $M_{t}^{f}$, object potential map $M_{t}^{o}$, area potential map $M_{t}^{a}$, and O2R relation potential map $M_{t}^{r}$, where blue represents positive LLM scores and yellow denotes negative LLM scores.}
	\label{FIG:LROG_eva}
\end{figure*}

\subsection{Results in Habitat simulation}
We categorize existing ObjectNav works, which have been evaluated on either the Gibson or MP3D datasets, into two approaches: end-to-end, which directly outputs the actions mainly utilizing RL, and modular-based, which predicts a map-based long-term goal, as listed in Table \ref{tab:eva_all}.\newline

\noindent {\bf End-to-end RL baselines } 
\begin{itemize}
    \item DD-PPO \cite{wijmans2019dd}: This is a standard approach utilizing end-to-end reinforcement learning, which is expanded across multiple nodes for distributed training.
    \item Red-Rabbit \cite{ye2021auxiliaryred}: This model enhances DD-PPO by incorporating additional auxiliary tasks, leading to greater efficiency in sample use and better adaptability to novel environments. 
    \item THDA \cite{maksymets2021thda}: This introduces the concept of "Treasure Hunt Data Augmentation" which refines the rewards and inputs used in reinforcement learning, thereby improving adaptability to unfamiliar scenes.
    \item Habitat-web \cite{ramrakhya2022habitat}: They utilizes a comprehensive approach of imitation learning, leveraging demonstrations conducted by human operators.
    \item RIM \cite{chen2023objectRIM}: RIM introduces a novel approach via an implicit spatial map. This map is continuously updated with each new observation using transformer architecture. To bolster spatial analytical skills, they integrate auxiliary tasks, empowering their model to not only develop explicit maps but also to anticipate visual details, assign semantic tags, and deduce potential actions.
\end{itemize}

\noindent {\bf Modular-based baselines } 
\begin{itemize}
    \item FBE \cite{yamauchi1997frontier}: Employing a traditional method of frontier-based exploration, this strategy generates a 2D map of occupancy and moves towards the closest frontiers of the map. Upon identifying the target object through semantic segmentation, it approaches the target following a calculated local strategy before halting.
    \item ANS \cite{chaplot2020learningans}: This strategy formulates a modular reinforcement learning approach aimed at enhancing the coverage of an area. It adopts an identical approach to FBE for identifying objectives and implementing cessation.
    \item SemExp \cite{semexp}: Known as the leading modular technique in ObjectNav, this method employs reinforcement learning for dynamic interaction training, enabling the policy to determine long-term objectives effectively. 
    \item PONI \cite{ramakrishnan2022poni}: It is a supervised, image-to-image, UNet-based modular approach that separates perception from navigation tasks and learns without direct environmental interaction, utilizing area and object potential functions which are trained on semantic maps.
    \item L3MVN \cite{yu2023l3mvn}: It is a framework is designed to construct a map of the environment and identify long-term objectives through frontier analysis, utilizing large language model insights for more effective exploration and search activities.
    \item FSE \cite{FSE}: This is a structure aims to develop a map of the surroundings and determine long-term targets by leveraging frontiers in a modular-based manner, employing deep reinforcement learning techniques for optimized exploration and search efficiency.
    \item STUBBORN \cite{luo2022stubborn}: It employs a modular strategy that designates the objective to one of the four corners of a local section of the map, adapting through rotation upon encountering a cul-de-sac. Additionally, it utilizes a heuristic approach to integrate target detections over multiple frames.
    \item PEANUT \cite{zhai2023peanut}: They present a method to predict unseen object locations with incomplete maps, using global context. The lightweight model is efficiently trained with limited data and can be seamlessly integrated into ObjectNav frameworks without the necessity for reinforcement learning.
    \item MON \cite{MON}: MON presents a concise reinforcement learning setup with a hybrid policy designed for multi-object navigation (We extract the results of one-object navigation), focusing on minimizing unnecessary actions. The policy function predicts a probable object location, aiming to investigate the positions most likely associated with the target.

\end{itemize}

Through the results listed in Table \ref{tab:eva_all}, the proposed LROGNav achieves the best performance in 3 out of 4 metrics. Notably, the SPL, which represents the efficiency of the ObjectNav task, has improved by an average of 10.6\% compared to the second-best related works. RIM \cite{chen2023objectRIM} achieves the best performance in SR, which is also based on a multi-modal transformer framework. However, compared to the end-to-end approach RIM, the modular-based approach exhibits a smaller sim-to-real gap \cite{wang2023survey} \cite{li2022object} \cite{gervet2023navigating}, as evidenced by the real-world demonstrations provided for both methods.

The qualitative example of the proposed LROGNav, which utilizes three frontier-based potential maps regressed by multiple decoders, is shown in Fig. \ref{FIG:LROG_eva}. From these observations, it is evident that at the initial period ($T=20$), when there is not much semantic information collected and the distance to the target is far, the long-term goal is predominantly influenced by the area potential map $M_{t}^{a}$, as verified in the original PONI \cite{ramakrishnan2022poni}. Since the bed is a type of distinct, room-specific object usually found in bedrooms, the O2R potential map $M_{t}^{r}$ predicts mostly negative values. As the agent approaches closer to the target, $M_{t}^{o}$ predicts high values for frontiers close to the target accurately, and $M_{t}^{r}$ assigns positive values to the frontiers located in the bedroom. These predictions are fused into $M_{t}^{f}$, guiding the estimation of the long-term goal towards a high-potential position where the target object is located. The final trajectory demonstrates that the proposed LROGNav searches for the target object in an efficient manner. More examples from simulations are provided in the ablation study Section \ref{sec:ablation}, the Appendix materials, and on the project webpage (\url{https://sunleyuan.github.io/ObjectNav}).

\begin{figure*}[!t]
	\centering
		\includegraphics[scale=.6]{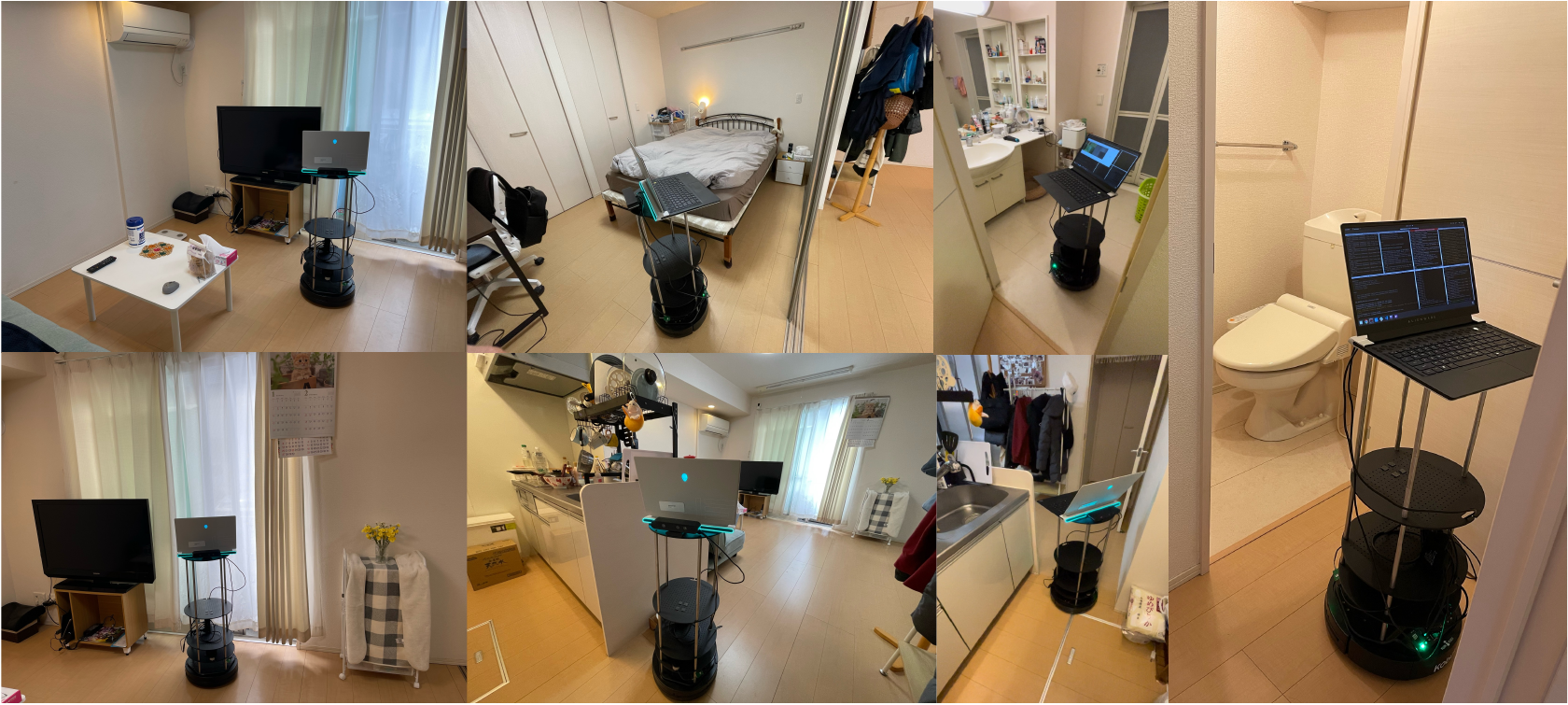}
	\caption{The environment for real-world experiments with Kobuki robot. The robot operates in various rooms: the living room, bedroom, bathroom/washroom, kitchen, and toilet, respectively. }
	\label{FIG:home_exp}
\end{figure*}


\subsection{Results in real world with Kobuki robot}

We deploy the LROGNav framework on the Kobuki robot using ROS to connect the PC (Intel i9-12900H, up to 5.0GHz Turbo, RTX 3080Ti) to the robot hardware platform. To reduce the sim-to-real gap for real-world experiments, we maintain the same height (0.88m) and radius (0.18m) for the camera (Astra RGB-D) and the robot base as configured in the Habitat simulator. Although the Kobuki base can output odometry through a wheel encoder fused with an internal inertial measurement unit, the pose in the Habitat simulator is perfectly accurate without any noise \cite{semexp} \cite{gervet2023navigating}. However, map-based approaches are very sensitive to the quality of mapping \cite {wang2023survey}. To address this issue, we implement a 2D LiDAR-based Hector SLAM \cite{kohlbrecher2011flexible} for more robust real-time localization results, which are fed into the LROGNav for real-world experiments. For safety reasons, the LiDAR is also used for obstacle avoidance, as the camera is positioned high and its depth range is limited compared to the LiDAR (a RPLiDAR A1 is positioned at 25cm height). However, for a fair comparison, this information is not utilized in the mapping module of the LROGNav framework. The hardware configuration described is similar to those used in other ObjectNav-related works that conduct real-world experiments, such as SemExp \cite{semexp}, L3MVN \cite{yu2023l3mvn} and FSE \cite{FSE}.
\begin{CJK}{UTF8}{min} 
\begin{figure}[!t]
	\centering
		\includegraphics[scale=.56]{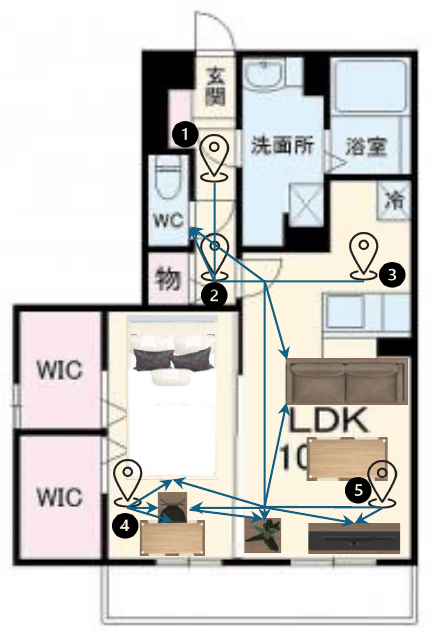}
	\caption{The floor map for the experiments shows the house (50.36$m^{2}$) layout. Five starting locations and target objects are marked on the map. Auxiliary lines represent the shortest path lengths from five different positions to six target objects, including: \texttt{"chair", "couch", "potted plant", "bed", "toilet", "tv"}. The Japanese characters in the map denote the following: \texttt{"玄関" -- "entrance", "物" -- "storage room", "洗面所" -- "washroom", "浴室" -- "bathroom", "冷" -- "refrigerator"}. }
	\label{FIG:home}
\end{figure}
\end{CJK}

The experiment is conducted in a house environment, as illustrated in Fig. \ref{FIG:home_exp}. The interior layout of the house is depicted in Fig. \ref{FIG:home}. We designed the experiments following the protocols in ROS4VSN \cite{gutierrez2023visual}. The robot starts from five different locations distributed throughout the house, and the target objects belong to six categories, consistent with the Gibson dataset evaluation. We calculate the SR and manually measure the optimal path length for each testing episode to collect the SPL metric. The quantitative evaluation results are shown in Table \ref{tab:home_exp}.

In Table \ref{tab:home_exp}, we collect the results for each testing episode. The average SR and SPL are 82.8\% and 74.9\%, respectively, in this house with an average optimal testing length of 4.7$m$ and an area of 50.36$m²$. Compared with the results published in \cite{gervet2023navigating}, which evaluated the model across 6 homes over 60 episodes, SemExp achieves the SR of 90\% and SPL of 64\%. However, directly comparing these quantitative results is not meaningful since the tests were not conducted under identical scenarios and protocols. Nevertheless, there are still phenomena to be observed behind the numbers. For instance, our tests were conducted in a Japanese-style house, which differs significantly from the American-style houses used in \cite{gervet2023navigating}. For example, the main reason for failure in our scenario is collisions in narrow spaces, such as the corridor at position 1, as illustrated in Fig. \ref{FIG:home}. While areas in American houses are typically more spacious than those in Japanese-style ones, which may explain the higher SR, the smaller size could, conversely, lead to a higher SPL. For example, in our tests, some target objects were very close to the start position, such as at position 4 to find a chair. 

\begin{figure*}[!t]
	\centering
		\includegraphics[scale=.43]{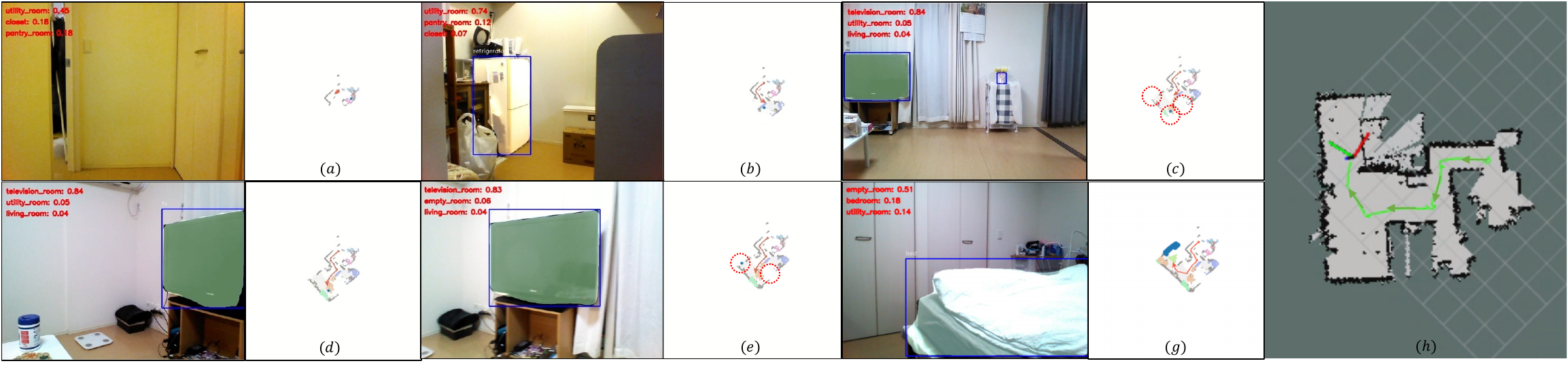}
	\caption{The qualitative evaluation example of "Find bed" in real-world experiments is presented as follows: (a) to (g) depict the RGB observations and real-time maps; (h) is the final trajectory map generated by Hector SLAM, of which only the pose estimation has been utilized in the LROGNav framework. }
	\label{FIG:home_eva}
\end{figure*}

\begin{table}[ht]
\caption{Quantitative evaluations of real robot experiments at home environment}
\renewcommand\arraystretch{1.5}
\begin{center}
\scalebox{0.66}{
\begin{tabular}{
>{}c 
>{}c 
>{}c 
>{}c 
>{}c 
>{}c }
\hline
\multicolumn{1}{c|}{{ Loc.}}                & \multicolumn{1}{c|}{{ Obj.}}   & \multicolumn{1}{c|}{{  \begin{tabular}[c]{@{}c@{}}Shortest\\ length (m)\end{tabular} }} & \multicolumn{1}{c|}{{ SR}} & \multicolumn{1}{c|}{{ SPL}} & { Fail reason}            \\ \hline
\multicolumn{1}{c|}{{ }}                    & \multicolumn{1}{c|}{{ chair}}  & { 8.0}                                                          & { \textbf{\texttimes}}                                               & { }                                                 & { Detection error} \\
\multicolumn{1}{c|}{{ }}                    & \multicolumn{1}{c|}{{ couch}}  & { 4.1}                                                          & { \textbf{\texttimes}}                                               & { }                                                 & { Collisions}      \\
\multicolumn{1}{c|}{{ }}                    & \multicolumn{1}{c|}{{ plant}}  & { 7.2}                                                          & { \checkmark}                                               & { 76}                                               & { }                \\
\multicolumn{1}{c|}{{ }}                    & \multicolumn{1}{c|}{{ bed}}    & { 8.1}                                                          & { \checkmark}                                               & { 72}                                               & { }                \\
\multicolumn{1}{c|}{{ }}                    & \multicolumn{1}{c|}{{ toilet}} & { /}                                                            & { /}                                               & { / }                                                 & { Door closed}     \\
\multicolumn{1}{c|}{\multirow{-6}{*}{{ 1}}} & \multicolumn{1}{c|}{{ tv}}     & { 8.0}                                                          & { \textbf{\texttimes}}                                               & { }                                                 & { Collisions}      \\ \hline
\multicolumn{1}{c|}{{ }}                    & \multicolumn{1}{c|}{{ chair}}  & { 6.3}                                                          & {\textbf{\texttimes}}                                               & { }                                                 & { Detection error} \\
\multicolumn{1}{c|}{{ }}                    & \multicolumn{1}{c|}{{ couch}}  & { 2.3}                                                          & { \checkmark}                                               & { 78}                                               & { }                \\
\multicolumn{1}{c|}{{ }}                    & \multicolumn{1}{c|}{{ plant}}  & { 5.5}                                                          & { \checkmark}                                               & { 77}                                               & { }                \\
\multicolumn{1}{c|}{{ }}                    & \multicolumn{1}{c|}{{ bed}}    & { 6.4}                                                          & { \checkmark}                                               & { 75}                                               & { }                \\
\multicolumn{1}{c|}{{ }}                    & \multicolumn{1}{c|}{{ toilet}} & { 1}                                                            & {\textbf{\texttimes}}                                               & { }                                                 & { Collisions}      \\
\multicolumn{1}{c|}{\multirow{-6}{*}{{ 2}}} & \multicolumn{1}{c|}{{ tv}}     & { 6.3}                                                          & { \checkmark}                                               & { 79}                                               & { }                \\ \hline
\multicolumn{1}{c|}{{ }}                    & \multicolumn{1}{c|}{{ chair}}  & { 7.2}                                                          & { \checkmark}                                               & { 73}                                               & { }                \\
\multicolumn{1}{c|}{{ }}                    & \multicolumn{1}{c|}{{ couch}}  & { 3.2}                                                          & { \checkmark}                                               & { 75}                                               & { }                \\
\multicolumn{1}{c|}{{ }}                    & \multicolumn{1}{c|}{{ plant}}  & { 6.4}                                                          & { \checkmark}                                               & { 93}                                               & { }                \\
\multicolumn{1}{c|}{{ }}                    & \multicolumn{1}{c|}{{ bed}}    & { 7.3}                                                          & { \checkmark}                                               & { 78}                                               & { }                \\
\multicolumn{1}{c|}{{ }}                    & \multicolumn{1}{c|}{{ toilet}} & { 3.7}                                                          & { \textbf{\texttimes}}                                               & { }                                                 & { Collisions}      \\
\multicolumn{1}{c|}{\multirow{-6}{*}{{ 3}}} & \multicolumn{1}{c|}{{tv}}     & { 7.2}                                                          & { \checkmark}                                               & { 91}                                               & { }                \\ \hline
\multicolumn{1}{c|}{{ }}                    & \multicolumn{1}{c|}{{ chair}}  & { 0.6}                                                          & { \checkmark}                                               & { 93}                                               & { }                \\
\multicolumn{1}{c|}{{ }}                    & \multicolumn{1}{c|}{{ couch}}  & { 3.5}                                                          & { \checkmark}                                               & { 82}                                               & { }                \\
\multicolumn{1}{c|}{{ }}                    & \multicolumn{1}{c|}{{ plant}}  & { 1.3}                                                          & { \checkmark}                                               & { 88}                                               & { }                \\
\multicolumn{1}{c|}{{ }}                    & \multicolumn{1}{c|}{{ bed}}    & { 1}                                                            & { \checkmark}                                               & { 95}                                               & { }                \\
\multicolumn{1}{c|}{{ }}                    & \multicolumn{1}{c|}{{ toilet}} & { 7.8}                                                          & { \checkmark}                                               & { 77}                                               & { }                \\
\multicolumn{1}{c|}{\multirow{-6}{*}{{ 4}}} & \multicolumn{1}{c|}{{ tv}}     & { 0.6}                                                          & { \checkmark}                                               & { 95}                                               & { }                \\ \hline
\multicolumn{1}{c|}{{ }}                    & \multicolumn{1}{c|}{{ chair}}  & { 3.4}                                                          & { \checkmark}                                               & { 35}                                               & { }                \\
\multicolumn{1}{c|}{{ }}                    & \multicolumn{1}{c|}{{ couch}}  & { 3.7}                                                          & { \checkmark}                                               & { 55}                                               & { }                \\
\multicolumn{1}{c|}{{ }}                    & \multicolumn{1}{c|}{{ plant}}  & { 2.6}                                                          & { \checkmark}                                               & { 83}                                               & { }                \\
\multicolumn{1}{c|}{{ }}                    & \multicolumn{1}{c|}{{ bed}}    & { 3.5}                                                          & { \checkmark}                                               & { 82}                                               & { }                \\
\multicolumn{1}{c|}{{ }}                    & \multicolumn{1}{c|}{{ toilet}} & { 8.0}                                                          & { \checkmark}                                               & { 70}                                               & { }                \\
\multicolumn{1}{c|}{\multirow{-6}{*}{{ 5}}} & \multicolumn{1}{c|}{{ tv}}     & { 0.7}                                                          & { \checkmark}                                               & { 1}                                                & { }                \\ \hline
\multicolumn{2}{c}{{ Avg.}}                                                                                              & { 4.7}                                                          & { 82.8}                                            & { 74.9}                                             & { }                \\ \hline
\end{tabular}
}
\end{center}
\label{tab:home_exp}
\end{table}

Fig. \ref{FIG:home_eva} presents a qualitative example of finding a bed in this home scenario. The robot starts at position 1 and needs to travel through the corridor, observe the kitchen, and then explore the living room before finally locating the bedroom with the bed. The optimal shortest path length for this episode is 8.1$m$, with an SPL achievement of 72\%. Initially, the long-term goal is placed in a space where more semantic information has been detected (see (a)). When the agent is in the kitchen space, since only one refrigerator has been detected and the viewpoint is narrow, the CLIP model does not correctly predict the observation as a kitchen. However, as the target bed has low relations to other rooms except the bedroom, the long-term goal is set outside of the kitchen space in (b). When the robot traverses the living room, it faces three options, as shown in (c): to the right (towards the bedroom), to the left (the living room with a TV), and to the bottom left (an unexplored area of the living room without any semantic information). The model initially selects the left in (c), likely because it generally prefers directions with more semantic information. This strategy benefits navigation as it prevents the robot from backtracking or missing targets it has already explored. After reaching the previous long-term goal in (d), and thanks to accurate CLIP-based room estimation, the LROGNav model predicts the long-term goal towards the right frontier in (e), opting not to explore the unexplored living room area in the bottom left. This decision validates the LROGNav concept, which utilizes LLM-based O2R knowledge to enhance the efficiency of ObjectNav tasks. Ultimately, the robot moves towards the bedroom and successfully finds the bed, as illustrated in (f).

In general, thanks to the smaller sim-to-real gap found in map-based modular approaches \cite{semexp}\cite{yu2023l3mvn}\cite{gervet2023navigating}\cite{wang2023survey} as compared to end-to-end approaches \cite{chen2023objectRIM}\cite{chen2023think}, SR can typically be maintained in real-world applications if the challenges related to obstacle avoidance and object detection/segmentation are not overly significant. However, SPL should be the primary concern that requires further efforts.

\begin{table}[!t]
\caption{Ablation study on network architecture and promptings.}
\renewcommand\arraystretch{1.9}
\begin{center}
\scalebox{0.6}{
\begin{tabular}{cccccc|cc}
\hline
\multicolumn{1}{c|}{\begin{tabular}[c]{@{}c@{}}No./\\ Components\end{tabular}} & \multicolumn{1}{c|}{\begin{tabular}[c]{@{}c@{}}UNet\\ (PONI)\end{tabular}} & \multicolumn{1}{c|}{\begin{tabular}[c]{@{}c@{}}Multi-modal \\ Transformer arc.\end{tabular}} & \multicolumn{1}{c|}{COT} & \multicolumn{1}{c|}{\begin{tabular}[c]{@{}c@{}}Positive\\ prompts\end{tabular}} & \begin{tabular}[c]{@{}c@{}}Negative\\ prompts\end{tabular} & \multicolumn{1}{c|}{SR $ \uparrow $} & SPL $ \uparrow $  \\ \hline
\multicolumn{1}{c|}{1}                                                         & \checkmark                                                                          &                                                                                              &                          &                                                                                 &                                                            & 73.6                    & 41   \\ \cline{1-1}
\multicolumn{1}{c|}{2}                                                         &                                                                            & \checkmark                                                                                            &                          &                                                                                 &                                                            & 75.2                    & 42.1 \\ \hline
\multicolumn{1}{c|}{3}                                                         &                                                                            & \checkmark                                                                                            &                          & \checkmark                                                                               &                                                            & 76.3                    & 43.0 \\ \cline{1-1}
\multicolumn{1}{c|}{4}                                                         &                                                                            & \checkmark                                                                                            &                          &                                                                                 & \checkmark                                                          & 76.1                    & 43.8 \\ \hline
\multicolumn{1}{c|}{5}                                                         &                                                                            & \checkmark                                                                                            &                          & \checkmark                                                                               & \checkmark                                                          & 78.0                    & 44.1 \\ \cline{1-1}
\multicolumn{1}{c|}{\textbf{6 (Ours)}}                                                         &                                                                            & \checkmark                                                                                            & \checkmark                        & \checkmark                                                                               & \checkmark                                                          & 78.3                    & 44.6 \\ \cline{1-1}
\multicolumn{1}{c|}{7}                                                         & \checkmark                                                                          &                                                                                              & \checkmark                        & \checkmark                                                                               & \checkmark                                                          & 76.7                    & 43.1 \\ \hline
\end{tabular}
}
\end{center}
\label{tab:promptings}
\end{table}
\begin{figure}[!t]
	\centering
		\includegraphics[scale=.88]{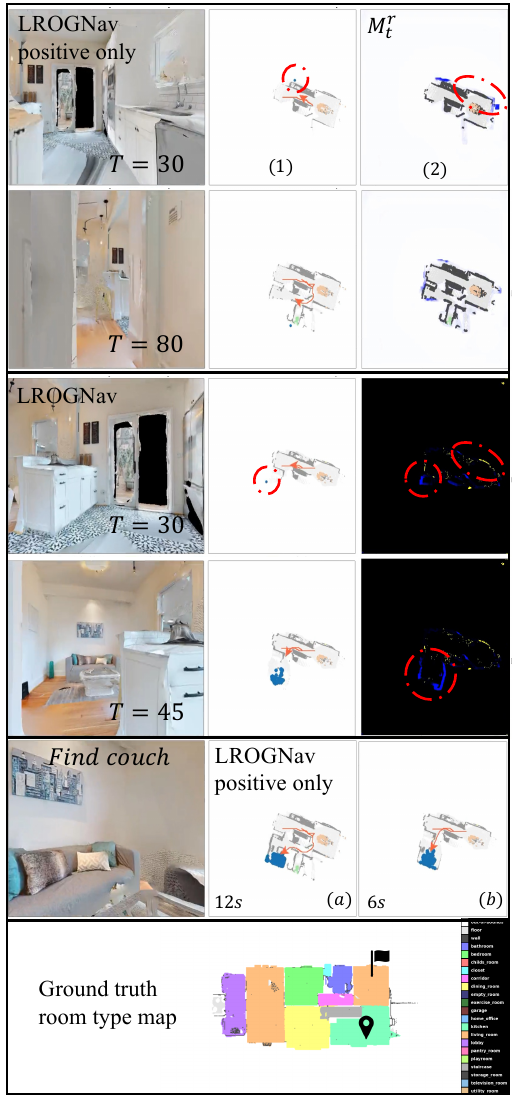}
	\caption{The qualitative example of "positive only" and "positive with negative" promptings. LROGNav with positive only takes 12s, full LORGNav costs 6s.}
	\label{FIG:VSpromoting}
\end{figure}

\subsection{Ablation study and discussions}
\label{sec:ablation}

In this section, we demonstrate the contributions of each component in the proposed LROGNav framework using ablation study evaluations. Additionally, we explore discussions related to ObjectNav as furue works.

$\bullet$ (1) How do "positive with negative" and "positive or negative only" promptings compare in the proposed framework?

In Table \ref{tab:promptings}, we categorize our promptings into modules, including Chain of Thought (COT), negative, and positive manners. From this ablation study, it emerges that "positive only" slightly outperforms "negative only" in terms of SR, as seen in No. 3 and No. 4. Conversely, "negative only" demonstrates an advantage in SPL compared with "positive only", as shown in No. 3 and No. 4. Differently from \cite{shah2023navigation}, which also employs COT in combination with positive and negative promptings, the impact of COT in our case (see No. 5 and No. 6) is not as significant as in their study. A possible reason is that we deploy the COT strategy only once offline, whereas they implement the promptings online during navigation each time the agent acquires new observations, which allows the advantages of COT to accumulate over multiple instances. Additionally, we observe that the most significant improvement among these components comes from changing the network architecture from UNet to multi-modal Swin-Unet (see comparisons between No. 1 and No. 2, and No. 6 and No. 7).

\begin{figure*}[!t]
	\centering
		\includegraphics[scale=.79]{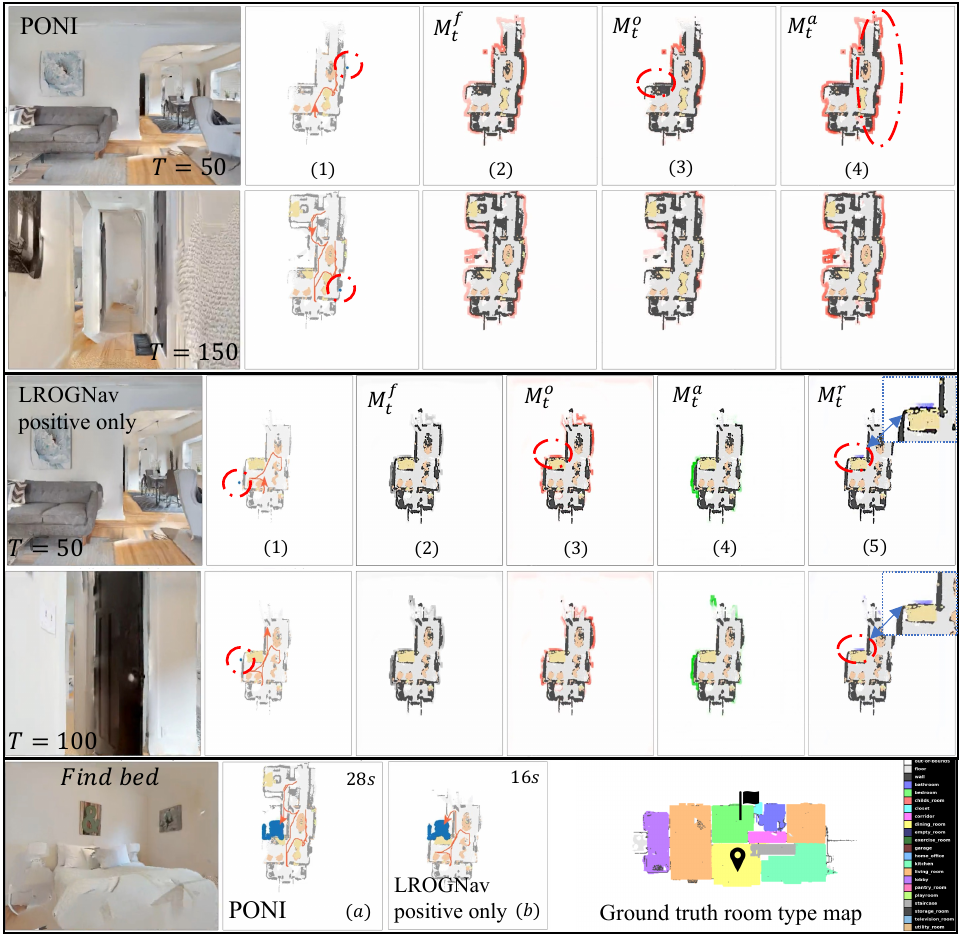}
	\caption{Qualitative examples of PONI and LROGNav with only positive promptings. We display the potential maps for each approach at two timestamps in the first four rows. The bottom row illustrates the final trajectory comparison: PONI required 28 seconds, whereas LROGNav with only positive promptings took 16 seconds to find the bed. The start and target locations are marked on the ground truth map. }
	\label{FIG:VSPONI}
\end{figure*}

We also present qualitative examples of PONI and LROGNav with only positive promptings, as illustrated in Fig. \ref{FIG:VSPONI}. Thanks to the multi-modal Swin-Unet architecture and our proposed LLM-based O2R relation potential decoder, LROGNav with positive promptings takes 10 seconds less than the PONI baseline in this testing episode. Observing the visualizations of long-term goal predictions in column (1) of two methods, we can note that PONI predicts the goal on the incorrect side compared to ours, which results in the agent spending more time exploring. This discrepancy is primarily due to our model's more accurate prediction of the area potential map $M_{t}^{a}$, as shown in column (4). Additionally, the object potential map $M_{t}^{o}$ (column 3) and the O2R relation potential map $M_{t}^{r}$ (column 5) in our approach accurately assign high/positive values to the bed and bedroom, significantly improving navigation efficiency.

Fig. \ref{FIG:VSpromoting} illustrates the comparison between the "positive only" and "positive with negative" approaches within the LROGNav framework. Through the visualizations in column (5) of the O2R relation potential map $M_{t}^{r}$, we can observe that the positive values are denoted in blue located in bedroom and negative values in yellow. This combination makes LROGNav more efficient than the "positive only" approach, saving more time on exploration. Generally, negative scores deter the agent from venturing into areas unlikely to contain the goal, while positive scores attract the agent to areas where finding the goal is more probable. Using only positive scores makes the navigation strategy more progressive, whereas relying solely on negative scores makes it more conservative. The combination of both approaches strikes a balance between them.

$\bullet$ (2) What is the performance of the different combinations (subsets) of the three decoders? Which component contributes the most to LROGNav?

To address this question, we conducted exhaustive ablation studies on various combinations of multi-decoders, as shown in Table \ref{tab:multideocders}. Through comparisons utilizing only one decoder at a time (No.1-3), it turns out that the O2R potential decoder $U^{r}$ achieves the best performance. Conversely, similar to PONI, the worst performance occurs when solely using the object potential decoder $U^{o}$. The likely reason is that even when using only $U^{r}$, the agent retains some level of exploration ability, moving from one room to another, while also possessing knowledge related to target objects. However, relying solely on $U^{o}$ limits this exploration ability, which has been verified as very important for the ObjectNav task \cite{semexp} \cite{chen2020learning} \cite{ye2021auxiliary}.

Similar to the performance observed when using only one decoder, combinations of any two among them have been tested in No.4-7. The results can be summarized as follows: $U^{r}$ contributes more than $U^{a}$, which in turn contributes more than $U^{o}$. This suggests that predicting only the potential location of the target object is difficult to achieve good performance if you ignore exploration ability and common-sense knowledge.

Moreover, we test LROGNav with ground truth (GT) segmentation results (No.8 and 9) in the Habitat simulator, as both PONI \cite{ramakrishnan2022poni} and SemExp \cite{semexp} conducted, to verify that the main source of error in ObjectNav is due to segmentation errors. This is because segmentation affects the quality of mapping and the agent's behavior when the target object has been observed.

\begin{table*}[!t]
\caption{Ablation study comparing multi-decoders with PONI. The number before the slash is taken from the PONI paper.}
\renewcommand\arraystretch{1.9}
\begin{center}
\scalebox{0.66}{
\begin{tabular}{ccccc|cccc}
\hline
\multicolumn{5}{c|}{PONI VS LROGNav Ablation Study}                                                                                                                                                                                                                                                                                                                                                        & \multicolumn{2}{c|}{Gibson}                                                                                                                                      & \multicolumn{2}{c}{MP3D}                                                                                                                    \\ \hline
\multicolumn{1}{c|}{\begin{tabular}[c]{@{}c@{}}No./\\ Components\end{tabular}} & \multicolumn{1}{c|}{\begin{tabular}[c]{@{}c@{}}Object \\ Potential\end{tabular}} & \multicolumn{1}{c|}{\begin{tabular}[c]{@{}c@{}}Area \\ Potential\end{tabular}} & \multicolumn{1}{c|}{\begin{tabular}[c]{@{}c@{}}Object-to-Room \\ Potential\end{tabular}} & \begin{tabular}[c]{@{}c@{}}GT \\ Segmentation\end{tabular} & \multicolumn{1}{c|}{\begin{tabular}[c]{@{}c@{}}SR $ \uparrow $\\ PONI/LROGNav\end{tabular}} & \multicolumn{1}{c|}{\begin{tabular}[c]{@{}c@{}}SPL $ \uparrow $\\ PONI/LROGNav\end{tabular}} & \multicolumn{1}{c|}{\begin{tabular}[c]{@{}c@{}}SR $ \uparrow $\\ PONI/LROGNav\end{tabular}} & \begin{tabular}[c]{@{}c@{}}SPL $ \uparrow $ \\ PONI/LROGNav\end{tabular} \\ \hline
\multicolumn{1}{c|}{1}                                                         & \checkmark                                                                                &                                                                                &                                                                                          &                                                            & 65.1/68.3                                                                      & 37.9/39.8                                                                       & 30.8/31.4                                                                      & 12.0/13.1                                                  \\ \cline{1-1}
\multicolumn{1}{c|}{2}                                                         &                                                                                  & \checkmark                                                                              &                                                                                          &                                                            & 72.7/73.5                                                                      & 39.4/40.8                                                                       & 31.1/31.9                                                                      & 11.8/13.3                                                  \\ \cline{1-1}
\multicolumn{1}{c|}{3}                                                         &                                                                                  &                                                                                & \checkmark                                                                                        &                                                            & /74.1                                                                          & /41.9                                                                           & /31.7                                                                          & /14.1                                                      \\ \hline
\multicolumn{1}{c|}{4}                                                         & \checkmark                                                                                & \checkmark                                                                              &                                                                                          &                                                            & 73.6/75.2                                                                      & 41.0/42.0                                                                       & 31.8/35.7                                                                      & 12.1/13.9                                                  \\ \cline{1-1}
\multicolumn{1}{c|}{5}                                                         & \checkmark                                                                                & \checkmark                                                                              &                                                                                          & \checkmark                                                          & 86.5/88.3                                                                      & 51.5/53.2                                                                       & 58.2/60.7                                                                      & 27.5/31.0                                                  \\ \cline{1-1}
\multicolumn{1}{c|}{6}                                                         & \checkmark                                                                                &                                                                                & \checkmark                                                                                       &                                                            & /75.4                                                                          & /43.0                                                                           & /37.2                                                                          & /14.7                                                      \\ \cline{1-1}
\multicolumn{1}{c|}{7}                                                         &                                                                                  & \checkmark                                                                              & \checkmark                                                                                        &                                                            & /77.4                                                                          & /43.9                                                                           & /43.7                                                                          & /17.6                                                      \\ \hline
\multicolumn{1}{c|}{\textbf{8 (Ours)}}                                                         & \checkmark                                                                                & \checkmark                                                                              & \checkmark                                                                                        &                                                            & /78.3                                                                          & /44.6                                                                           & /50.1                                                                          & /19.1                                                      \\ \cline{1-1}
\multicolumn{1}{c|}{9}                                                         & \checkmark                                                                                & \checkmark                                                                              & \checkmark                                                                                        & \checkmark                                                          & /91.6                                                                          & /57.3                                                                           & /63.6                                                                          & /35.9                                                      \\ \hline
\end{tabular}
}
\end{center}
\label{tab:multideocders}
\end{table*}

\begin{figure}[!t]
	\centering
		\includegraphics[scale=.5]{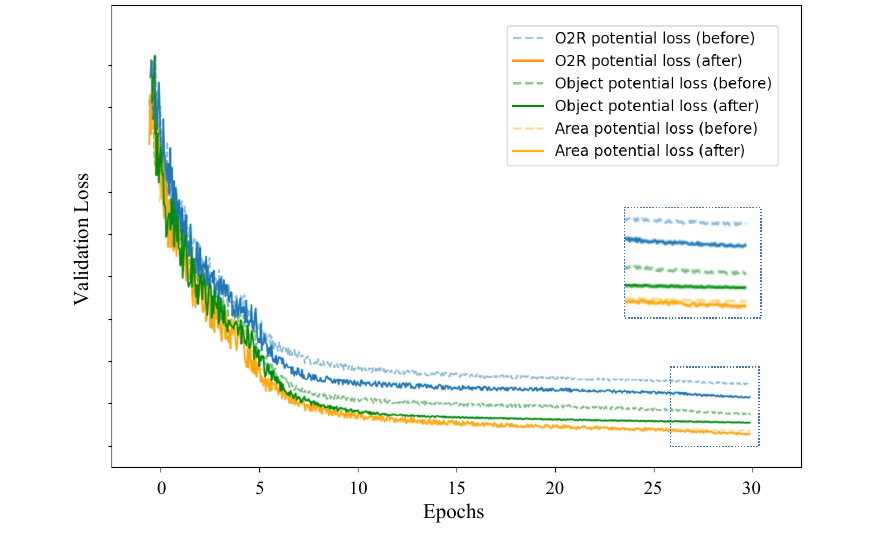}
	\caption{The loss curve of three decoders on validation. Before: only utilizting the RGB-D as input modlity like PONI, after: multi-modal inputs in LORGNav.}
	\label{FIG:losscurve}
\end{figure}

\begin{figure}[!t]
    \centering
    \begin{subfigure}[b]{0.35\textwidth}
        \includegraphics[width=\textwidth]{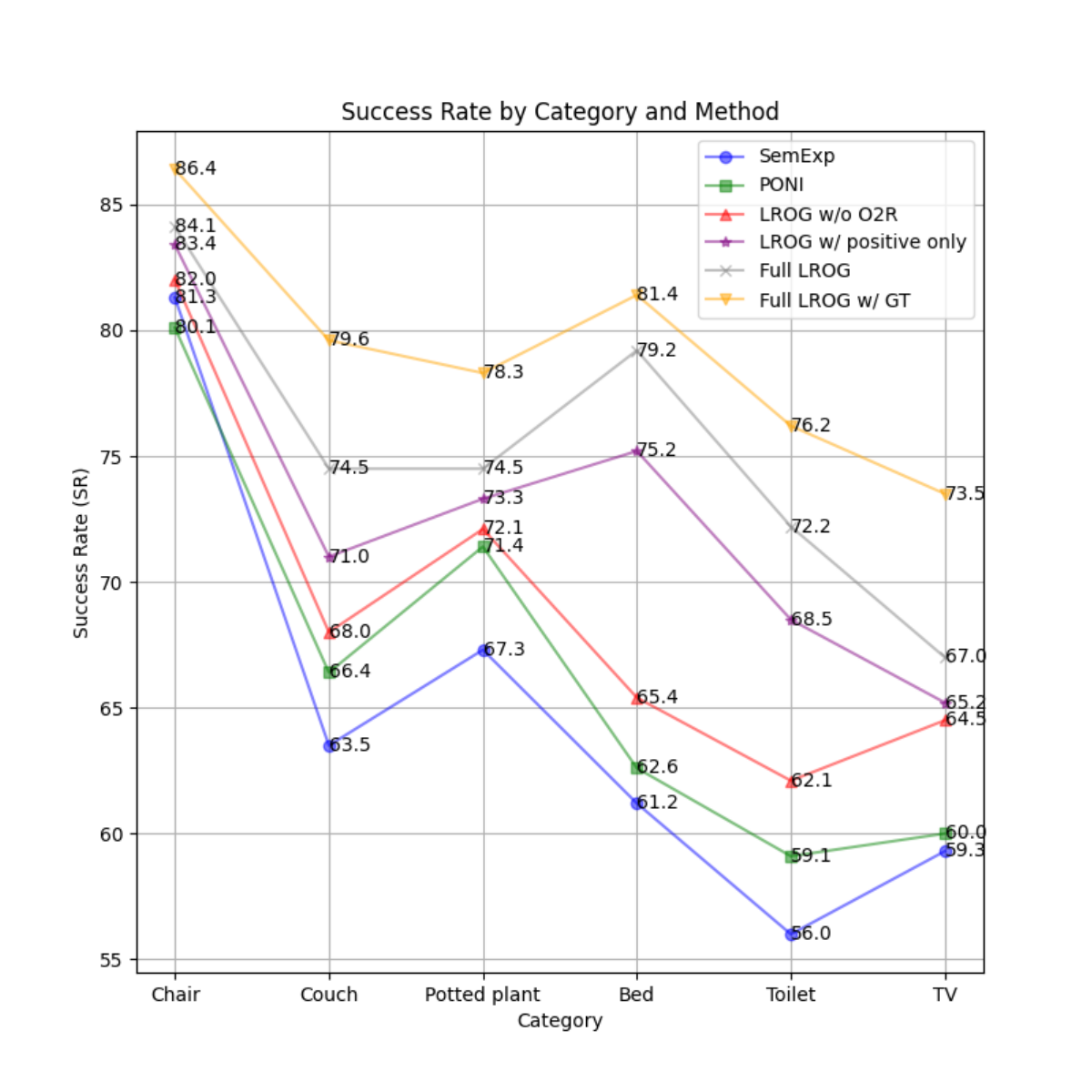}
        \caption{SR across different categories and approaches/modules }
        \label{FIG:SR}
    \end{subfigure}
    \hfill
    \begin{subfigure}[b]{0.35\textwidth}
        \includegraphics[width=\textwidth]{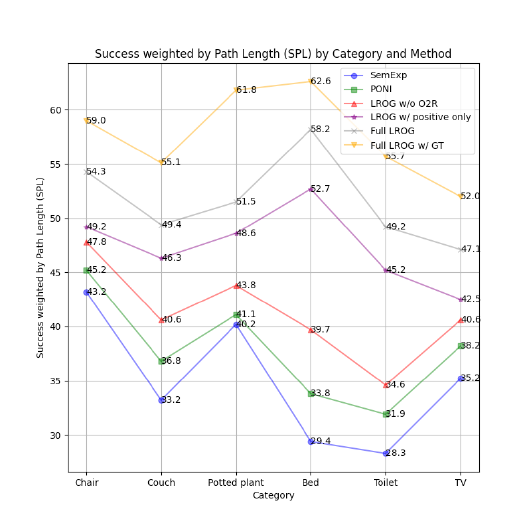}
        \caption{SPL across different categories and approaches/modules }
        \label{FIG:SPL}
    \end{subfigure}
    \caption{Ablation study on SR and SPL comparisons for different categories.}
    \label{FIG:category_eva}
\end{figure}

$\bullet$ (3) What is the performance across different object categories, and which components contribute most to each category?

We collected 200 testing episodes randomly from both the Gibson and MP3D datasets and evaluated the SR and SPL for six target objects listed in Fig. \ref{FIG:category_eva}. We found that the categories of "bed" and "toilet" are significantly improved compared to "chair" after we deployed the LLM-based O2R decoders, even when using only positive promptings. This improvement is attributed to "bed" and "toilet" being distinct, room-specific objects, meaning they predominantly appear in bedrooms and bathrooms, respectively. In contrast, "chair" is a ubiquitous object with a relatively high likelihood of appearing in multiple rooms. Additionally, after utilizing the GT segmentation, the performance in the "potted plant" category has significantly improved. The possible reason is that potted plants are more likely to appear in corners of the house, which are difficult to detect successfully."

\begin{table}[!t]
\caption{Ablation study on multi-modal inputs.}
\renewcommand\arraystretch{1.9}
\begin{center}
\scalebox{0.52}{
\begin{tabular}{c|cccc|cc}
\hline
Multi-modal inputs & \multicolumn{1}{c|}{RGB-D images} & \multicolumn{1}{c|}{\begin{tabular}[c]{@{}c@{}}Target object\\ category modality\end{tabular}} & \multicolumn{1}{c|}{\begin{tabular}[c]{@{}c@{}}Room-related \\ modality\end{tabular}} & \begin{tabular}[c]{@{}c@{}}Distance and \\ direction modality\end{tabular} & \multicolumn{1}{c|}{SR} & SPL \\ \hline
1                  & \checkmark                                 &                                                                                            &                                                                                   &                                                                        &                   75.2      &  42.9   \\
2                  & \checkmark                                 & \checkmark                                                                                          &                                                                                   &                                                                        &                    75.1     &   43.1  \\
3                  & \checkmark                                 & \checkmark                                                                                          & \checkmark                                                                                 &                                                                        &               77.2          & 43.9     \\
4                  & \checkmark                                 & 
& \checkmark                                                                                 & \checkmark                                                                      &                     77.3    & 44.0     \\
\textbf{5 (ours)}           & \checkmark                                 & \checkmark                                                                                          & \checkmark                                                                                 & \checkmark                                                                     &     78.3                    &   44.6  \\
6                  & \checkmark                                 & \checkmark                                                                                          & \checkmark                                                                                 &                                                                        &              78.0           & 44.1     \\ \hline
\end{tabular}
}
\end{center}
\label{tab:multi-modal}
\end{table}

$\bullet$ (4) Does all the modality is useful for LROGNav?

We separate the multimodal inputs into four types of embeddings: RGB-D images (as in PONI), target object category modality, room-related modality, and distance with orientation modality, as listed in Table \ref{tab:multi-modal}. Compared to adding only the target object embeddings, the combination of target object with room-related embedding improves performance more (see No.2 and 3). It is worth noting that our framework can work without the room-related modality as inputs (No.2), which means the model is capable of implicitly learning the O2R relation decoder with only object semantic knowledge in the environment. In conclusion, each component contributes to improving the results to varying degrees within the multimodal Swin-Transformer architecture of LROGNav. We also plot the loss curves of the three decoders in Fig. \ref{FIG:losscurve}. Through the visualization, we can observe that the O2R relation potential task is more challenging to train than the object potential task. After incorporating more modalities, the losses associated with the O2R potential decoder and the object potential decoder decrease. This indicates that the multimodal inputs, by utilizing more information, assist in the model's training, thereby achieving better performance than solely using unimodal RGB-D images, as was the case with PONI.

$\bullet$ (5) How does the performance of different combinations of positional-related embeddings?

Except for the channel-wise and modality-type embeddings, which have been introduced in Section \ref{sec:network}, there are also two traditional position embeddings (PE) compared in the Swin-Transformer \cite{liu2021swin}: absolute PE and relative PE. As shown in Table \ref{tab:positional}, we conduct exhaustive ablation studies to select the best PE combination for LROGNav. The results indicate that both channel-wise and modality-type PEs are useful for improving performance, while combining them with relative position bias yields the best results among all combinations. In contrast, the absolute PE degrades the performance (No.3 and 4), which is not suitable for the shifted-window attention mechanism in the Swin-Transformer architecture, echoing findings in \cite{liu2021swin}.

\begin{table}[!t]
\caption{Ablation study on position-related embeddings.}
\renewcommand\arraystretch{1.9}
\begin{center}
\scalebox{0.56}{
\begin{tabular}{c|cccc|cc}
\hline
No.               & \multicolumn{1}{c|}{abs. pos.} & \multicolumn{1}{c|}{rel. pos.} & \multicolumn{1}{c|}{channel-wise} & modality-type & \multicolumn{1}{c|}{SR} & SPL  \\ \hline
1                 &                                &                                &                                   &               & 75.6                    & 42.8 \\
2                 & \checkmark                                &                                &                                   &               & 75.5                    & 43.1 \\
3                 &                                & \checkmark                                &                                   &               & 76.5                    & 43.7 \\
4                 & \checkmark                                & \checkmark                                &                                   &               & 75.8                    & 43.5 \\ \hline
5                 &                                & \checkmark                                & \checkmark                                   &               & 76.5                    & 43.9 \\
6                 &                                & \checkmark                                &                                   & \checkmark               & 76.4                    & 43.7 \\ \hline
7                 & \checkmark                                &                                & \checkmark                                   & \checkmark               & 77.4                    & 44.1 \\
\textbf{8 (ours)} &                                & \checkmark                                & \checkmark                                   & \checkmark               & 78.3                    & 44.6 \\
9                 &                                &                                & \checkmark                                   & \checkmark               & 77.6                    & 44.2 \\ \hline
\end{tabular}
}
\end{center}
\label{tab:positional}
\end{table}

$\bullet$ (6) What is the advantage of data-driven combined with LLM (proposed LROGNav) compared with pure LLM-based approach (L3MVN \cite{yu2023l3mvn})?

\begin{figure}[!t]
    \centering
    \begin{subfigure}[b]{0.46\textwidth}
        \includegraphics[width=\textwidth]{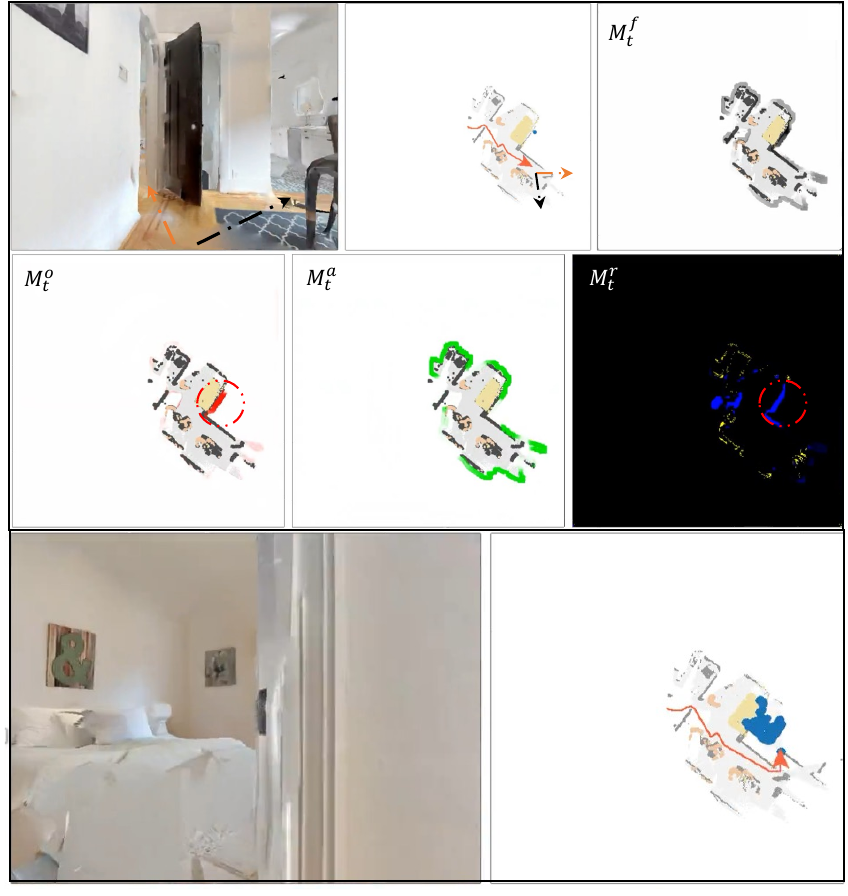}
        \caption{Left-or-right decisions in the Habitat simulator: When there is limited semantic information available, but the left-or-right options need to be decided. The sequence displays real-time RGB observations with trajectory maps, potential maps, the final successful location of the bed, and the entire trajectory.}
        \label{fig:leftrighsim}
    \end{subfigure}
    \hfill
    \begin{subfigure}[b]{0.46\textwidth}
    \hspace*{0.5mm} 
        \includegraphics[width=\textwidth]{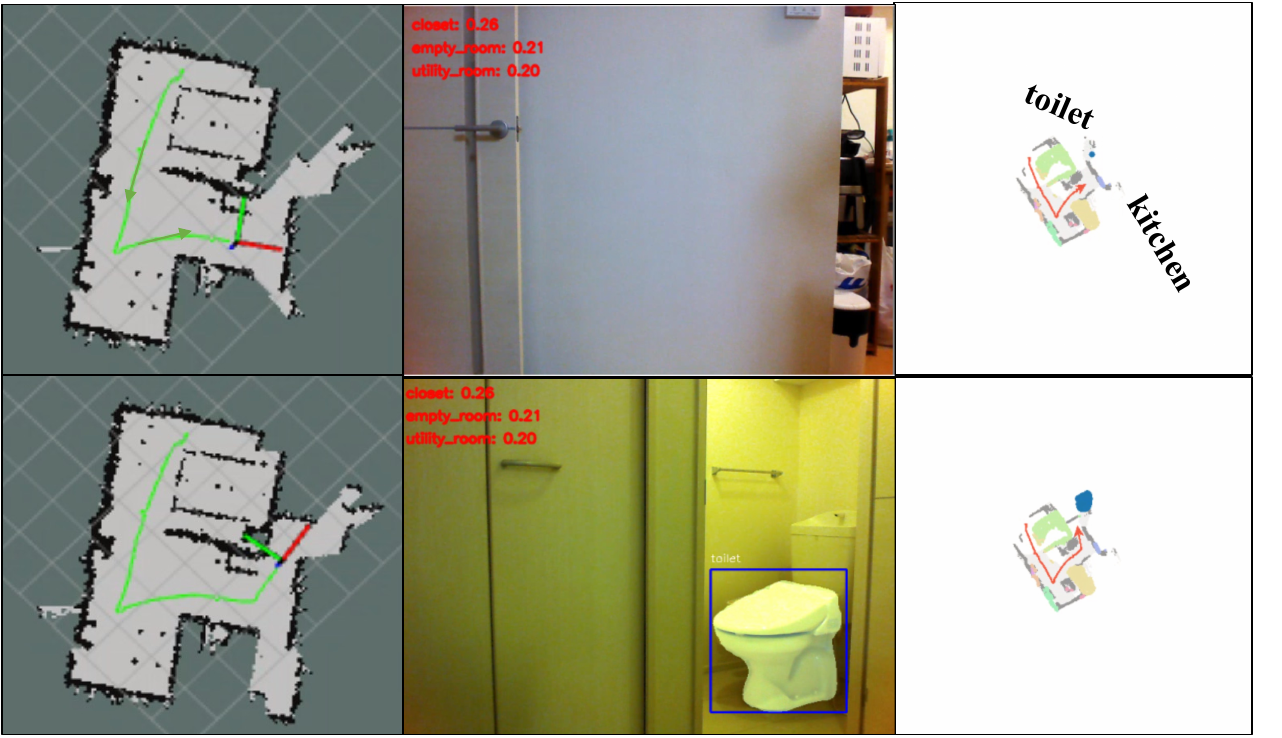}
        \caption{Left-or-right decisions in the Habitat simulator: from left to right, the sequence includes the real-time trajectory map using Hector SLAM, the RGB observation, and the semantic map.}
        \label{fig:lefrightreal}
    \end{subfigure}
    \caption{Left-or-right decisions in Habitat simulator and real-world experiments.}
    \label{fig:leftright}
\end{figure}

We use a 'Left-or-right' example, as shown in Fig. \ref{fig:leftright}, to address this issue. Assume there is only limited semantic information available in front of the agent (e.g., Fig. \ref{fig:leftrighsim} ); a purely LLM-based approach, such as L3MVN \cite{yu2023l3mvn}, needs to decide whether to go left or right for the next step. Theoretically, an LLM-based approach solely relies on the observation at that moment, which has a fifty-fifty chance of making the correct decision. However, the data-driven approach has prior knowledge of the geometry/layout of the house, and also, the historical map as the most significant reasoning source used to make a better prediction that has the potential to be more efficient than the LLM approach.

In the case of the proposed LROGNav, both the object potential map $M_{t}^{o}$ and the O2R relation potential map $M_{t}^{r}$ successfully predict the target object and its related room on the correct left side, thanks to the knowledge from the data-driven training process. In the example of a real-world experiment, as illustrated in Fig. \ref{fig:lefrightreal}, the agent faces a wall with a door to the toilet (the target) on the left and a path to the kitchen (unknown to the robot) on the right. The agent makes the correct decision based on the history map information and knowledge acquired from a large dataset.

In the future, if the LLM-based approach could utilize historical observations through textual promptings, thereby challenging the LLM's ability to reason about spatial geometry, the comparison with the map-based, data-driven approach would become both fairer and more interesting.

$\bullet$ (7) Which part could be improved for the real-world application of ObjectNav?
\begin{figure}[!t]
    \centering
    \begin{subfigure}[b]{0.46\textwidth}
        \includegraphics[width=\textwidth]{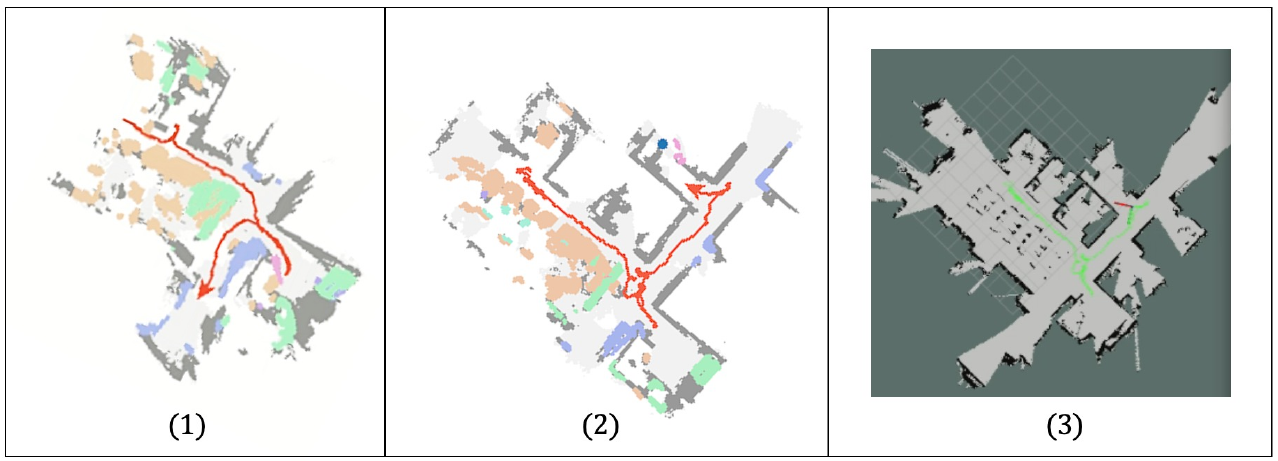}
        \caption{Mapping quality comparisons include: (1) map created using Kobuki's wheel-inertial odometry within LROGNav's mapping module, (2) map generated with LiDAR-based Hector SLAM within LROGNav's mapping module, and (3) map produced with LiDAR-based Hector SLAM in ROS.}
        \label{fig:mappingissue}
    \end{subfigure}
    \hfill
    \begin{subfigure}[b]{0.455\textwidth}
    \hspace*{-0.6mm} 
        \includegraphics[width=\textwidth]{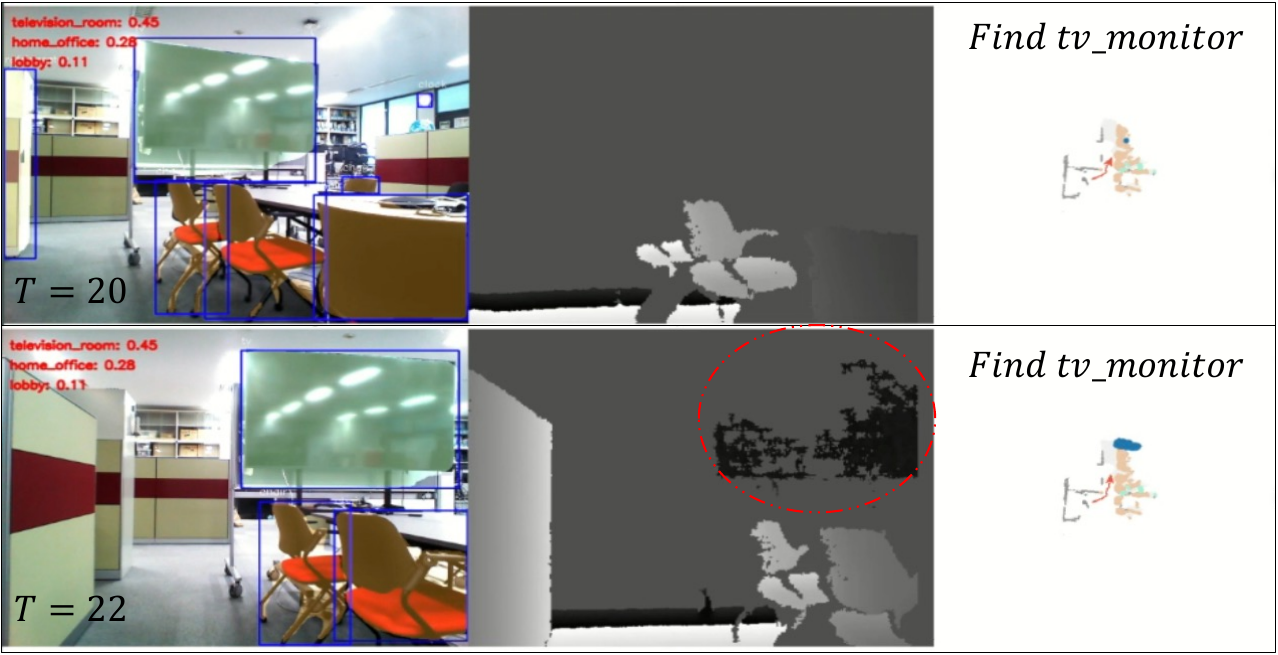}
        \caption{Top row: the target object has been detected in the RGB but without depth information. Botttom row: the target object has been located when depth has been partially detected. }
        \label{fig:depthissue}
    \end{subfigure}
    \hfill
     \begin{subfigure}[b]{0.455\textwidth}
        \includegraphics[width=\textwidth]{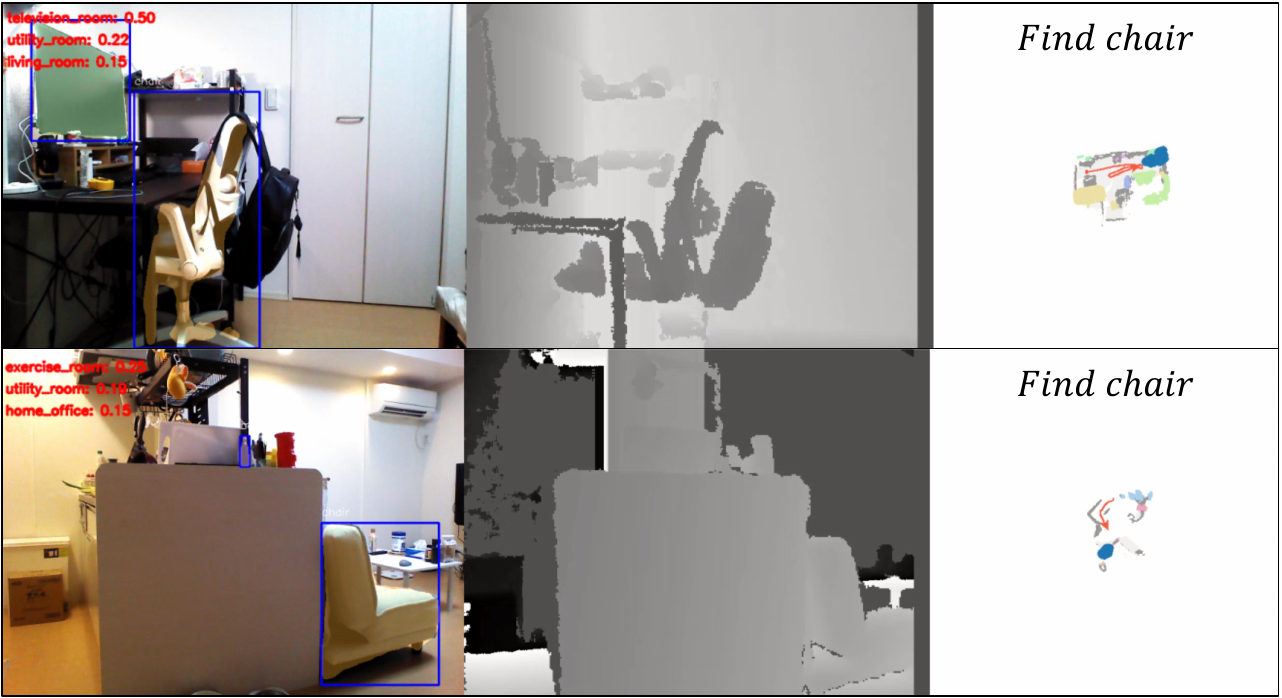}
        \caption{The detection issues. Top row: missing the chair detection results in backtracking. Bottom row: the couch is incorrectly detected as a chair.}
        \label{fig:detectionissue}
    \end{subfigure}
    \caption{Mapping, depth and detection issues in real-world experiments.}
    \label{fig:realissue}
\end{figure}

To address this issue, we discuss three aspects based on observations during real-world experiments. In Fig. \ref{fig:mappingissue}, we illustrate maps generated using different sensors and approaches. To conduct a fair comparison of the sim-to-real gap, as in \cite{gervet2023navigating}, we only use the pose estimation from Hector SLAM, instead of relying directly on LiDAR-based SLAM for mapping. This configuration is necessary since the pose in the Habitat simulator is 100\% accurate and noise-free. Using the more error-prone odometry from the Kobuki move base directly would significantly degrade map quality (see Fig. \ref{fig:mappingissue} (1)).

In the future, we plan to deploy more robust pose estimation algorithms using Visual LiDAR Odometry (VLO), such as V-LOAM \cite{zhang2015visual}, Lidar-Inertial Odometry (LIO), for example, Traj-LIO \cite{zheng2024traj}, and other odometry estimation works ranked highly in the KITTI benchmark \cite{geiger2012we}. Additionally, improvements to the mapping module in map-based approaches are necessary. Currently, most of these approaches extract the mapping module directly from SemExp \cite{semexp}, which merely projects the semantic point cloud to a top-down view and accumulates the map using ego-motion estimation. Compared with current semantic Simultaneous Localization and Mapping (SLAM) technology \cite{yang2022semantic} \cite{hempel2022online} \cite{yang2024enhanced}, there are still gaps to learn from to enhance the quality of semantic mapping. This includes implementing loop closure detection \cite{sun2021visual} and back-end bundle adjustment optimization \cite{schops2019bad}, among others. In conclusion, the map-based approach is a double-edged sword; compared to the end-to-end approach, it exhibits a smaller sim-to-real gap. However, the quality of the map plays a decisive role in the effectiveness of the entire system, echoing findings from recent survey works \cite{wang2023survey} \cite{li2022object}.

The second issue is illustrated in Fig. \ref{fig:depthissue} and relates to depth. Although we use Mask R-CNN for object detection in RGB observations, which has some robustness to reflective objects, there are occasions when a target object, despite being detected in RGB, cannot be captured in depth. This may be due to limitations in the range of the depth camera or challenges with illumination. Consequently, the agent cannot locate the target object on the map, resulting in a loss of efficiency. If we can utilize the signal that the target has already been detected in RGB, or integrate the depth from LiDAR, which is more robust to environmental variations, the performance of system can be further improved.

The last issue concerns object detection, as shown in Fig. \ref{fig:depthissue}. Both examples pertain to object detection problems: the first misses the target while the other represents a detection failure. The possible reasons include a narrow viewpoint (the camera not directly facing the target) or the ability of object detector. To address this issue, we could employ more accurate state-of-the-art (SOTA) object segmentation algorithms, such as Mask-DINO \cite{li2023mask} and DPLNet \cite{dong2023efficient}, among others. Additionally, to address the mis-detection issue, employing a double-check strategy, which involves re-verifying from any viewpoint after the target object has been successfully detected, could be straightforward yet effective.

\section{Conclusion}
In this study, we presented a modular-based ObjectNav framework utilizing common sense knowledge of object-to-room relationships extracted from LLMs, named LROGNav. The chain-of-thought promptings, both positive and negative, are incorporated into GPT to calculate the relationship score between each object and room category, serving as the common sense injected into a room segmentation dataset based on Gibson and Matterport3D. We proposed utilizing multimodal inputs to train a multi-channel Swin-UNet with an encoder-decoder architecture. The results are demonstrated both in the Habitat simulation and in the real world using the Kobuki mobile robot, particularly through realistic demonstrations across various room spaces, efficiently searching for the target. Exhaustive ablation studies verify the function of each component/configuration in LROGNav. The proposed LROGNav achieved an average 10.6\% improvement in SPL compared with the second-best SOTA methods. In consideration of real-world experiments, we discuss the issues encountered when conducting ObjectNav tasks with real robots. We hope that our work will assist the community of embodied navigation in understanding how to apply LLM-based knowledge to the real-world engineering applications. 

While the semantic content in navigation can vary significantly between different environment. For example, the semantic implications of navigating through a crossroad are likely distinct from those of moving around a shopping center. A generalist common sense knowledge extractor based on LLMs, adaptable to different environments, would be an exciting future direction to explore.

\section*{Declaration of competing interest}
The authors declare that they have no known competing financial interests or personal relationships that could have appeared to influence the work reported in this paper.
\section*{Data availability}
No data was used for the research described in the article.
\section*{Acknowledge}
We acknowledge the support from the research project, JPNP20006, commissioned by the New Energy and Industrial Technology Development Organization (NEDO), and JSPS KAKENHI Grant Number 23H03426 in Japan.
\bibliographystyle{model1-num-names}

\bibliography{cas-refs}

\clearpage

\label{promptings}

\begin{figure*}[!t]
\centering
\begin{minipage}{\textwidth} 
\texttt{Here are the appendix materials, including details about promptings, more examples of dataset generation, and qualitative evaluations.}
\end{minipage}
\end{figure*}
\begin{figure*}[bh]
	\centering
		\includegraphics[scale=.42]{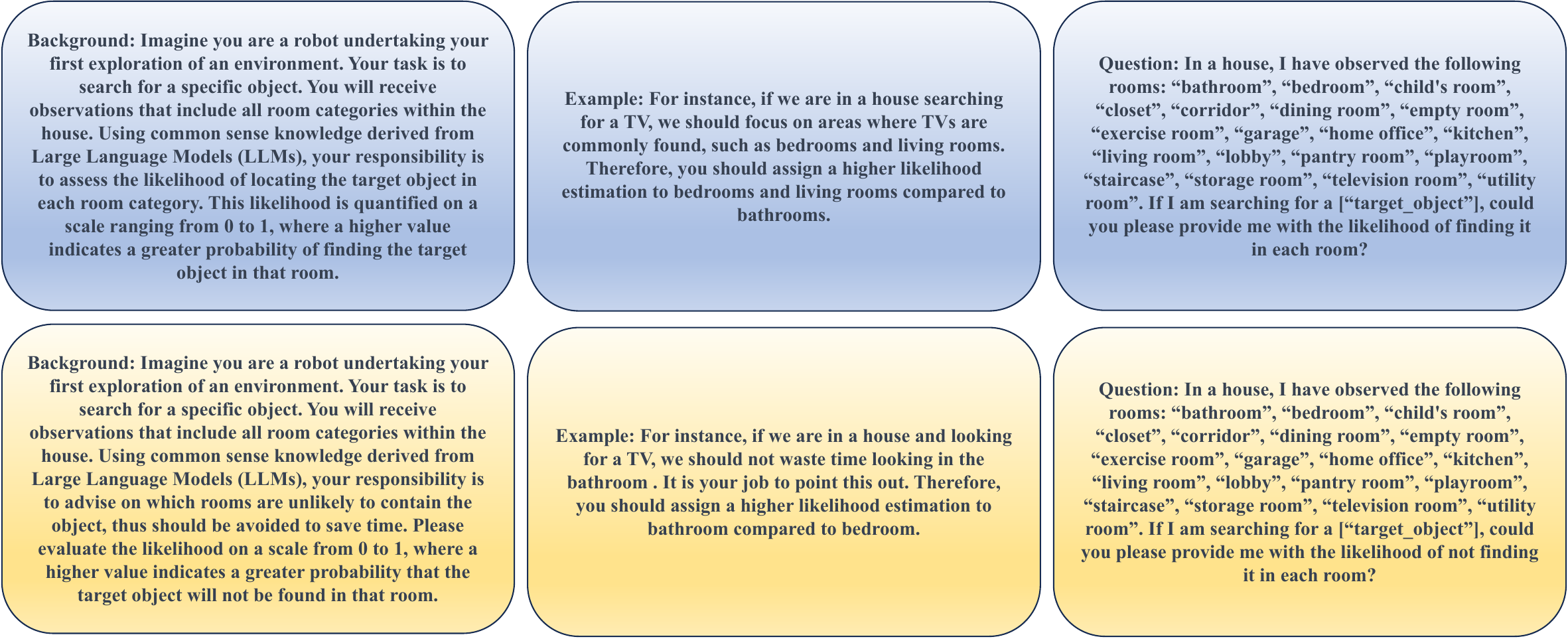}
	\caption{Top: positive promptings, bottom: negative promptings}
	\label{FIG:promptings}
\end{figure*}

\begin{figure*}
	\centering
		\includegraphics[scale=.86]{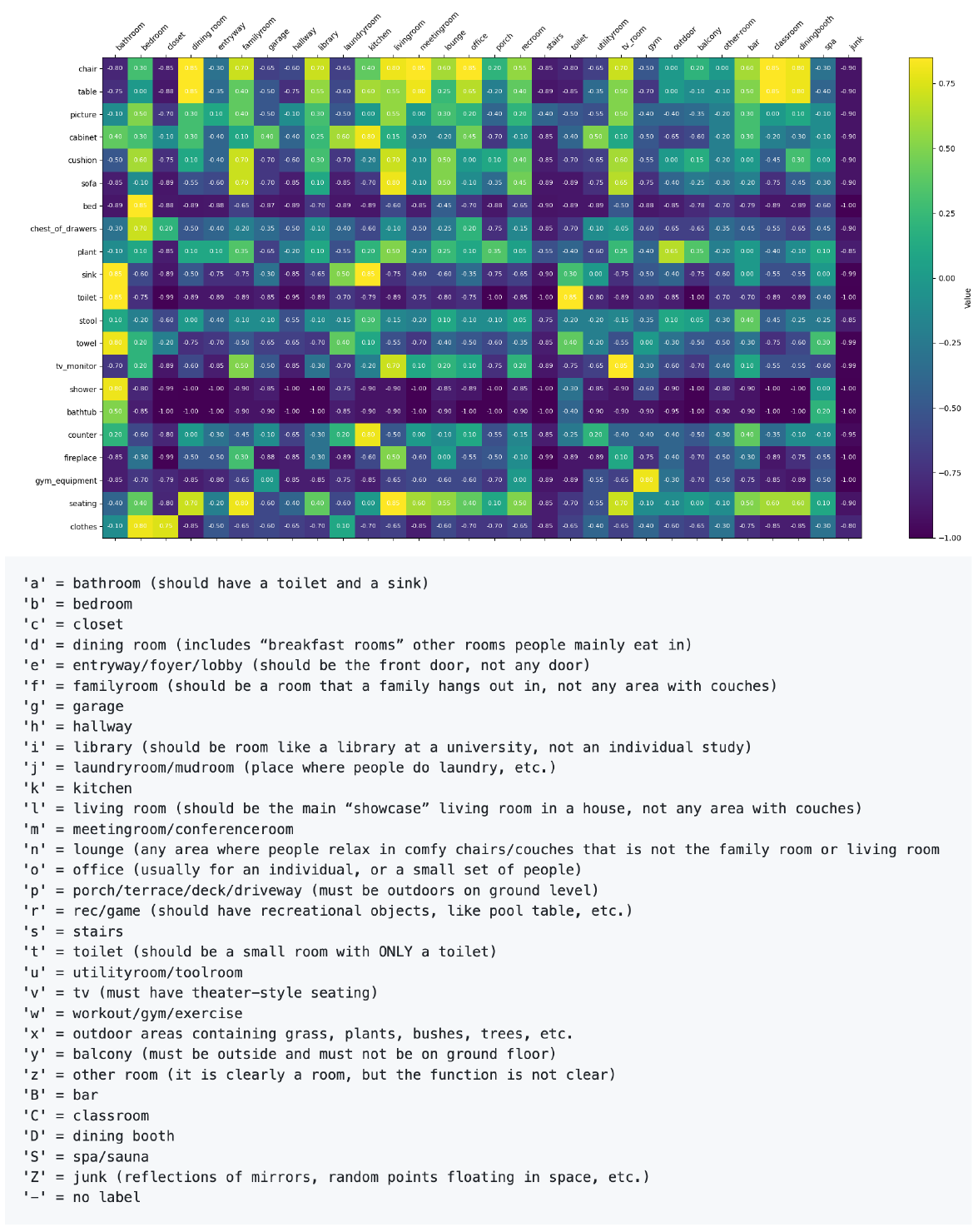}
	\caption{The Object-2-Room relationship matrix of MP3D dataset utilizes LLM-based knowledge. There are conventions for each room/region category, extracted from \url{https://github.com/niessner/Matterport/blob/master/data_organization.md}, which are also used by the LLM for reasoning about common sense knowledge.}
	\label{FIG:matterpord3d_matrix}
\end{figure*}

\begin{figure*}
	\centering
		\includegraphics[scale=.85]{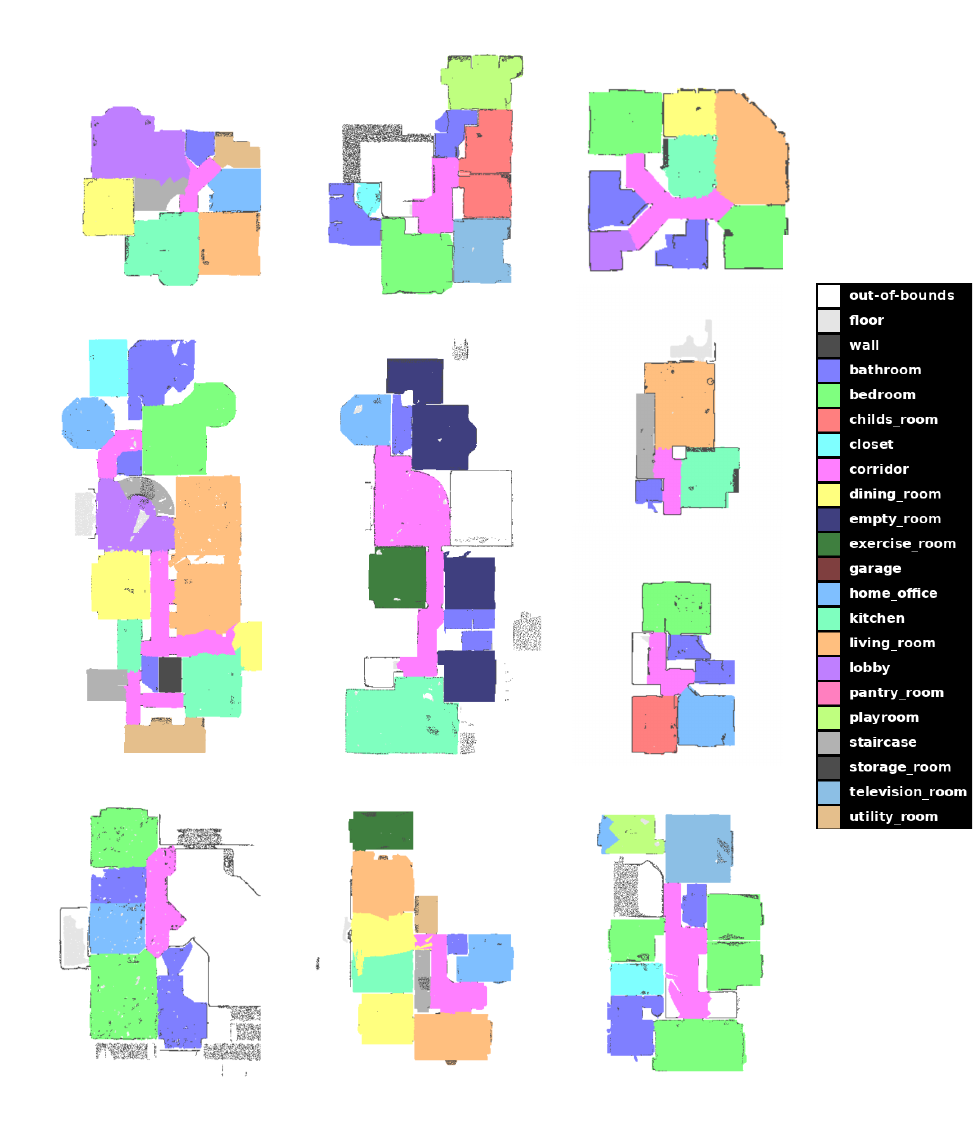}
	\caption{Room semantic segmentations derived from Gibson dataset}
	\label{FIG:gibson_room_semantic}
\end{figure*}

\begin{figure*}
	\centering
		\includegraphics[scale=0.55]{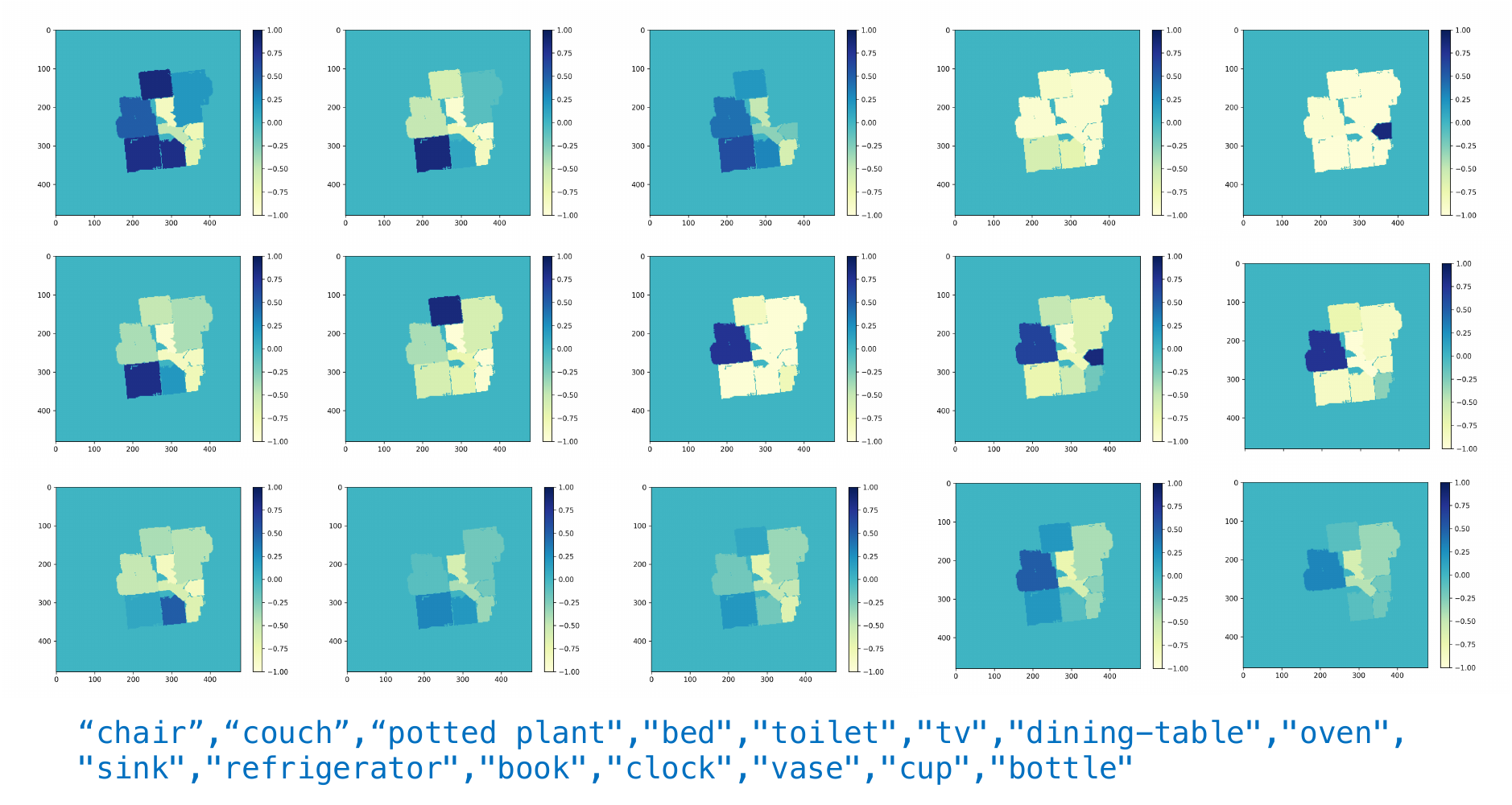}
	\caption{LLM-based complete object-to-room maps for 15 object categories which list in the bottom of figure.}
	\label{FIG:obj_room_wholemap}
\end{figure*}

\begin{figure*}
	\centering
		\includegraphics[scale=0.52]{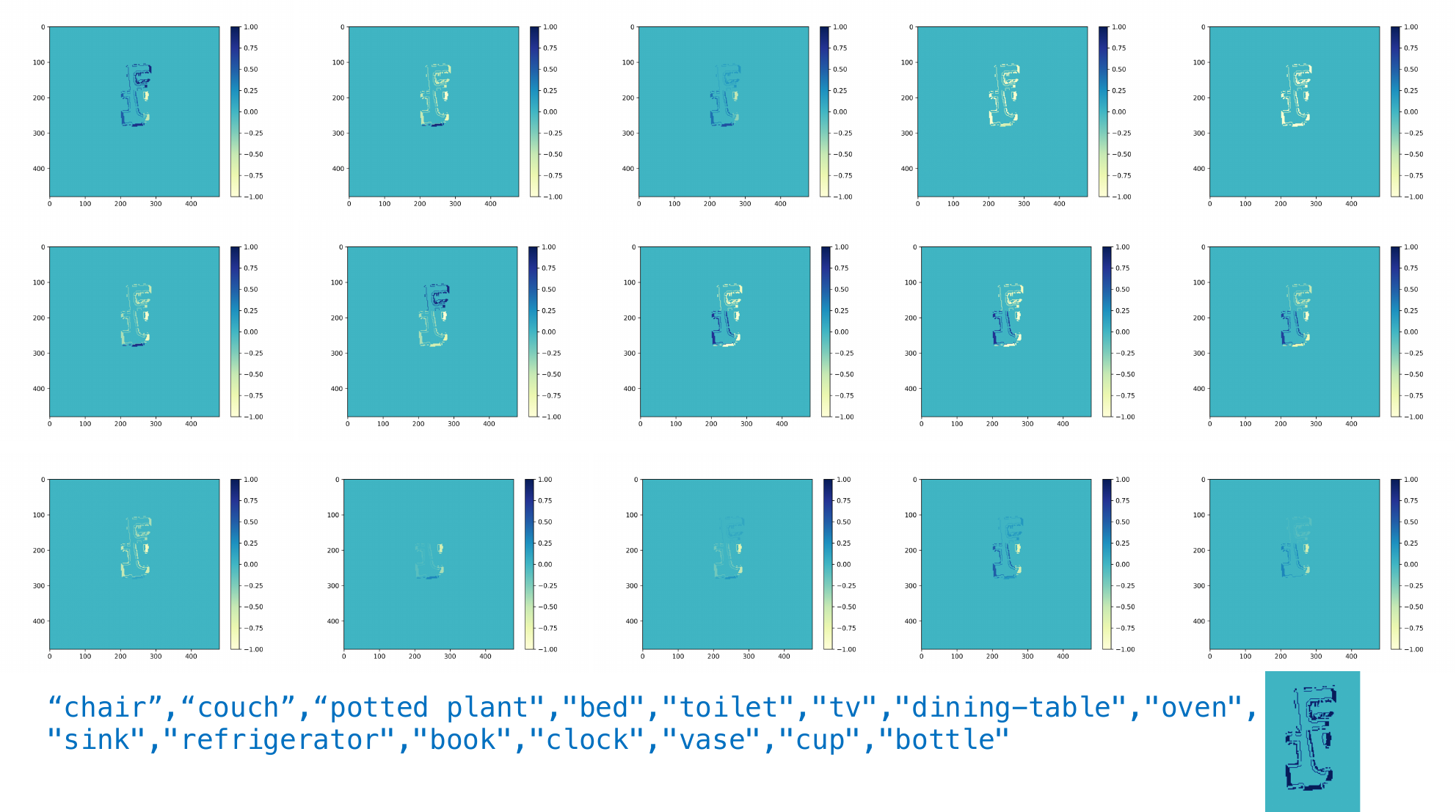}
	\caption{One partial object-to-room map for 15 object categories, extracted from Fig. \ref{FIG:obj_room_wholemap}, is illustrated. The figure at the bottom right shows the shape of this partial map, then the LLM-based O2R relationship score assigned to the frontier respectively.}
	\label{FIG:obj_room_partialmap}
\end{figure*}

\begin{figure*}
	\centering
		\includegraphics[scale=0.85]{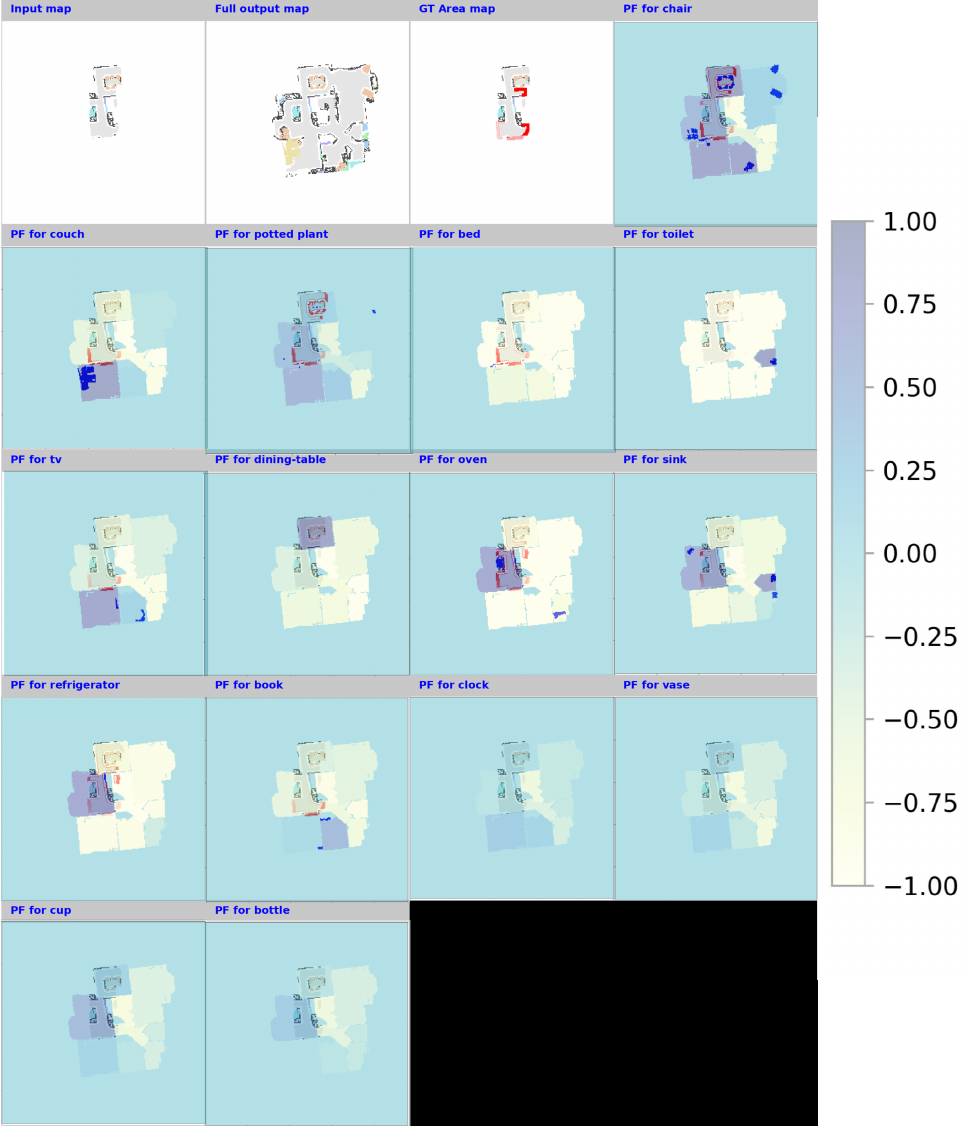}
	\caption{LLM-based complete object-to-room map overlaps with the ground truth positions of 15 object categories, highlighted in blue. In most cases, the target objects are located in rooms with high LLM-based O2R scores, which could verify that the LLM-based O2R relationship score aligns with common sense knowledge.}
	\label{FIG:obj_room_gt_vis}
\end{figure*}

\begin{figure*}
	\centering
		\includegraphics[scale=.76]{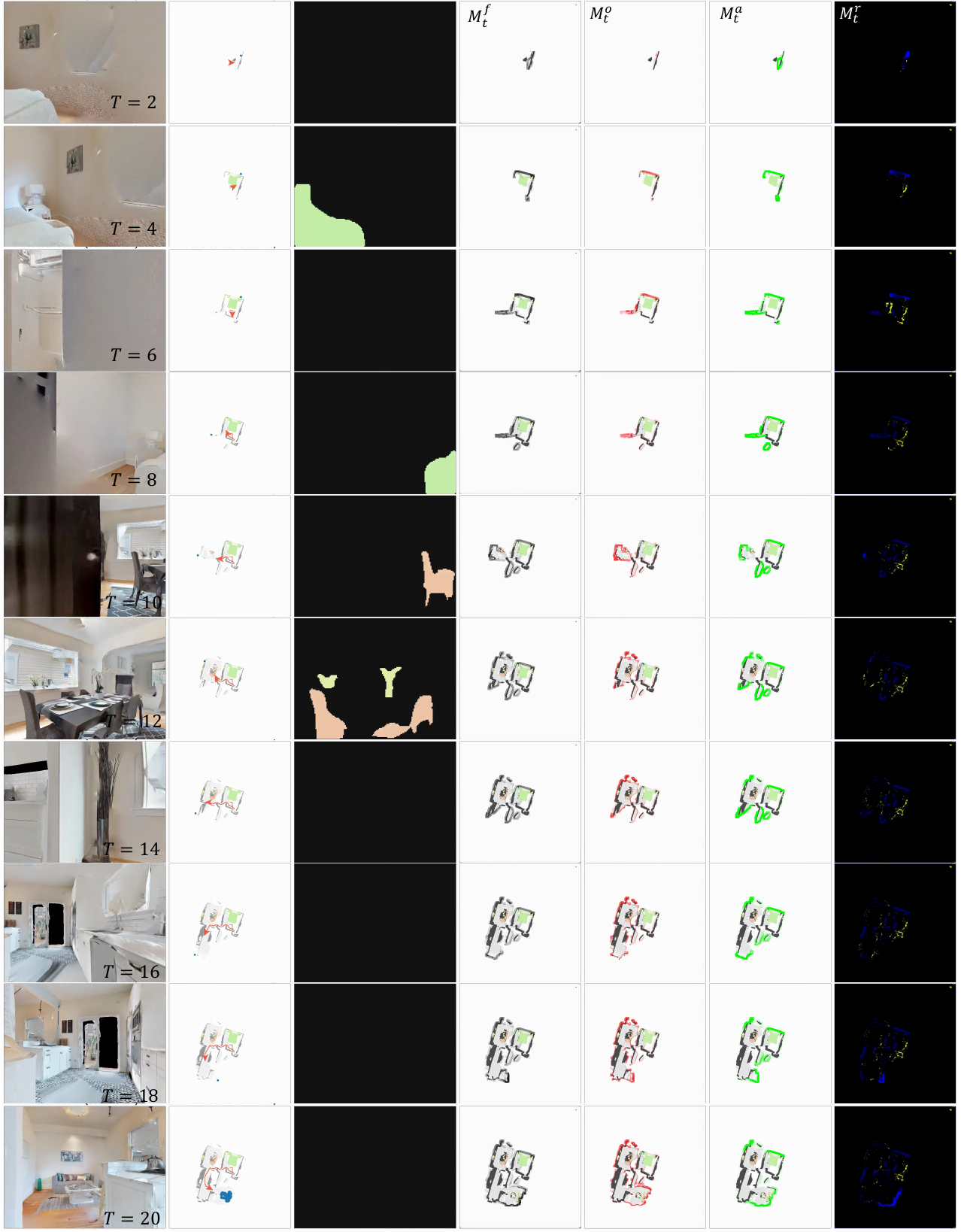}
	\caption{Qualitative examples from different timestamps of LROGNav's "Find couch" task. From left to right in each row: RGB image, real-time map, object segmentation results with Mask R-CNN, fused potential map, and the object, area, and O2R potential maps, respectively.}
	\label{FIG:examples}
\end{figure*}

\end{document}